\documentclass[12pt]{article}

\usepackage{graphicx} 
\usepackage{a4wide}
\usepackage{cancel}
\usepackage{amsmath}
\usepackage{comment}
\usepackage{amsfonts}
\usepackage{minitoc}
\usepackage{titletoc}
\usepackage[T1]{fontenc}
\usepackage[utf8]{inputenc}
\usepackage{rotating}

\usepackage{booktabs}
\usepackage[margin=1in]{geometry}
\usepackage[dvipsnames]{xcolor}%
\usepackage{color}
\usepackage{lscape}
\usepackage{bbold}
\usepackage{umoline}
\usepackage{fancyhdr}
\usepackage{authblk}
\usepackage{chngcntr}
\usepackage{etoolbox}
\usepackage{lipsum}
\usepackage{float}
\usepackage{afterpage}
\usepackage[numbers,sort&compress]{natbib}
\usepackage[section]{placeins}
\usepackage{mwe} 
\usepackage{array}     
\usepackage{tabulary}
\usepackage{caption}
\DeclareCaptionFormat{myformat}{\textsc{#1#2}#3\par}
\usepackage{multicol}
\usepackage{amssymb}
\usepackage{tikz}
\usepackage{sgame}
\usepackage{tabularx}
\usepackage{multirow}
\usepackage[bottom,multiple]{footmisc}
\usepackage[table]{colortbl}
\usepackage{url}
\usepackage{setspace}
\usepackage{pdfpages}
\usepackage{chronosys}
\usepackage{xstring}
\usepackage{ragged2e}
\usepackage{subcaption}  
\usepackage{enumitem}  
\usepackage[hidelinks]{hyperref}
\usepackage[titletoc,title]{appendix}
\hypersetup{
    colorlinks=true,
    citecolor=blue,
    linkcolor=blue,
    filecolor=blue,
    urlcolor=blue,
}
\usepackage{cleveref}
\usepackage{newpxtext,newpxmath}

\makeatletter
\def\ps@pprintTitle{%
  \let\@oddhead\@empty
  \let\@evenhead\@empty
  \def\@oddfoot{\reset@font\hfil\thepage\hfil}
  \let\@evenfoot\@oddfoot
}
\makeatother

\title{\vspace{-1.1cm}\Huge \textbf{
Deep Learning for Crime Forecasting: The Role of Mobility at Fine-grained Spatiotemporal Scales
}}

\author[1]{Ariadna Albors Zumel 
}
\author[1]{Michele Tizzoni}
\author[2]{Gian Maria Campedelli}

\affil[1]{\textit{University of Trento}}
\affil[2]{\textit{Fondazione Bruno Kessler}}

 \date{}
 \vspace{1.5cm}

\begin{document}

\begin{singlespace}
\maketitle
\renewcommand\thefootnote{}
\footnotetext{This paper has been published in the \textit{Journal of Quantitative Criminology}: \href{https://doi.org/10.1007/s10940-025-09629-3}{10.1007/s10940-025-09629-3}}
\renewcommand\thefootnote{\arabic{footnote}}
\end{singlespace}
\vspace*{-2.6em}
\begin{abstract}
   \begin{spacing}{1}

\noindent
\textbf{Objectives:} To develop a deep learning framework to evaluate if and how incorporating micro-level mobility features, alongside historical crime and sociodemographic data, enhances predictive performance in crime forecasting at fine-grained spatial and temporal resolutions.\\
\noindent
\textbf{Methods:} We advance the literature on computational methods and crime forecasting by focusing on four U.S.\ cities (i.e., Baltimore, Chicago, Los Angeles, and Philadelphia). We employ crime incident data obtained from each city's police department, combined with sociodemographic data from the American Community Survey and human mobility data from Advan, collected from 2019 to 2023. This data is aggregated into grids with equally sized cells of 0.077 sq.\ miles (0.2 sq.\ kms) and used to train our deep learning forecasting model, a Convolutional Long Short-Term Memory (ConvLSTM) network, which predicts crime occurrences 12 hours ahead using 14-day and 2-day input sequences. We also compare its performance against three baseline models: logistic regression, random forest, and standard LSTM.\\
\noindent
\textbf{Results:} Incorporating mobility features improves predictive performance, especially when using shorter input sequences. Noteworthy, however, the best results are obtained when both mobility and sociodemographic features are used together, with our deep learning model achieving the highest recall, precision, and F1 score in all four cities, outperforming alternative methods. With this configuration, longer input sequences enhance predictions for violent crimes, while shorter sequences are more effective for property crimes.\\
\noindent
\textbf{Conclusion:} These findings underscore the importance of integrating diverse data sources for spatiotemporal crime forecasting, mobility included. They also highlight the advantages (and limits) of deep learning when dealing with fine-grained spatial and temporal scales.\\

   \end{spacing}
\end{abstract}

\thispagestyle{empty} 

\clearpage
\pagenumbering{arabic} 

\newgeometry{left=1in,right=1in,top=25mm,bottom=25mm}
\section{Introduction}\label{introduction}
The increasing availability of data, coupled with advancements in analytical techniques and the progressive democratization of programming languages, has contributed to
facilitate the computational study of criminal phenomena \citep{campedelli2022machine}. One of the areas that most benefited from the convergence of these three factors is the spatiotemporal prediction of crime, which is particularly important for theory and policy testing and development, as it can shed light on the behavioral dynamics that explain crime occurrence and provide actionable knowledge that can be leveraged to anticipate and prevent crime \citep{predictivepolicing, Berk2021}.

The spatial study of crime has its roots in the first half of the 19th century \citep{cullen2010encyclopedia,turnbull2000atlas,chamarad2006} and became extremely popular in the second half of the 20th century. One of the most widely accepted findings to emerge from this area of research is that crime tends to concentrate in small geographical areas. This observation led to the so-called \textit{criminology of place}, first introduced by \citep{Sherman1989HotSO}, which focuses on the dynamics of crime in microgeographic units within cities, such as addresses, facilities, street segments, or small clusters of street segments \citep{Sherman1989HotSO, WEISBURD}. Since then, the growing interest in these micro-dynamics has led to practical applications in crime prevention, such as hot spots policing \citep{hotspots}.

Along with the spatial study of crime, another factor that has long been considered in the empirical and theoretical literature is the temporal dimension of crime \citep{Mohler2011SelfExciting,Sleeuwen2021}. The two dimensions, space and time, are intrinsically interconnected: both components of a crime event are fundamental due to their correlation as time constraints result in space constraints \citep{Andresen2015}. Given this fundamentally indissoluble link, crime forecasting systems rely on both spatial and temporal information in order to deliver more accurate predictions.

However, beyond the spatial and temporal aspects of crime, many other factors play a role in shaping its concentration and patterned nature. One of the most studied factors is the social context, often examined through dimensions such as socioeconomic inequalities and ethnic diversity. Poverty, for instance, often fosters conditions that increase the likelihood of crime by limiting people’s access to opportunities and resources \citep{Sharkey2016poverty,Jarjoura2002,Graif2017,Berk1980Poverty}. Meanwhile, neighborhoods with significant ethnic diversity sometimes experience crime patterns shaped by the social tensions and discrimination that residents face \citep{legewie2016contested, olzak1992dynamics}. \par
Besides these socio-economic dimensions, another central phenomenon impacting the spatiotemporal concentration of crime is human mobility, as suggested by various theories uncovering its link to crime incidents \citep{CohenFelson1979,SampsonGroves1989}. For example, areas with high foot traffic or frequent visitors often experience more crime due to the increased presence of potential victims and offenders. Conversely, more deserted areas can foster crime due to the absence of individuals who can discourage offenders from committing crimes \citep{CohenFelson1979}. Similarly, shifts in population dynamics—such as the influx of new residents or temporary workers—can disrupt established community networks, creating opportunities for criminal behavior \citep{SampsonGroves1989}.\par
Due to the nonlinear and multidimensional spatiotemporal dynamics that often characterize crime patterns, quantitative criminologists---along with scholars from other fields--- have thus recently become increasingly interested in the promising properties of machine and deep learning methods. Compared to more traditional statistical methods, these algorithms have, in fact, often shown to be more effective in capturing complex patterns, outplaying less flexible methods in prediction and forecasting tasks \citep{reviewpaper}. However, despite the growing interest in these technological trends and the growing recognition of the importance of mobility in explaining crime, as of the current date, there exist very few studies on crime forecasting employing deep learning models and combining historical crime data with human mobility data. Moreover, such studies use relatively large geographic units of analysis (e.g., census tract or police beats) or have a large temporal granularity (i.e., weeks, months, or years), resulting in predictions that cannot easily yield actionable recommendations. Additionally, these studies often focus on only one city or employ just one year of historical data, leading to models with low generalizability to different contexts \citep{Kounadi2020review,MobilityDLWu,stec2018forecasting,Mandalapu_2023}.\par
To address these gaps, in this research we propose a deep learning framework---based on Convolutional Long Short-Term Memory (ConvLSTM) layers---to perform spatiotemporal crime forecasting in microgeographic units, utilizing historical crime data in conjunction with micro-level mobility and sociodemographic data from 2019 to 2023 across four U.S.\ cities. More specifically, the main contribution of this work is that, to the best of our knowledge, this is the first study to use mobility data in spatiotemporal crime forecasting with very fine spatial and temporal resolutions (0.077 sq. miles and 12 hours, respectively), covering a period of five years of historical data and including four different cities. This design thus enables a broader comparative analysis, allowing for a more in-depth exploration of the benefits and limitations of predicting crime occurrences at highly detailed geographical and temporal levels. This, in turn, advances our understanding of crime both theoretically and practically, while further investigating whether and how artificial intelligence can serve as an effective ally for criminology. Our short-term forecasting effort, focused on hourly time scales and fine-grained spatial resolution, is motivated by two interrelated goals. By testing the predictive capabilities of deep learning approaches, we aim not only to advance theoretical insights into the mechanisms underlying crime occurrence and distribution, but also to generate practical knowledge that can inform policy decisions and situational interventions. These may include, for example, shift-by-shift allocation of patrol officers, but also extend to non-policing measures, such as adjusting lighting in high-risk areas, modifying urban infrastructure to reduce anonymity or escape routes, deploying community outreach teams during peak risk hours, or temporarily altering the accessibility of specific spaces (e.g., parks, transit hubs) based on dynamic risk assessments.

The remainder of the paper proceeds as follows. In Section~\ref{background}, we outline the main theories that motivate the use of mobility data for crime forecasting and our contributions to the literature. In Sections~\ref{data} and \ref{methodology}, we describe the datasets used, the extensive data processing carried out, and the deep learning algorithm we designed to perform the crime forecasting task. In Section~\ref{results}, we evaluate the performance of the model and the impact of using mobility data on forecasting outcomes. Lastly, in Section~\ref{discussion}, we discuss the results, highlight the limitations of this study, and address its ethical implications.

\section{Background}\label{background}
\subsection{Theoretical framework}\label{theory}
\subsubsection{Crime concentration}\label{concentration}
The primary framework of this study is the one stemming from the \textit{law of crime concentration}, proposed by Weisburd in his 2014 Sutherland Address to the American Society of Criminology. It states that ``for a defined measure of crime at a specific microgeographic unit, the concentration of crime will fall within a narrow bandwidth of
percentages for a defined cumulative proportion of crime" \citep{WEISBURD}. This law was inspired by early descriptive findings on the study of crime that appeared in the 19th century as well as empirical evidence amassed over the last decades of the 20th century and claims that most crime concentrates in small areas regardless of the city \citep{Braga2017}. The law of crime concentration, one of the most universally accepted features of crime as a human phenomenon, is among the many applications of the \textit{Pareto principle}, which posits that while numerous factors may contribute to an observed outcome, a small subset of these factors often accounts for the majority of the observations \citep{paretolaw}. This principle has been shown to govern many social phenomena. For instance, it appears in education, particularly in the dynamics of classroom disruption \citep{skiba2006zero}; in network theory applied to social influence in online communities \citep{watts2007influence}; in the field of economics, specifically regarding wealth distribution \citep{piketty2014capital}; and in other areas of criminology, for example, asserting that most crimes in a population are committed by a small number of individuals \citep{terence2003deli}.\par
We frame the present study starting from the law of crime concentration because, if crime is clustered in space, it follows that there are statistical patterns to be inferred. Learning these patterns, along with their sources of variation, is the key goal of our forecasting framework. \par
However, while the law of crime concentration represents the fundamental intuition behind this work, there exist complementary theoretical traditions that should be taken into consideration to more comprehensively understand how concentration occurs and, importantly, how to explain and predict concentration in space and time. Modern urban environments are increasingly shaped by factors such as commuting patterns, population flows, gentrification, and urban sprawl \citep{Antipova2018city}: these fluid phenomena are crucial to capturing crime patterns, making it imperative to incorporate such dynamic elements into the study of crime, especially when seeking to forecast crime emergence.
\subsubsection{Crime and human mobility}\label{mobility}
As a social phenomenon, human mobility both influences and is influenced by the environments people engage with. These dynamics also play a critical role in the occurrence of criminal events, which arise from the interplay between offenders, potential targets, and their surroundings.  Certain conditions—such as a lack of guardianship, transient populations, or economic disadvantage—can amplify the likelihood of crime occurring. Therefore, to better understand these processes, researchers have proposed theories that examine how human mobility, social structures, and neighborhood characteristics contribute to the spatial and temporal distribution of crime.

For instance, crimes are more likely to occur when offenders and potential targets converge in environments that lack capable guardianship. This concept is central to \textit{Routine Activity Theory} (RAT), introduced by \citep{CohenFelson1979}. According to RAT, offenders act when they possess both the motivation and the capacity to commit a crime, often selecting victims either deliberately, based on characteristics like identity or possessions, or opportunistically, based on accessibility or vulnerability. Guardians—such as police officers, family members, or institutions like law enforcement—serve as deterrents but must be able to act effectively upon witnessing criminal behavior. The spatial dimension of RAT is further refined by \textit{crime pattern theory} \citep{Brantingham1993}, which incorporates elements from \textit{geometric theory}. Geometric theory explores how the built environment shapes the geographic distribution of crime, leading individuals to form awareness spaces consisting of their main routine activity nodes \citep{geometrictheroy}. Hence, the intersection of offender awareness spaces and potential crime opportunities is where crime hot spots will arise \citep{piquero2015handbook, WEISBURD}. RAT is therefore important for crime forecasting and for this study specifically, as it provides a conceptual framework for understanding the relationship between human mobility patterns and the dynamics of crime opportunities. Based on these theoretical prepositions, incorporating mobility data can enhance predictive accuracy by allowing us to detect changes in routine activities (e.g., population density, and business hours) that influence the likelihood of crime. The theoretical postulations regarding the relationship between crime incidents and human mobility have been backed by extensive empirical evidence 
\citep{stec2018forecasting,Bogomolov2015,kadar2018,DeNadai2020}. Benefiting from the growing integration between traditional data and more innovative sources (for a review see \cite{luca2023crime}), these studies usually include mobility features in the form of \textit{footfall}\footnote{The footfall is defined as the number of people in a given location during a defined time span.}, showing that this feature can ameliorate the performance of crime forecasting models when compared to models that only use historical crime features.

Persistent crime patterns, however, are also closely linked to the socioeconomic conditions of neighborhoods. Communities with high levels of poverty, residential instability, and ethnic diversity often experience weakened social networks and limited informal social control. These conditions form the basis of the \textit{Social Disorganization Theory} (SDT) developed by the Chicago School \citep{SampsonGroves1989}. Social disorganization fosters environments where subcultures of crime and deviance flourish, further contributing to higher crime rates \citep{smelser2001international}. Despite some challenges and inconsistencies in its empirical research \citep{SocialDisorganizationTheory1,SocialDisorganizationTheory2}, SDT remains highly relevant as it provides a lens through which we can investigate how human mobility, neighborhood dynamics, and social structures intersect to influence crime rates. Hence, using mobility data in combination with sociodemographic data allows us to more comprehensively capture patterns and trends, as well as changes in these patterns and trends, occurring in urban environments. Leveraging both dimensions, we argue, would improve our ability to understand when and where crime will happen.

\subsection{Contributions to crime forecasting}\label{literature}
Crime forecasting research employing advanced computational methods encompasses a range of scales and methodologies, each designed to address specific contexts and objectives. A review of the past decade's literature reveals five core elements in study design. Here, we outline our contributions and innovations with respect to each of these elements.\par
\paragraph{\textit{Element 1: What urban focus?}}\mbox{}\\
Firstly, most studies focus on a single city in the U.S., such as New York City or Chicago \citep{Duan2017DeepCN,cesario2024,Zhao2022,Wheeler2021}. A smaller number of studies aim for broader comparability. For example, \citep{MobilityDLWu} analyzed crime patterns in four U.S.\ cities highlighting the benefits of comparative analysis in diverse urban environments. Another notable example is \citep{rotaru2022}, which examined event-level predictions of urban crime across eight U.S.\ cities. Drawing inspiration from these multi-city approaches, our study investigates four U.S.\ cities to enhance generalizability of our results, seeking to underscore common patterns and potential differences across diverse urban areas.

\paragraph{\textit{Element 2: What geographical scale?}}\mbox{}\\
Secondly, the choice of geographic unit, which reflects a fundamental trade-off between spatial detail and computational accuracy, has mostly focused on relatively large areas of 1 km$^2$ or larger \citep{MobilityDLWu, kang2017,stec2018forecasting,kang2017,Wheeler2021}. Despite this, there are some studies that used very fine geographic units of analysis. For instance, \citep{drawve2016predictability} used a highly detailed grid made of square cells measuring 91 $\times$ 91 meters in their study of gun violence in Little Rock, AR, and \citep{zhuang2017crime} used a grid of cells measuring 183 $\times$ 183 meters, approximately 0.122 km² to crime incidents in Portland, OR. However, in such studies, the corresponding temporal granularity was relatively large (i.e., six months and two weeks, respectively). Our analysis adopts a spatial unit of 0.077 sq. miles,\footnote{0.077 sq. miles are equivalent to 0.2 sq. kilometers in the metric system.} offering a compromise that captures micro-spatial patterns effectively while remaining computationally manageable, also avoiding excessive sparsity, which would have rendered any forecasting exercise theoretically and practically useless.

\paragraph{\textit{Element 3: What temporal granularity?}}\mbox{}\\
As anticipated in the previous paragraph, the temporal granularity of predictions further distinguishes contributions in this field. Many studies focus on daily predictions \citep{MobilityDLWu,Duan2017DeepCN,WangYuan2019,kang2017,stec2018forecasting,Xia2021}, while some explore larger granularities. For instance, \citep{magnusson2023} employed a coarser approach, generating yearly forecasts for gun violence. Similarly, \citep{lin2018} examined monthly predictions to highlight long-term trends. Our work focuses on a 12-hour prediction interval, a granularity never explored in the literature in combination with fine spatial resolution. By narrowing the prediction window and using a very fine spatial granularity, we aim to provide more actionable insights for real-time resource allocation and urban safety measures, underscoring not only the potential of this method but also its limits.

\paragraph{\textit{Element 4: What temporal focus?}}\mbox{}\\
Temporal coverage of the historical data is another critical dimension, influencing the ability to capture trends and variability. Many studies rely just on one year of data, focusing on short-term predictions \citep{MobilityDLWu,ChenLiao2023,Liang2021}, while few others try to capture long-term patterns by extending their analysis to eight years or more \citep{WangYuan2019,cesario2024}. Other researchers, including \citep{magnusson2023} and \citep{Duan2017DeepCN}, settled on datasets spanning five years, finding a balance between temporal depth and data availability. Our study aligns with this medium-term focus, leveraging five years of data to identify meaningful patterns.

\paragraph{\textit{Element 5: How many data sources?}}\mbox{}\\
Lastly, recent advancements in crime forecasting have increasingly emphasized the importance of integrating diverse data sources. While traditionally previous studies often relied exclusively on crime data \citep{WangYuan2019, Duan2017DeepCN,rotaru2022,Xia2021}, more recent approaches highlight the added value of contextual information as well. In general, the use of mobility data has increased substantially in the past years in criminology \citep{BrowningHumanMobilityCrime2021a}, and has played a major role in the computational social science revolution \citep{LazerComputationalSocialScience2009a}. Mobility is measured through various sources, including transportation data \citep{RosesSimulatingOffenderMobility2020a}, GPS tracking \citep{WangCrimeRateInference2016a, HuangDeepCrimeAttentiveHierarchical2018c}, and social media or other types of digital traces \citep{MallesonSpatiotemporalcrimehotspots2015, kadar2018, HippUsingSocialMedia2019a}, and it allows researchers to capture individual and collective behaviors and flows at fine-grained geographical and temporal scales. Despite this growing interest and the evident potential of mobility data to enrich crime analysis, their role in short-term crime forecasting remains underexplored and largely unquantified. In particular, we still lack systematic evidence and empirical estimations of the extent to which incorporating mobility features can enhance the accuracy of predicting when and where crimes are likely to occur—an insight that would be highly valuable to both scholars and practitioners. For this reason, our study embraces a multidimensional perspective, integrating crime, mobility, and sociodemographic data to capture the complex interplay of social, spatial, and temporal dynamics in crime occurrence, in the attempt to disentangle the contribution that each dimension provides to the forecasting ability of deep learning.\\

\noindent
Therefore, by building on these five elements, our study advances the field by adopting a multi-city framework with exceptionally fine spatial and temporal resolution, utilizing five years of data and integrating three distinct data sources. These design choices align with recent advancements while pushing the boundaries of crime forecasting research.

\section{Data}\label{data}
\subsection{Sample Selection}
In this research, we analyzed four U.S.\ cities: Baltimore, MD (Bal); Chicago, IL (Chi); Los Angeles, CA (Las); and Philadelphia, PA (Phi). These cities were chosen due to their high crime rates and their good coverage of mobility data (the specific criteria we defined and used are explained in detail in Section~\ref{mobility-data}). In addition, their diversity with respect to location, urban structure, and sociodemographics made them particularly appealing for our comparative analysis between cities. For instance, Baltimore is a port city focused on shipping and trade, with a significant African American majority. In contrast, Los Angeles is nearly six times larger, characterized by a sprawling car-dependent structure, and is predominantly home to Latino and Asian communities. A more detailed overview of the four cities can be found in Table~\ref{description-cities}. Using multiple cities represents a relevant feature of this study since, as previously mentioned, most research on crime forecasting focuses on only one city, making it difficult to argue for the applicability of those models to all contexts.

\begin{table}[ht]
    \fontsize{11pt}{11pt}\selectfont
    \centering
    \begin{tabularx}{0.94\textwidth}{ccccccccc}
    \toprule
      & Region & Population & Area & Black & Hispanic & Asian & Poverty & Crime rate\\
    \midrule
    Bal & Northeast & 565,239 & 210 & 61.2\% & 5.9\% & 2.6\% & 19.6\% & 6,997\\
    Chi & Midwest & 2,664,452 & 591 & 28.8\% & 29.0\% & 7.0\% & 16.9\% & 4,381\\
    Las & West & 3,820,914 & 1,216 & 8.6\% & 48.1\% & 11.8\% & 16.6\% & 3,332\\
    Phi & Northeast & 1,550,542 & 347 & 40.1\% & 15.7\% & 7.6\% & 22.7\% & 4,037\\
    \bottomrule
    \end{tabularx}
    \caption{Comparison of the four U.S.\ cities we selected. Information obtained from the U.S.\ Census Bureau QuickFacts for 2023, except for the crime rates, which were obtained from FBI's UCR data and represent the crime rates per 100,000 people in 2019. The area is in sq. kilometers.}
    \label{description-cities}
\end{table}
In the following subsections, we present the three types of data sources used to create city-specific datasets, as well as the data processing that had to be performed for each of them.

\subsection{Crime data}\label{crime_data}
We used city-wide crime datasets containing crime incidents from 2019 to 2023, publicly available through the Open Data portals of the four cities. The final versions of the datasets contained four variables: the date and time of the crime incident, the category of the crime incident, and the location of the crime given as latitude and longitude coordinates.\par
Although several crime categories were available depending on the city, we first selected those that were common across all four settings and, among these, the ones more likely to be reported to the police. This latter criterion was important to ensure that the crime data we used for the analysis were not severely impacted by the risk of underreporting, which would have introduced non-trivial issues to our forecasting exercises \citep{darkfigures,Akpinar2021,Wu2020}. Specifically, we selected the following crime categories: burglary, motor vehicle theft (MVT), assault, homicide, and robbery. The first two are property crimes, while the remaining three are violent crimes. Our analysis is thus three-fold: we developed models forecasting all crime together, as well as models focusing on property and violent crimes separately.
The percentages of each type of crime and the total number of crime incidents for each city are shown in Table~\ref{crime-types}. From these data, we observe that assault is the most common crime in all four cities, accounting for around half of the total incidents, while homicide is the least common.

\begin{table}[h]
    \fontsize{11pt}{11pt}\selectfont
    \centering
    \begin{tabularx}{0.82\textwidth}{lcc@{\hskip 1em}ccc@{\hskip 1em}cc}
    \toprule
    & \multicolumn{2}{c}{Property Crimes} && \multicolumn{3}{c}{Violent Crimes} &  \\
    \cmidrule(r){2-3} \cmidrule(l){5-7}
    & Burglary & MVT && Assault & Homicide & Robbery & Total incidents\\
    \midrule
    Bal & 15.14\% & 18.53\% && 53.75\% & 1.22\% & 11.36\% & 132,002\\
    Chi & 14.81\% & 29.68\% && 38.00\% & 1.29\% & 16.23\% & 268,627\\
    Las & 15.01\% & 24.73\% && 51.19\% & 0.39\% & 8.68\% & 438,905\\
    Phi & 11.15\% & 16.91\% && 61.36\% & 0.84\% & 9.73\% & 265,097\\
    \bottomrule
    \end{tabularx}
    \caption{Percentage of each crime type over the total amount of crime incidents of those categories between 2019 and 2023 for each of the four U.S.\ cities we selected.}
    \label{crime-types}
\end{table}

\begin{figure}[h]
  \centering
  \includegraphics[width=1\columnwidth]{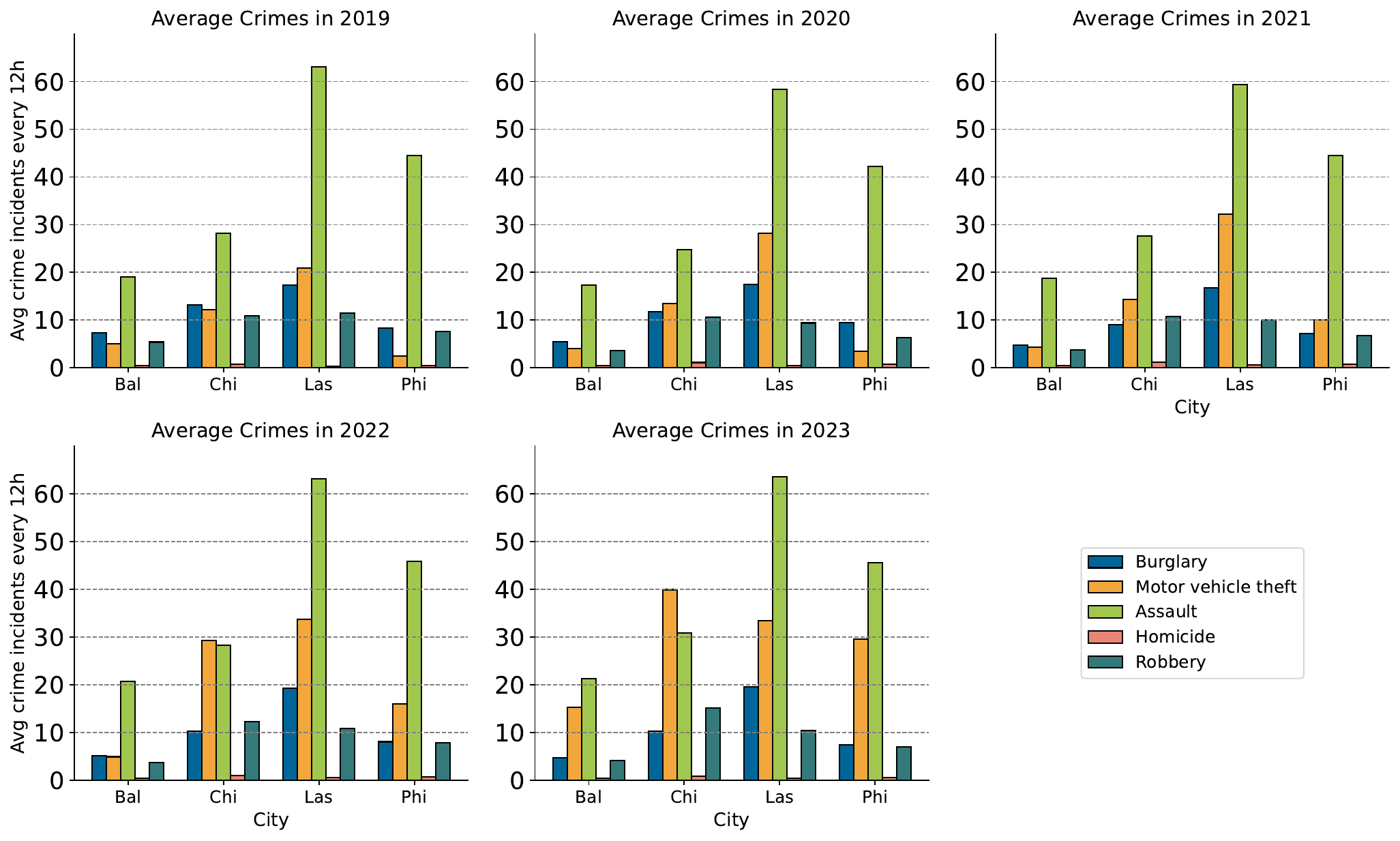}
  \caption{Yearly average of crime incidents recorded every 12 hours for each city and crime type.}
  \label{dailyavg}
\end{figure}

Furthermore, since our main focus is on 12-hour blocks (half-days) for crime predictions, Figure~\ref{dailyavg} presents the average number of crime incidents every 12 hours for each city. This visual representation reveals that most types of crime average fewer than 40 incidents per day, except assault, which exceeds this threshold in Los Angeles and Philadelphia. Furthermore, homicide averages are, as expected, much lower than the other crime categories across all four cities, as shown in Table~\ref{crime-types}, potentially posing a further challenge for our forecasting model due to data sparsity. These findings are consistent across all five years of data.\par
Using this dataset, we generated a single feature represented as a binary variable: 0 if no crime occurred and 1 if at least one crime occurred within each spatial unit of analysis. This approach was chosen due to the extreme sparsity of the data (clearly illustrated in \ref{crime_distrib}), which made using a binary variable effectively equivalent to examining the actual crime counts. In Baltimore, the percentage of instances---i.e., spatial cells at a given time---with no crime (0s) was, on average, 97.96\%$\pm$0.73, while for Chicago, Los Angeles, and Philadelphia, the corresponding percentages were 99.03\% $\pm$ 0.34, 99.39\% $\pm$ 0.19, and 98.38\% $\pm$ 0.62, respectively. This significant imbalance, with the overwhelming majority of cells experiencing no crime (0s), makes the prediction task particularly challenging since the scarcity of positive instances limits the algorithm's ability to learn effectively from these events, increasing the risk of overfitting or bias toward predicting no crime.

\subsection{Mobility data}\label{mobility-data}
The source of micro-level mobility data was Advan, a location intelligence company that is mainly focused on providing micro-level data to the financial and real estate industries. They provide anonymized and GDPR\footnote{General Data Protection Regulation.} and CCPA\footnote{California Consumer Privacy Act.} compliant mobile data from 100 million cellphones, offering a wide set of information about a given location, such as foot traffic and demographics \citep{advanwebsite}. Advan data starts from 2019 and is updated weekly. Specifically, we exploited their ``Weekly Patterns" dataset \citep{advan2022foot}, which contains data from 11,590,987 points of interest (POIs) distributed across the U.S.\ and Canada. POIs are significant for various reasons, such as their cultural, commercial, or logistical importance. Common examples of POIs include hotels, restaurants, gas stations, stadiums, churches, shopping malls, parks, hospitals, and other places that attract visitors or serve as hubs of social activity in urban contexts.

In order to select the subset of cities to be used for this study, we were aware that not only the number of POIs was relevant, but also the relationship POIs had with population size and the city's area since population density and POI density allow for a better understanding of the potential human mobility coverage of each POI. For this reason, we defined a variable called the \textit{people-to-POI ratio}, representing the number of people per POI. This ratio was calculated by dividing the population density by the POI density, hence indicating the number of people per POI within 1 km$^2$. Therefore, since this is a normalized value, it offers a fairer way to compare the cities, particularly in terms of the richness of the data in capturing as much as possible the underlying mobility dimension of a city. We computed this ratio for the main 735 cities in the U.S.\ and obtained a list ordered by ascending people-to-POI ratio. From this list, we focused on cities with high crime rates, verified that (a) each did not fall in the upper extreme of the distribution in terms of people-to-POI ratio\footnote{A very high people-to-POI ratio would indicate likely poor coverage and representativity of POIs in a urban context.} and (b) each had complete crime data from 2019 to 2023 publicly available. This resulted in the final selection of four U.S.\ cities mentioned in the previous subsection (i.e., Baltimore, Chicago, Los Angeles, and Philadelphia). Table~\ref{geo-attributes} presents the people-to-POI ratio for each of the four cities, with a lower ratio indicating better POI coverage. In \ref{poi_ratio}, we visualize the histogram of the distribution of the people-to-POI ratio for all 735 cities.

\begin{table}[h]
    \fontsize{11pt}{11pt}\selectfont
    \centering
    \begin{tabularx}{0.75\textwidth}{lcccc}
    \toprule
     & Area & Population density & Num.\ of POIs & People-to-POI ratio\\
    \midrule
    Bal & 210 & 2,930 & 28,218 & 21.78\\
    Chi & 591 & 4,581 & 95,049 & 28.46 \\
    Las & 1,216 & 3,273 & 102,478 & 38.80 \\
    Phi & 347 & 4,518 & 55,496 & 28.25\\
    \bottomrule
    \end{tabularx}
    \caption{Relevant geographical attribute of each of the four cities selected. The area is in $km^2$, the population density in people per $km^2$, the Num.\ of POIs is the total number of unique POIs available for that city (not all of them were used during all four years), and the people-to-POI ratio in people per POI per $km^2$.}
    \label{geo-attributes}
\end{table}

From this dataset, we constructed several mobility features, which we classified into two groups. First, we used the variable ``top\_category'' to assign each POI type to one of 11 categories, following the NAICS system\footnote{NAICS, or the North American Industry Classification System, is a standardized framework used by businesses and governments to classify industries and organize economic data across North America.}. The 11 categories are: \textit{(1) Utilities and construction, (2) Manufacturing, (3) Retail and Wholesale Trade, (4) Transportation and Warehousing, (5) Business and Professional Services, (6) Educational Services, (7) Health care and Social Assistance, (8) Arts, entertainment, and recreation, (9) Accommodation and food services, (10) Public Administration, (11) Other Services}. More details are provided in \ref{secA}. Using the variable ``visits\_for\_each\_hour'', we then computed the \textit{footfall} for each POI category. These features allowed us to obtain dynamic quantities reflecting how many people visited each POI category in each cell during each time window, thus representing the underlying mobility characteristics of all our cities, mapping potential time-specific patterns at the daily level (i.e., dynamics changing between day and night), as well as meso-level (i.e., different dynamics on different days of the week) and macro-level (e.g., structural changes due to gentrification) seasonality and trends.

Second, we calculated the POI category diversity using the \textit{Shannon Diversity Index}\footnote{The Shannon Diversity Index (also called the Shannon-Wiener Index) is used to characterize entity diversity in a given community or set of elements. The higher the index value, the greater the entity diversity.}, defined as:
\begin{equation}
H = -\sum_{i=1}^{n}p_i \cdot \ln(p_i),
\end{equation}
where $p_i$ represents the proportion of each POI category $i$ within each spatial unit of analysis, and $n$ is the total number of POI categories (11 in our case). We computed this measure to map the heterogeneity of a given spatial unit in terms of human activities carried out, hypothesizing that this information can be valuable in understanding its potential crime attractors or targets, as well as representing the complexity of its human and social fabric. To calculate the proportion of each POI category within each spatial unit, we first determined the total number of POIs across all categories within each cell by summing the individual counts. The proportion for each category was then computed by dividing the count of that category by the total count within the cell. This process yielded a total of 12 features: 11 representing footfall per POI category and one representing POI diversity. Maps displaying the distribution of POI diversity for each of the four U.S.\ cities can be found in \ref{poi_diversity}.

\subsection{Sociodemographic data}
Besides crime and mobility data, we enriched our feature space by including a range of sociodemographic variables. These features allow us to take into consideration the broader social context of each geographic unit beyond the type of activities and premises that are captured by our mobility data.\par
The data we used were obtained from the American Community Survey. Specifically, we used the ACS 5-year estimates for 2019, 2020, and 2021\footnote{\url{https://www.census.gov/data/developers/data-sets/acs-5year.2021.html\#list-tab-1036221584}}\footnote{The data for 2022 and 2023 had not been released at the time of this study, so we decided to use the 2021 data for both years, as we do not expect significant variations.}.
These datasets contain more than 20,000 variables related to the social, economic, demographic, and housing characteristics of the U.S.\ population for different geographic granularities (e.g., region, county, county subdivision, and block group).\par
Firstly, we chose the block group as our geographic level since it is the smallest granularity available. Afterward, we selected 26 variables related to gender, age, race/ethnicity, employment, income, education, and marital status—features that have been identified as correlated with crime in the scientific literature \citep{socioecvars, socialfactorscrime, flowers1989demographics}. The specific variables are listed in \ref{secB}. Their raw values (in the case of medians) or percentages (in the case of counts) were used as our 26 sociodemographic features.

\section{Methodology}\label{methodology}
\subsection{Data preparation}\label{preprocessing}
\begin{figure}[h]
        \centering
        \includegraphics[width=0.77\textwidth]{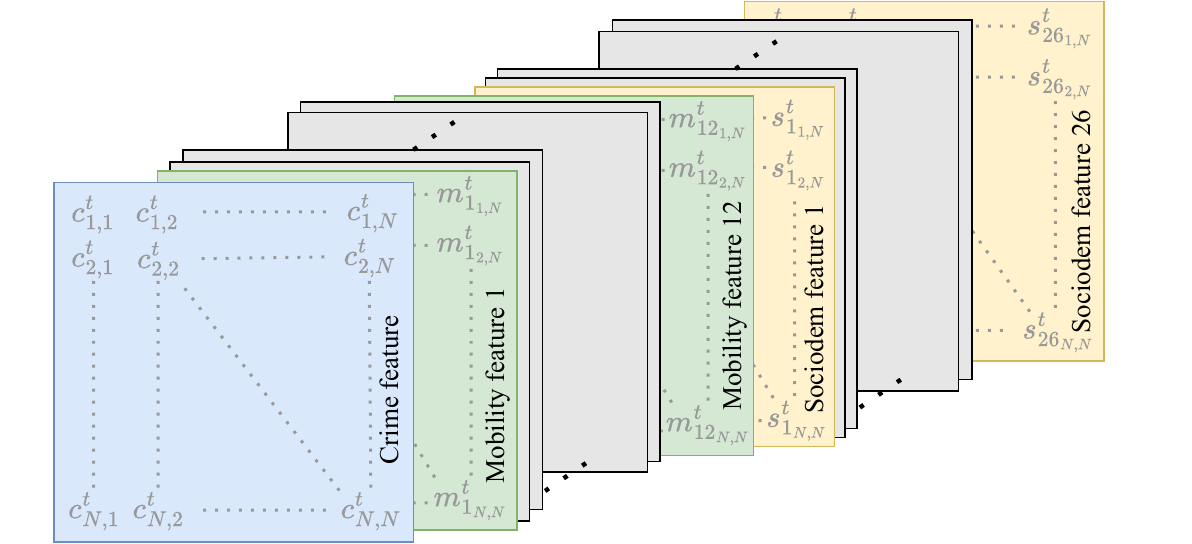}
        \caption{Illustration of the structure that we used for our $N\times N$ frames with 39 channels for each 12-hour block $t$. The first channel corresponds to the crime feature, the following 12 channels are the mobility features, and the last 26 channels are the sociodemographic features.}
        \label{fig:datastructure}
\end{figure}

To formalize the microgeographic unit of analysis, we used a grid-based approach, which consisted of defining a grid containing the city borders and then calculating the values of all our features for each spatial cell during every 12-hour period throughout the study's time span. Therefore the problem definition becomes as follows: Consider that we divide a city into an $N\times N$ grid (with $N$ varying for each city to ensure that each cell has an area of 0.077 sq. miles), hence into a matrix of the form $\mathbf{G}=(g_{i,j})_{0< i,j\le N}$. Then for each grid cell $g_{i,j}$, we compute 39 sets of predictive features for each 12-hour block. The first predictive feature is the historical crimes, where we have that the crime feature at the 12-hour block $t$ is represented by $\mathbf{C}_{i,j}^t=\{c_{i,j}^{t-T},c_{i,j}^{t-T+1},\dots,c_{i,j}^{t-1}\}$ where $T$ is our look-back period, hence how many blocks of 12 hours before $t$ we use to predict this feature at the 12-hour block $t$. The second set is the 12 mobility features, which are denoted as $\mathbf{M}_{i,j}^t=\{m_{y_{i,j}}^{t-T},m_{y_{i,j}}^{t-T+1},\dots,m_{y_{i,j}}^{t-1}\}$ for each mobility feature $y$ with $y=1,2,\dots,12$ and for each 12-hour block $t$. Lastly, we have the 26 sociodemographic features, hence for each sociodemographic feature $z$ with $z=1,2,\dots,26$ and each 12-hour block $t$, we get the features $\mathbf{S}_{z_{i,j}}^t=\{s_{z_{i,j}}^{t-T},s_{z_{i,j}}^{t-T+1},\dots,s_{z_{i,j}}^{t-1}\}$\footnote{Since the size of each block group varies significantly, the preprocessing of the sociodemographic features involved identifying the block group(s) that intersected with each cell, and extracting the corresponding sociodemographic data. In cases where multiple block groups overlapped with a cell, a simple average was computed. For cells with no intersection, the value was set to NaN.}. Consequently, and borrowing this example from computer vision, each frame can be interpreted as an $N\times N$ image with 39 channels, where each channel represents a different feature (see Figure~\ref{fig:datastructure}). In other words, our forecasting task can be thought of as similar to predicting the next frame in a video: given a series of previous frames, each characterized by a specific set of features, what will the next frame look like? \par
These frames are then grouped into sequences, with the length determined by the look-back (LB) period $T$. Each feature within these sequences is then normalized using min-max scaling, ensuring consistent value ranges across the dataset. Also, to maintain uniform grid sizes across all four cities and to augment the number of samples available, we randomly sampled five $M\times M$ subsections for each sequence of frames, keeping only the sequences with at least 2 crimes in that area in order to balance the dataset as much as possible. For this study, we set $M=16$, which covers approximately 19.31 sq. miles (50 km$^2$), and used two LB periods of $T=28$ and $T=4$ (equivalent to 14 days and 2 days respectively) to assess the optimal amount of data required to maximize forecasting outcomes. This dimension of the analysis not only bears practical importance but also theoretical relevance: a shorter LB period would indicate that crime dynamics cluster and change in short time frames, while a longer LB would suggest that crime dynamics are more complex and require long-term memory.

Therefore, the goal is to use T consecutive 12-hour frames to predict the crime occurrence at $T+1$, where the crime occurrence is defined as a $16\times 16$ matrix with either $y_{i,j}=0$ (for no crime) or $y_{i,j}=1$ (when there is at least one crime) for each cell $g_{i,j}$. It should be noted that this binary problem is essentially equivalent to predicting crime counts due to the extreme sparsity of the crime data, as discussed in Section~\ref{crime_data}.

Using this approach, since we had a total of 3,640 half-day frames, using a sliding window of length $T+1$ resulted in a total of $3,640-T$ sequences. These sequences were split into training and test sets in a 90/10 ratio, maintaining their chronological order. To prevent data leakage between the last sequences of the training set and the first ones of the test set, the final $T$ sequences from the training set were removed.

\begin{figure}[H]
        \centering
        \includegraphics[width=1\textwidth]{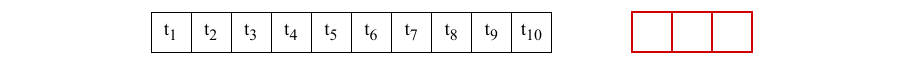}
        \caption{Illustration of a toy dataset with 10 half-day frames (left) and a sliding window of size $T+1=3$ (right).}
        \label{chrono-spliting}
\end{figure}

To understand how this procedure works, suppose we have 10 half-day frames and a sliding window of length three (i.e., $T=2$, meaning that we use the first two frames to predict the crime in the third one), as shown in Figure~\ref{chrono-spliting}.
As we slide our window over the data, we obtain eight (i.e., $10-T$) sequences: $(t_1,t_2,t_3)$, $(t_2,t_3,t_4)$, $(t_3,t_4,t_5)$, $(t_4,t_5,t_6)$, $(t_5,t_6,t_7)$, $(t_6,t_7,t_8)$, $(t_7,t_8,t_9)$, and $(t_8,t_9,t_{10})$. Applying a 90/10 chronological split assigns the first seven samples to the train set and the last one to the test set. Next, to avoid data leakage, we remove the last $T$ sequences of the training set, since they contain frames that are part of the test set. This results in a training set with five sequences and a test set with one sequence.

For our actual dataset, when $T = 14$, this procedure resulted in training and test sets containing 3,222 and 362 sequences, respectively, which would increased to 16,110 and 1,810 samples after extracting the five subgrid sections. However, sequences with no crime events were discarded, leading to the loss of some samples. Additionally, to keep the same sample size for all data configurations (i.e., for the four cities and three crime aggregations), we retained the number of samples corresponding to the smallest dataset among those. This yielded a final training set size of 12,546 sequences and a test set size of 1,510 sequences.

Lastly, before using these datasets for our analysis, we applied a spatial mask to exclude all cells that met at least one of the following criteria: (i) they are located outside city borders, (ii) they consist entirely of water, or (iii) they do not intersect with any census block group. A visualization of the included and excluded cells is provided in \ref{masks}.

\subsection{Convolutional LSTMs}
Our main model is based on ConvLSTM layers, proposed by \citet{shi2015convolutional}. This type of neural network consists of two main components:
Long Short-Term Memory (LSTM) layers and convolution operations.

LSTMs are powerful neural networks designed to handle temporal patterns. For instance, they have been used to predict daily streamflow through weekly forecast horizons \citep{Nearing2024}, for speech recognition by modeling temporal dependencies in audio signals \citep{Graves2013SpeechRW}, and for determining the probability that certain social media content will gain popularity \citep{LEE2012134}. They are able to perform such tasks by using an internal structure referred to as \textit{memory cells}. Each cell contains a \textit{cell state} ($c$), a \textit{hidden state} ($h$), and three types of gates that determine the flow of information using \textit{sigmoid} and \textit{tanh} activation functions. The types of gates are the \textit{forget gate} $f_t$, the \textit{input gate} $i_t$, and the \textit{output gate} $o_t$. Their corresponding equations for the multivariate version of the LSTM (i.e., the fully connected LSTM or FC-LSTM), where both inputs and outputs are 1-dimensional vectors, are as follows:
\begin{align}
    f_t&=\sigma(W_{xf}x_t+W_{hf}h_{t-1}+W_{ci}\odot c_{t-1}+b_f)\\
    i_t&=\sigma(W_{xi}x_t+W_{hi}h_{t-1}+W_{cf}\odot c_{t-1}+b_i)\\
    o_t&=\sigma(W_{xo}x_t+W_{xc}h_{t-1}+W_{co}\odot c_t+b_o),
\end{align}
where $h_{t-1}$ is the previous hidden state, $x_t$ is the current input, $W$ are the weight matrices (where the subscripts indicate the two variables that are connected by this matrix), $b_x$ is the bias for that gate, and $\sigma$ represents the sigmoid function \citep{ZHAO2019486}. The goal of the forget gate is to determine which information from the previous cell state $c_{t-1}$ should be forgotten. The aim of the input gate is to select which new information should be added to the current cell state $c_t$. Lastly, the output gate is used to assess what information should be included in the current hidden state $h_t$.\par
Each candidate cell is formalized through the equations related to the cell state $c_t$ and the hidden state $h_t$:
\begin{align}
    c_t&=f_t c_{t-1}+i_t\tanh(W_{xc}x_t+W_{hc}h_{t-1}+b_c)\\
    h_t&=o_t\odot\tanh(c_t).
\end{align}
These equations highlight how the candidate cell is used in combination with the forget gate and the input gate in order to obtain the current cell state. Subsequently, the cell state and the output gate are employed to obtain the hidden state, which summarizes all the information learned. Lastly, the hidden state is utilized to make the final prediction $y_t$ by using the appropriate activation function for the task at hand.

Therefore, memory cells allow LSTMs to retain only the information that is most relevant to the forecasting task. The advantage of these internal structures is that they trap the gradient\footnote{The gradient is a vector that points in the direction of the steepest increase in a multivariate function.} in the cell, preventing it from becoming too large or too small and thus avoiding in this way the vanishing/exploding gradients problem that significantly limited its predecessor, the recurrent neural networks \citep{Goodfellow-et-al-2016,Sherstinsky2020,CORTEZ2018315}.

However, while LSTMs are effective at capturing temporal information, they are not engineered to account for spatial information. To address this, ConvLSTMs extend traditional LSTMs by replacing the one-dimensional vectors with three-dimensional tensors, where the first dimension represents time and the other two are the spatial dimensions (the number of rows and the number of columns) of the input frame.Therefore, the matrix multiplications in the FC-LSTM equations---used in both the input-to-state and state-to-state transitions---are replaced by convolutional operations, and the weight matrices are replaced by convolutional kernels \citep{shi2015convolutional}.

To illustrate how the introduction of convolutional operations allows the model to preserve spatial information, consider a simple example where the input $x_t$ is a single-channel $3\times3$ spatial grid and the output should be of the same shape. Then the operation $Wx_t$ is done as follows. In the case of LSTM, $x_t$ is flattened into a $9\times 1$ vector and this operation becomes a simple matrix multiplication with W being a $9\times9$ weight matrix. Hence, the operation results in a $9\times1$ vector, which is then reshaped into a $3\times3$ matrix to recover its 2-dimensional shape. Therefore, we can see that the fully connected operation disregards the spatial structure of $x_t$. In contrast, for ConvLSTM, assume $W$ is a $2\times2$ convolutional kernel. The convolutional operation is then performed by sliding this kernel over the input grid and performing element-wise multiplication of its values with the corresponding elements in the input grid, followed by summing the results. This operation is repeated for the rest of the grid, with the kernel sliding across the grid one element at a time in both the horizontal and vertical directions. To ensure the output has the same spatial dimensions as the input, zero-padding\footnote{Zero-padding is the process of adding extra zeros around the border of an input grid before applying the convolution operation. In our example, by adding one layer of zeros around the input, the grid becomes a $5 \times 5$ matrix, allowing the kernel to process 9 regions in total (one for each possible position of the kernel). This ensures that the output remains $3 \times 3$, preserving the spatial dimensions of the input.} is applied. Therefore, this operation results in a $3\times3$ matrix as output, preserving the spatial dependencies of the input, and maintaining the spatial structure of $x_t$ throughout the computation.

Another advantage of this type of neural network is that the use of three-dimensional tensors considerably reduces the number of parameters to be trained when compared to LSTMs. This ability to incorporate both spatial and temporal information makes ConvLSTMs appealing for capturing the complex spatiotemporal dynamical patterns in crime events, hence offering a suitable algorithmic framework for our task.

\subsection{Model architecture}
ConvLSTMs were initially defined for precipitation nowcasting, using past radar sequences to predict future radar frames \citep{shi2015convolutional}. Since then, they have been applied in various contexts, such as quantifying the contributions of climate change and human activities to vegetation change, and predicting short-term traffic flow \citep{Liang2024,Zheng2020}. Therefore, this type of neural network is typically used for video frame prediction, as it accounts for both the temporal and spatial components of the data, making it ideal for the data structure we constructed. While this is usually performed with either grayscale images (1 channel) or RGB images (3 channels)\footnote{Grayscale images have only 1 channel, representing intensity information, while RGB images contain separate channels for red, green, and blue.}, we adjusted the model architecture to train on images with 39 channels (corresponding to our 39 features) and output a 1-channel image.\par
\begin{figure}[h]
    \centering
        \includegraphics[width=0.9\textwidth]{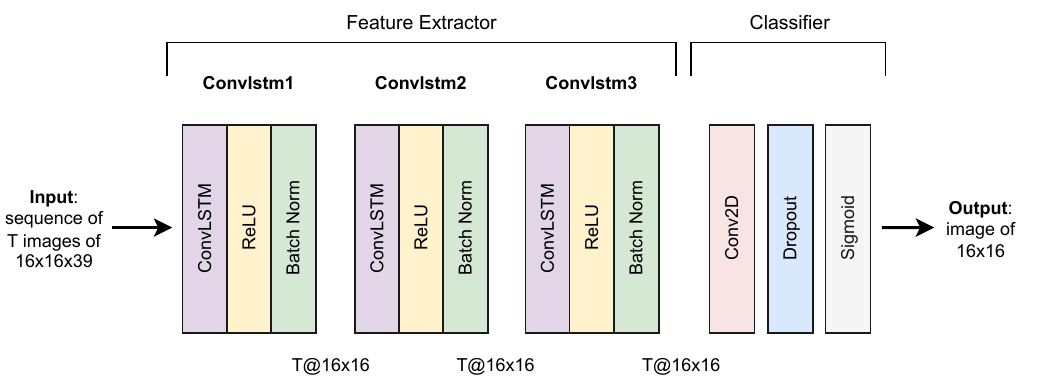}
        \caption{Architecture that we designed for our ConvLSTM model, where $T$ is our LB period.}
        \label{model-architecture}
\end{figure}

In order to engineer our model, we started with a basic architecture for video frame prediction using ConvLSTMs and adapted it to our specific problem. Then we fine-tuned several hyperparameters (see \ref{secC} for the specific values used) until we obtained the model architecture depicted in Figure~\ref{model-architecture}.

As already mentioned, the goal of this model is to predict crime occurrences at a future point in time (time $t$) based on past data. Specifically, our model analyzes a sequence of $T$ frames (from $t-T$ to $t-1$) to make this prediction, with each frame containing 39 channels, one for each feature. The crime occurrence is represented by a $16 \times 16$ matrix with zeros and ones, indicating no crime or at least one crime, respectively. Therefore, the input to the model consists of a sequence of $T$ frames, each with a shape of $16 \times 16 \times 39$ (the first value represents the height, the second the width, and the third the number of channels). This sequence is then processed by the first ConvLSTM block, called \textit{Convlstm1}, which sequentially processes each frame and updates the hidden states.  After processing all $T$ frames, the final hidden states of the ConvLSTM layer contain information about both the temporal and spatial context. This process is repeated through two more ConvLSTM blocks, allowing the network to learn more intricate patterns by capturing different levels of temporal and spatial abstraction. This section of the model is therefore used to extract the main features from the input data. Additionally, a \textit{ReLU} activation function and a \textit{batch normalization} layer are applied at the end of each ConvLSTM block to introduce nonlinearity and to improve the learning process by normalizing the inputs to the network, respectively. We then proceed to the classifier section of the model, where the final hidden states of the last ConvLSTM block are processed by a two-dimensional convolutional layer, which produces the final $16\times16$ output image. This step is followed by a dropout layer, which helps prevent overfitting, and a sigmoid function, which converts the values of each cell into probabilities, resulting in values between 0 and 1 that indicate the likelihood of at least one crime occurring at time $t$. These probabilities are then turned into a binary classification using a threshold of 0.5. We chose not to optimize the threshold because, due to the extreme sparsity of our data, increasing the threshold resulted in a sharp decrease in recall and a marginal increase in precision (see \ref{perf_thrs}).

\subsection{Model evaluation}
Once we obtain the final prediction, we evaluate the model's performance by calculating recall, precision, and F1 score. Specifically, recall indicates how well the model predicts all instances where at least one crime actually occurred, while precision reflects how often the model's prediction of at least one crime is correct. The F1 score then combines precision and recall into a single value, providing a balanced assessment of the model's performance\footnote{Recall, Precision, and F1 score are calculated as follows:
\[
\text{Recall} = \frac{tp}{tp + fn}, \quad
\text{Precision} = \frac{tp}{tp + fp}, \quad
\text{F1} = 2 \cdot \frac{\text{Precision} \cdot \text{Recall}}{\text{Precision} + \text{Recall}}.
\]
}. This evaluation is performed over four different random seeds (i.e., 0, 42, 123, and 999) to control for the role of stochasticity in model training and to obtain the average and standard deviation of each model's performance.

Given the unbalanced distribution in our target variable and the differential cost of wrong predictions between false positives and false negatives, we primarily focus on recall to quantify the predictive performance of the model. This is due to the high costs associated with false negatives, as they signify a failure to allocate crime prevention resources to a location in need.

Besides these two metrics, we also calculate a second version of recall and precision that considers false positives with neighboring positive cells\footnote{We define as neighboring cells all the cells that are touching the target cell. Hence, there can be between 3 and 8 neighboring cells, depending on the target cells' location in the grid.} as true positives. This choice is motivated by two aspects. First, the presence of crime in a location affects the nearby areas (and the same occurs with police presence) and spillover effects may lead to incidents in locations that fall within adjacent spatial cells. Second, given obfuscated georeferenced crime data, crime incidents that actually occurred in a given cell may be recorded differently and therefore fall in a neighboring one: calculating modified metrics would then allow us to predict them without losing such relevant data points. We calculated these two modified metrics as follows. Let $\mathbf{O}=(o_{i,j})_{1\le i,j\le M}$ denote the ground truth crime occurrence matrix and $\mathbf{P}=(p_{i,j})_{1\le i,j\le M}$ the predicted crime occurrence matrix, where $o_{i,j},p_{i,j}\in\{0,1\}$ and $M$ is the size of the matrix as defined in Section~\ref{preprocessing}. Next, let $\mathcal{N}_{i,j}(\mathbf{O})$ represent the set of indices of the cells with a \textit{Chebyshev distance}\footnote{The Chebyshev distance, also known as the chessboard distance, between two cells $(i_1,j_1)$ and $(i_2,j_2)$ in a grid is defined as:
\[
D((i_1,j_1),(i_2,j_2))=\max(|i_1-i_2|,|j_1-j_2|).
\]
This metric captures the maximum horizontal, vertical, or diagonal displacement between two cells.
} of 1 from cell $(i,j)\in\mathbf{O}$ (i.e., its nearest neighbors). Using these three elements, the classification outcomes are defined as follows:
\begin{equation}
    \text{Cell Classification: }
\left\{
\begin{array}{l}
\textit{True negatives}\: (tn): \text{ Cells where both } o_{i,j}=0 \text{ and } p_{i,j}=0; \\[10pt]
\textit{True positives}\: (tp): \text{ Cells where both } o_{i,j}=1 \text{ and } p_{i,j}=1; \\[10pt]
\textit{False negatives}\: (fn): \text{ Cells where } o_{i,j}=1 \text{ and } p_{i,j}=0; \\[10pt]
\textit{False positives}: \left\{
\begin{array}{ll}
fp_{nn}, & \text{ if } \exists(i',j')\in\mathcal{N}_{i,j}(\mathbf{O}) \text{ such that } \\
& o_{i',j'}=1, \text{ (neighboring false positive)} \\[10pt]
fp, & \text{ (otherwise, standard false positive)}
\end{array}
\right.
\end{array}
\right.
\end{equation}

Therefore, the modified metrics become:
\begin{equation}
    \text{Recall NN} = \frac{tp + fp_{nn}}{tp + fp_{nn} + fn }, \quad
\text{Precision NN} = \frac{tp + fp_{nn}}{tp + fp_{nn} + fp}.
\end{equation}

These modified metrics (called NN, an acronym for nearest neighbor) are particularly relevant and applicable in our case since having cells of size 0.077 sq. miles implies that a police officer would take approximately 5 minutes (if the cells share an entire border) or 8 minutes (if they only share a corner) walking (or 2.5 and 4 minutes running, respectively) to go from the center of one cell to the center of a neighboring cell, assuming an average walking speed of 5 km/h. This highlights how small our geographic unit of analysis is and underscores how NN performance can be relevant for first responders.

Lastly, to determine the effectiveness of our proposed model, we compare it against three baseline models, chosen to provide a comprehensive evaluation by spanning statistical methods, traditional machine learning techniques, and deep learning approaches: (1) Logistic Regression (LR), a commonly used baseline in statistics and machine learning that assumes linear relationships between the features and the target variable; (2) Random Forest (RF), a traditional machine learning ensemble approach well-regarded for its ability to handle nonlinear relationships; and (3) LSTM, a deep learning model capable of capturing temporal dependencies but unable to account for spatial dependencies.\footnote{We recognize that self-exciting models (see, for reference, \cite{Mohler2011SelfExciting} and \cite{Mohler2014}) were specifically designed to address the challenges of short-term crime forecasting and have demonstrated considerable value in the criminological literature. However, we deliberately chose not to include these models among our baseline comparisons for the present study. This decision is motivated by the fact that a fair and meaningful comparison would require extending the ConvLSTM architecture with attention mechanisms, which better capture the dependencies central to the concept of self-excitation. Such a modification would represent a significant methodological advancement in its own right and is therefore beyond the intended scope of this work.}

\section{Results}\label{results}
\subsection{General performance}\label{general}
\subsubsection{All crimes}
Firstly, we assessed the model's performance against the three chosen baselines---LR, RF, and LSTM---by considering all five crime categories together, to predict whether there will be a crime in the next time window without discriminating by type. On the one hand, this approach has the advantage that it contains more crime incidents, thereby making the dataset relatively more balanced, and stabilizing the learning process. On the other hand, combining crimes with distinct seasonal patterns may introduce further challenges by obscuring individual crime trends.

\begin{figure}[H]
        \includegraphics[width=\textwidth]{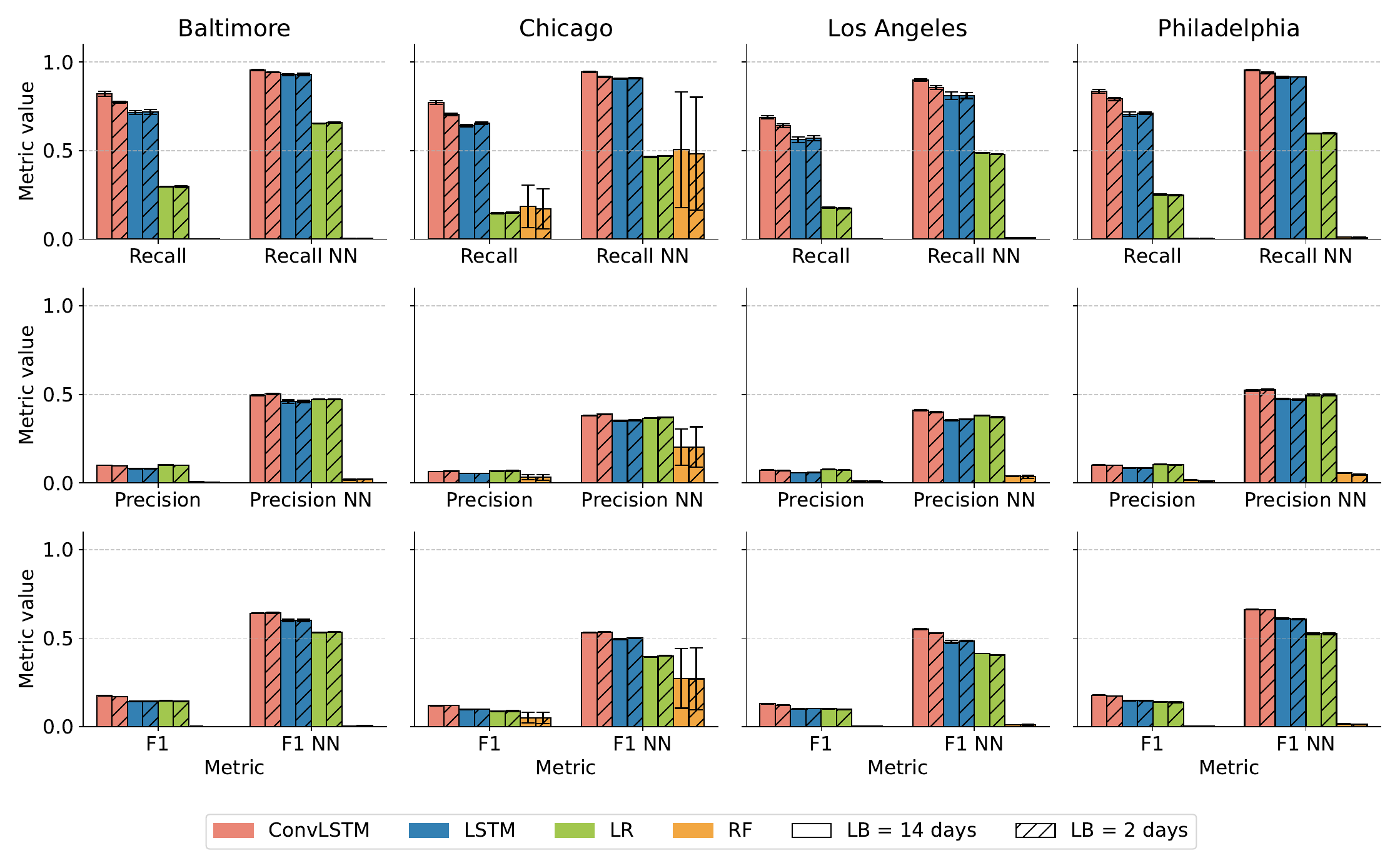}
        \caption{Average performance on the test set for our model and three baselines with and without taking into account the NN cells, considering all five crimes together and two different LB periods. Error bars indicate the standard deviation.}
        \label{all-crimes-together}
\end{figure}

The results in Figure~\ref{all-crimes-together} show that ConvLSTM is the overall best-performing model, which can be clearly seen when comparing the standard and modified F1 scores across the four models. It also achieves the highest standard recall values in all cities, with a notable increase in performance when comparing the results of the modified metrics to the standard metrics.
To exemplify, recall for our ConvLSTM model is always higher than 0.65 across all cities, meaning that we are able to correctly forecast crime presence for 65\% of the cell-time unit observations in which crime actually occurred. Precision is extremely low, being consistently below 10\%, meaning that only one in ten forecasts of crime presence are actually true positives. However, when considering the modified precision metric, performance increases substantially: on average, when we forecast the presence of crime for a given spatial cell, there is approximately a 50\% chance that crime will occur in some of the adjacent spatial cells in that same time unit. Given our fine-grained spatial scale and the highly unbalanced nature of our algorithmic task, this represents an important level of accuracy that does not hinder potential timely response in the short aftermath of a crime.

Regarding the baseline models, all three exhibit consistently lower standard recall values than ConvLSTM. Specifically, LR and RF remain below 0.30, correctly identifying fewer than 30\% of the actual positive instances (cells with at least one crime in a 12-hour block). However, LSTM achieves recall values close to those of ConvLSTM, highlighting how deep learning approaches in general are much more effective compared to less sophisticated (and less computationally costly) models. In terms of precision, LSTM and LR are comparable to ConvLSTM, albeit slightly lower. RF, despite showing similar patterns for Chicago, performs the worst across all metrics, with only marginal improvements for the modified metrics. Although LSTM is the second-best model for recall, its precision is lower than that of LR, highlighting the importance of the spatial patterns captured by ConvLSTM's convolutional operations for improving both recall and precision. It is also worth noting that, despite LSTM's comparable performance, it has approximately ten times more parameters than ConvLSTM, leading to higher computational costs and an increased risk of overfitting, particularly with limited data.

Lastly, the comparison of the LB periods reveals mixed results across models. For ConvLSTM, the 2-day LB period typically shows slightly lower performance compared to the 14-day period, although it achieves higher modified precision in Baltimore, Chicago, and Philadelphia. In contrast, LSTM generally performs better with the 2-day LB period. Meanwhile, LR demonstrates similar performance across both LB periods, while RF seems to benefit from longer LB periods. If we instead focus on the F1 scores, we observe that both LB periods have similar performance, with only Los Angeles showing noticeably higher performance when using the 14-day period for all models except LSTM. A robustness check using 7-day and 1-day LB periods (see \ref{LB7andLB1}) shows that the shorter LB period performs similarly or worse, except for the modified value in Baltimore. These mixed results suggest that the optimal LB period to be used to forecast crime depends both on the chosen algorithm and on the city itself.

\subsubsection{Violent crimes}
\begin{figure}[h]
        \includegraphics[width=1\textwidth]{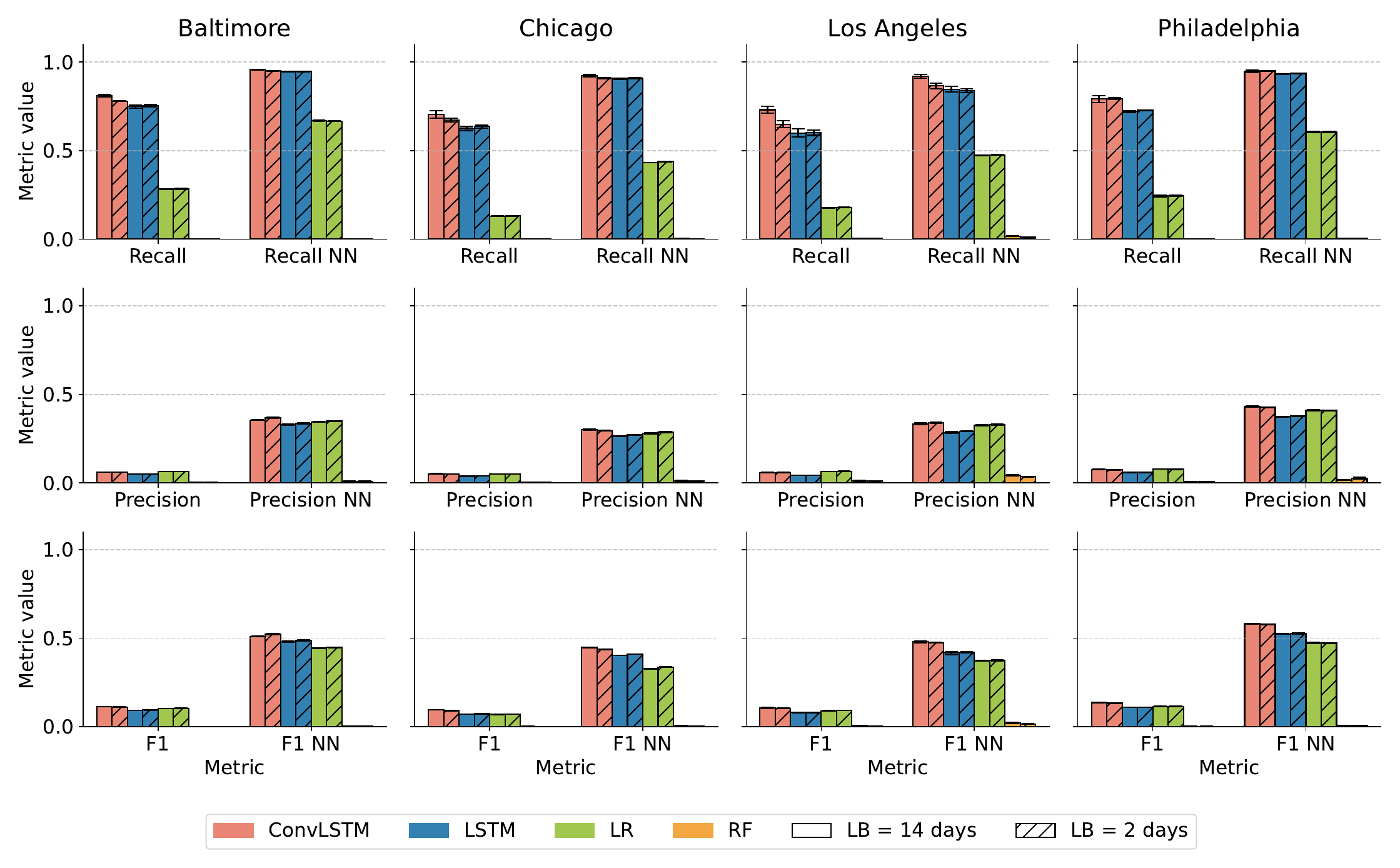}
        \caption{Average performance on the test set for our model and three baselines with and without taking into account the NN cells, considering the violent crimes and two different LB periods. Error bars indicate the standard deviation.}
        \label{fig:violent-crimes}
\end{figure}

Figure~\ref{fig:violent-crimes} presents the results when considering only violent crimes (i.e., assault, homicide, and robbery). Although this dataset contains fewer crime incidents, it remains relatively balanced since assault accounts for approximately half of the crime incidents in each city, as shown in Table~\ref{crime-types}.

Consistent with the previous results, the modified metrics achieve higher performance than the standard metrics across all models and cities, although the magnitude of these improvements varies. Notably, ConvLSTM remains the best-performing model for all cities, with generally higher modified recall values compared to when we used all crimes together. This suggests that violent crimes may exhibit stronger spatial dependencies, hence making the use of modified metrics more beneficial. However, precision decreases across all cities. In general, when comparing the F1 scores, we can conclude that the overall performance decreases relative to using all five crimes together.\par
With respect to the baseline models, LR achieves precision values similar to those of LSTM, slightly exceeding it in all cities for both the standard and modified metrics. Despite this, its recall remains consistently low. In contrast, RF performs poorly across all metrics and remains the worst-performing model, indicating its inability to capture meaningful patterns. Meanwhile, LSTM shows a slight increase in recall, but a decrease in precision compared to its results when considering all crimes combined.

These trends hold true across both LB periods. Nevertheless, we still observe that for ConvLSTM the 2-day LB period performs slightly worse than the 14-day period when using the standard recall and precision, except for the standard recall in Philadelphia. The results are more mixed when we examine the modified metrics instead. However, the F1 score is always slightly higher for the longer LB period, except for the modified value in Baltimore. The same pattern is observed when comparing the 7-day and 1-day LB periods, as shown in \ref{LB7andLB1}. These findings suggest that, when using ConvLSTM, violent crimes may often exhibit patterns that benefit from longer timescales.

\subsubsection{Property crimes}
Lastly, we evaluated the model's performance on property crimes, specifically burglary and MVT. This configuration had the lowest number of crime incidents, resulting in a highly unbalanced dataset. Nevertheless,  property crimes have been shown to be relatively easy to model due to their seasonal trends, dependency on neighborhood characteristics, and correlation with socio-economic factors \citep{linning2017crime}.\par

\begin{figure}[H]
        \includegraphics[width=1\textwidth]{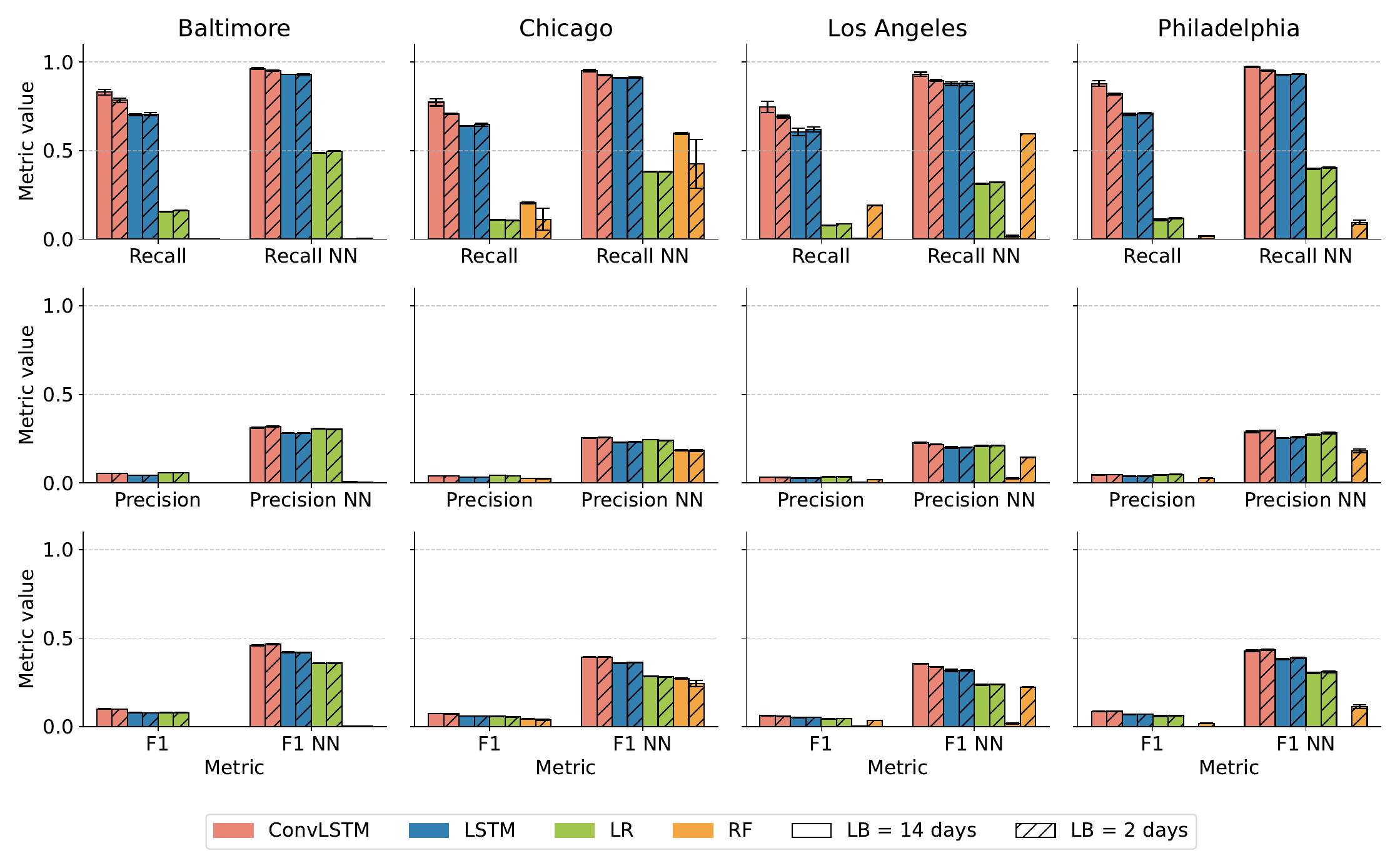}
        \caption{Average performance on the test set for our model and three baselines with and without taking into account the NN cells, considering the property crimes and two different LB periods. Error bars indicate the standard deviation.}
        \label{property-crimes}
\end{figure}

The results, presented in Figure~\ref{property-crimes}, indicate a further decline in overall performance compared to violent crimes, which becomes clear when looking at the F1 score values. For the baseline models, we observe that LR consistently achieves again slightly higher precision than LSTM across all cities. However, its standard recall remains very low (below 0.2). RF continues to be the worst-performing model, generally exhibiting very low recall and precision, although its performance improves in Chicago, Los Angeles, and Philadelphia, compared to its results for violent crimes. LSTM still achieves the highest recall among the baselines but shows slightly worse overall performance (as indicated by the F1 score) compared to its application to all crimes or violent crimes alone. Finally, the ConvLSTM model remains the best-performing model, maintaining strong recall values but showing a decrease in precision---a trend consistent with the baselines.

Therefore, these results indicate that the high data sparsity did not allow the model to learn the patterns in property crime properly when compared to models using all crimes together or only violent crimes.

Lastly, when comparing the values obtained for the two LB periods, we observe a general tendency for the 2-day LB period to achieve slightly better precision than the 14-day period, but lower recall—a pattern also observed for the other two crime aggregations. However, in this case, despite the F1 score indicating that the overall performance is very similar between the two LB periods, the shorter LB performs slightly better when using the modified metric for Baltimore and Philadelphia. This trend is also observed in Chicago, in addition to the two aforementioned cities, when comparing 7-day and 1-day LB periods (see \ref{LB7andLB1}). These results suggest that the patterns driving property crimes may generally operate over shorter timescales compared to those of violent crimes, although there is some variation depending on the city.

\subsection{Effect of mobility features}\label{mobility_features}
After evaluating the models' general performance, we sought to assess the impact of including mobility features in our deep learning framework. Thus, we retrained the best-performing model, ConvLSTM, using various feature sets: CMS, CM, CS, and C, where C represents crime features, M represents mobility features, and S represents sociodemographic features. This retraining was conducted using the best configuration alone, hence including all five crime types.\par

\begin{figure}[H]
    \centering
        \includegraphics[width=0.94\textwidth]{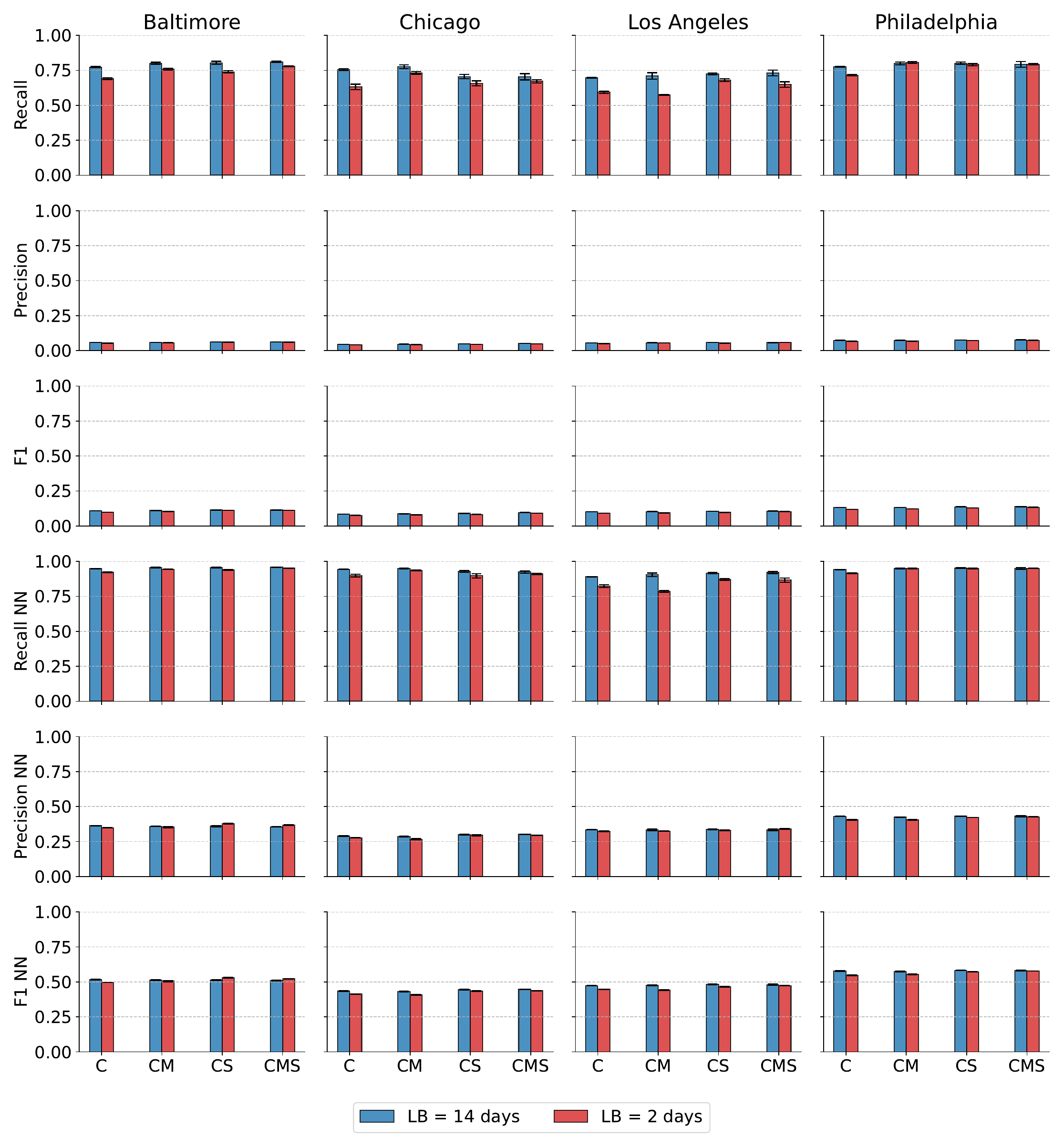}
        \caption{Average performance of the ConvLSTM model on the test set, trained using four different feature sets and two different LB periods. Error bars indicate the standard deviation.}
        \label{mobility-effects}
\end{figure}

Figure~\ref{mobility-effects} shows the results of the average performance and standard deviation for each of the four cities. If we focus on the results using an LB period of 14 days, the first observation is the considerable variation in standard recall within each city, depending on the feature set used for training. Looking more closely, we see that Chicago and Los Angeles have the lowest performance when focusing on the standard metrics, with the recall always below 0.76 and the precision always below 0.08. The same trend is observed when looking at the modified metrics.

Overall, based on both the standard and modified recall and precision values, the setup incorporating all three feature types consistently performs best, as expected. Specifically, among the 24 configurations seen in Figure~\ref{mobility-effects}, the C setup achieves the highest performance in 8.3\% of cases, the CM setup in 8.3\%, CS in 29.2\%, and CMS in 54.2\%. Therefore, the results indicate that incorporating additional feature types alongside crime data improves the model's ability to forecast effectively.

Conversely, when analyzing the results with an LB period of 2 days, performance consistently decreases slightly when using the standard metrics. However, Chicago and Los Angeles remain the cities with the lowest performance, and the CMS configuration again emerges as the best overall. In this case, the C setup does not achieve the highest performance in any instance, while the CM setup outperforms others in 12.5\% of cases, CS in 29.2\%, and CMS in 58.3\%. Thus, while shorter LB periods lead to a slight reduction in overall performance, the advantage of incorporating additional feature types alongside crime data becomes more evident.

Afterward, to further understand the impact of mobility features on predictive performance, we calculated the percentage difference between the results of CS and CMS (to assess the effect of incorporating mobility features) and CM and CMS (to evaluate the effect of sociodemographic features) for both LB periods. As shown in Table~\ref{percent-diff}, the results generally indicate a slight decrease in recall when a third feature set is added---both when going from CM to CMS and from CS to CMS---resulting in a negative percentage point difference. In contrast, precision increases in all configurations, with the exception of its modified value for Chicago when using a 14-day LB and Baltimore when using the 2-day LB. Another noteworthy observation is that the LB period of 2 days always shows a greater percentage point increase in overall performance (F1 score) compared to the 14-day period, suggesting that shorter LB periods benefit more from the incorporation of mobility features. Additionally, comparing the F1 scores between CMS-CM and CMS-CS, we can see that sociodemographic features have a greater impact than mobility features in increasing performance; however, both contribute to enhancing the model's ability to forecast crime, with the best performance achieved when they are used together.

\begin{table}[H]
    \fontsize{11pt}{11pt}\selectfont
    \centering
    \begin{tabular}{ccccccccccc}
    \toprule
     &  & \multicolumn{4}{c}{CMS - CM} && \multicolumn{4}{c}{CMS - CS} \\
    \cmidrule(r){3-6}
    \cmidrule(r){8-11}
    LB & Metric & Bal & Chi & Las & Phi && Bal & Chi & Las & Phi \\
    \midrule
    \multirow{6}{*}{\rotatebox{90}{14 days}}
        & Rec   & \textcolor{red}{-3.513} & \textcolor{red}{-4.743} & \textcolor{red}{-0.864} & \textcolor{red}{-1.036} && \textcolor{red}{-2.157} & \textcolor{ForestGreen}{0.427} & \textcolor{red}{-4.920} & \textcolor{red}{-0.270} \\
        & Prec   & \textcolor{ForestGreen}{8.381} & \textcolor{ForestGreen}{5.606} & \textcolor{ForestGreen}{4.372} & \textcolor{ForestGreen}{4.280} && \textcolor{ForestGreen}{2.689} & \textcolor{ForestGreen}{2.506} & \textcolor{ForestGreen}{3.197} & \textcolor{ForestGreen}{1.630} \\
        & F1    & \textcolor{ForestGreen}{6.988} & \textcolor{ForestGreen}{4.372} & \textcolor{ForestGreen}{4.088} & \textcolor{ForestGreen}{3.766} && \textcolor{ForestGreen}{2.077} & \textcolor{ForestGreen}{2.346} & \textcolor{ForestGreen}{2.185} & \textcolor{ForestGreen}{1.453} \\
        & Rec NN   & \textcolor{red}{-0.986} & \textcolor{red}{-1.168} & \textcolor{red}{-0.314} & \textcolor{red}{-0.206} && \textcolor{red}{-0.594} & \textcolor{ForestGreen}{0.068} & \textcolor{red}{-1.918} & \textcolor{red}{-0.161} \\
        & Prec NN  & \textcolor{ForestGreen}{3.266} & \textcolor{ForestGreen}{2.983} & \textcolor{ForestGreen}{1.309} & \textcolor{ForestGreen}{2.076} && \textcolor{ForestGreen}{0.727} & \textcolor{red}{-0.102} & \textcolor{ForestGreen}{1.758} & \textcolor{ForestGreen}{0.525} \\
        & F1 NN   & \textcolor{ForestGreen}{1.824} & \textcolor{ForestGreen}{1.737} & \textcolor{ForestGreen}{1.065} & \textcolor{ForestGreen}{1.339} && \textcolor{ForestGreen}{0.236} & \textcolor{red}{-0.099} & \textcolor{ForestGreen}{0.560} & \textcolor{ForestGreen}{0.283} \\
    \midrule
    \multirow{6}{*}{\rotatebox{90}{2 days}}
        & Rec   & \textcolor{red}{-3.185} & \textcolor{red}{-9.972} & \textcolor{ForestGreen}{4.853} & \textcolor{red}{-2.698} && \textcolor{ForestGreen}{3.362} & \textcolor{red}{-0.923} & \textcolor{red}{-2.775} & \textcolor{ForestGreen}{0.123} \\
        & Prec   & \textcolor{ForestGreen}{8.640} & \textcolor{ForestGreen}{14.239} & \textcolor{ForestGreen}{5.256} & \textcolor{ForestGreen}{6.390} && \textcolor{ForestGreen}{2.073} & \textcolor{ForestGreen}{6.917} & \textcolor{ForestGreen}{5.612} & \textcolor{ForestGreen}{3.069} \\
        & F1    & \textcolor{ForestGreen}{7.164} & \textcolor{ForestGreen}{11.945} & \textcolor{ForestGreen}{5.542} & \textcolor{ForestGreen}{5.175} && \textcolor{ForestGreen}{2.627} & \textcolor{ForestGreen}{6.737} & \textcolor{ForestGreen}{5.323} & \textcolor{ForestGreen}{2.839} \\
        & Rec NN   & \textcolor{red}{-0.668} & \textcolor{red}{-2.875} & \textcolor{ForestGreen}{2.561} & \textcolor{red}{-0.606} && \textcolor{ForestGreen}{1.102} & \textcolor{ForestGreen}{0.040} & \textcolor{red}{-0.377} & \textcolor{ForestGreen}{0.348} \\
        & Prec NN  & \textcolor{ForestGreen}{4.567} & \textcolor{ForestGreen}{7.510} & \textcolor{ForestGreen}{1.929} & \textcolor{ForestGreen}{3.906} && \textcolor{red}{-0.243} & \textcolor{ForestGreen}{0.533} & \textcolor{ForestGreen}{1.615} & \textcolor{ForestGreen}{1.353} \\
        & F1 NN     & \textcolor{ForestGreen}{2.695} & \textcolor{ForestGreen}{4.482} & \textcolor{ForestGreen}{2.314} & \textcolor{ForestGreen}{2.333} && \textcolor{ForestGreen}{0.404} & \textcolor{ForestGreen}{0.555} & \textcolor{ForestGreen}{1.375} & \textcolor{ForestGreen}{1.105} \\
    \bottomrule
    \end{tabular}
    \caption{Percentage difference in model performance between the CS and CMS configurations (to assess the effect of M), and between the CM and CMS configurations (to assess the effect of S), for each city and using LB periods of 14 and 2 days. Positive values (in green) indicate improved performance, while negative values (in red) indicate a decrease.}
    \label{percent-diff}
\end{table}
Lastly, as a robustness check and to investigate further how long-term versus short-term patterns influence crime prediction, we repeated the analysis in this subsection using two additional LB periods, 7 days and 1 day. The results, which can be found in \ref{features_extra}, also show lower overall performance for the shorter LB period and an improvement (or very small decrease) in performance when incorporating a third feature set, regardless of whether it is the mobility or the sociodemographic feature. We also observe that the shorter LB period still has a greater percentage point increase in overall performance (F1 score). However, the relative advantage between 1 and 2 days, as well as between 7 and 14 days, is not clear from this analysis.

\section{Discussion and conclusions}\label{discussion}
Extensive research has explored the spatiotemporal concentration of crime and how resulting patterns can be exploited for crime forecasting \citep{Johnson2010, Kounadi2020review}. In recent years, thanks also to the vast availability of data, studies in this field have started to explore more advanced computational methods (e.g., machine and deep learning), to investigate a range of questions \citep{campedelli2022machine}. Moreover, motivated by several theories highlighting the correlation between human mobility and the dynamics of crime, some studies started incorporating mobility data in combination with more traditional data \citep{Graif2021, MobilityDLWu, DeNadai2020, kadar2018}. Despite this, many questions remain unanswered as current scholarship is characterized by relatively coarse spatial and temporal granularities, very few years of historical data, and/or focus on a single city. This paper aimed to advance the literature on crime forecasting and spatial criminology by offering four main contributions.

Firstly, we used five years of human mobility data, in combination with crime and sociodemographic data from four U.S.\ cities, to assess the extent to which these features could enhance the prediction of the spatiotemporal occurrence of crime. Secondly, we defined a small geographic unit of analysis of 0.077 sq. miles and a fine temporal granularity of 12 hours in order to assess whether this fine-grained spatiotemporal unit of analysis allows us to meaningfully forecast crime in four different urban contexts in the U.S. To evaluate this, we designed a deep learning model using ConvLSTM layers and compared its performance to three baseline models to determine whether the performance improvement of using an advanced computational method compensated for the complexity involved in building it. Lastly, we compared the results from two different LB periods to assess the impact of longer versus shorter time windows on predictive performance.

The results led to several key findings. We first demonstrated that ConvLSTM consistently outperformed the three baseline models, particularly in recall, but struggled with precision, especially when using standard metrics. LSTM showed the closest performance in both recall and precision, though it remained slightly below ConvLSTM. LR, while competitive in precision, suffered from low recall, and RF failed to capture meaningful patterns across all data configurations. Therefore, on the one hand, the effort involved in building such an elaborate model was worthwhile, as the other models either struggled with high-dimensional data or, as in the case of LSTM, had to be highly complex to achieve comparable performance. Moreover, it should be noted that ConvLSTM’s high recall values suggest it can predict most instances where at least one crime is likely to occur, minimizing false negatives as intended. However, the low precision results highlight how the extremely unbalanced nature of our forecasting tasks---a byproduct of the fine-grained spatial and temporal scales of the study---makes this approach highly imperfect in terms of reducing false positives.

Concerning the crime types, we found that our model performed best when using all crime types together, rather than separating them into violent and property crimes. This outcome was expected since this configuration results in a more balanced dataset, hence leading to more stable training. Additionally, the comparison across four different LB periods indicated that violent crimes tend to yield higher overall predictive performance with longer LB periods, whereas property crimes appear to benefit from shorter ones.

When evaluating our model on different sets of features, we were able to determine that the overall best performance was achieved when using all three feature sets together (i.e., crime, mobility, and sociodemographic data), while the worst performance occurred when only historical crime data were used. This suggests that including additional data sources, in combination with crime data, allows the algorithm to learn more effectively the patterns leading to crime occurrence. Moreover, the percentage difference between CS and CMS indicates that mobility data improve performance, especially when using shorter LB periods, stressing how mobility captures aspects of an urban context that transcend its static socio-economic ecology. \par
These findings not only contribute to those strands of criminological literature concerned with spatiotemporal crime forecasting and the investigation of the link between mobility and crime, but are also relevant to policymakers. In fact, they contribute to the ongoing debate on the challenges associated with the use of computational approaches by law enforcement agencies \citep{Lum2016, Brayne2017, Brantingham2018, Meijer2019, Berk2021}. Although offering an exhaustive, systematic analysis of the limits of deep learning for crime forecasting is beyond the scope of this work, our results underscore that not even extremely sophisticated algorithms can reach high performance in both recall and precision when focusing on fine-grained spatial and temporal scales. This finding---detected for all cities and crime types---bears important implications. Relying on the predictions of our models, for instance, would certainly allow first responders and law enforcement to predict practically all actual offenses, but at the cost of wrongfully targeting certain micro-areas due to high rates of false positives, possibly leading to unintended consequences. While we show that this issue is largely diminished when considering the modified nearest neighbor metrics, the low baseline precision scores speak to the limits that computational models still face when dealing with such a sparse, unbalanced, and volatile phenomenon as crime.

\subsection{Limitations}\label{limitations}
This work, despite its promising results, is not without limitations. First, the use of a grid-based approach, while being very common in the literature (see, among others, \cite{Rummens2017, Zhao2017, Mohler2014} ) imposes an arbitrary division on our space, and it has been shown that network-based models can outperform grid-based models for crime prediction \citep{gridsbad}. Furthermore, an arbitrary division of urban space may also pose a problem, for instance, in terms of administrative or police precinct boundaries, since two points in the same spatial cell might fall under the jurisdiction of two different police precincts. However, a grid-based approach was mandatory given our design focused on the adaptation of computer vision models to crime forecasting.

A second potential limitation is that the crime data we used suffered from location obfuscation due to privacy concerns, which limits how finely we can define our spatial unit of analysis. However, our cell size is still considerably smaller than the typical block size in American cities (see \cite{major2015invention} for some estimates), including those analyzed in this research. Additionally, given the low precision obtained with our current grid cell size, we are concerned that further reducing the cell size would exacerbate the problem of false positives. Therefore, we argue that obfuscation does not constitute a major issue in our current design, as it is not as micro in scale.

Finally, arguably the main limitation of this study is that it does not rely on individual trajectories to capture mobility dynamics in the cities under analysis. More nuanced mobility features, such as mobility flows, have been shown to further improve forecasting performance \citep{mobilityflows} and would certainly have offered a richer set of opportunities to study how crime concentrates and varies over time along with variation in human movements across different parts of the same urban context. Unfortunately, we were unable to obtain such information due to the limited availability of mobility data for crime research.

\subsection{Ethical Implications}\label{ethics}
This study adheres to ethical research practices, particularly regarding the use of mobility and crime data. The data used in this study do not contain individual-level information; instead, they represent aggregate mobility information at the premise level, further aggregated at the spatial cell level. This fully ensures the privacy and anonymity of individuals while allowing us to study broader patterns of human activity and their relationship with crime.

We also acknowledge the inherent limitations in the reporting of crime data. Certain crimes may be underreported in specific areas of a city due to socioeconomic, cultural, or institutional factors, potentially introducing biases into the analysis. Nonetheless, we have sought to design our framework focusing on those crimes that are less likely to suffer from this issue, as well as crimes that are less prone to be the subject of specific policing decision-making processes, such as those concerning drug-related offenses.

Additionally, our model's statistical results, particularly its low precision, raise significant concerns that could have ethical implications if deployed in real-world scenarios. Relying on these results to allocate police resources risks unjustified overpolicing certain areas, as false positives are pervasive across cities and crime types, possibly exacerbating existing inequalities or disproportionately impacting specific communities.

We therefore emphasize that the results of this study are intended to advance the theoretical understanding of crime and to contribute to the development of methodological approaches in spatiotemporal crime forecasting, rather than to serve as direct recommendations for enforcement actions. Future research should focus on addressing these ethical concerns, improving model precision, and incorporating diverse stakeholder perspectives to ensure fair and equitable algorithmic design.

\section*{Acknowledgments}
Anonymized for submission. We thank Laia Albors Zumel for comments and suggestions regarding earlier versions of this manuscript and Giuseppe Veltri for partial financial support in the first phase of this research.

\section*{Author Contributions}
This work is the result of the joint efforts by all authors, A.A.Z., M.T., and G.M.C. A.A.Z., M.T., and G.M.C. jointly contributed to design the study and to its methodological setup, A.A.Z. performed the coding and modeling part, A.A.Z. and G.M.C. both contributed to the analysis of the results, A.A.Z. wrote the first draft of the paper and M.T. and G.M.C. reviewed and refined it. M.T. contributed funding to acquire the mobility data. G.M.C. supervised the project.

\section*{Competing Interests}
The authors declare that they have no financial or non-financial interests, either directly or indirectly, related to the content of this work.

\section*{Data and Code Availability}
The code and links to the raw data used in this study have been deposited in Zenodo and are accessible via DOI: 10.5281/zenodo.16042193.

\bibliographystyle{ieeetr}
\bibliography{sn-bibliography.bbl}

\appendix
\renewcommand{\thesection}{Appendix \Alph{section}}
\section{Crime distribution}\label{crime_distrib}
\begin{figure}[H]
        \centering
        \includegraphics[width=1\textwidth]{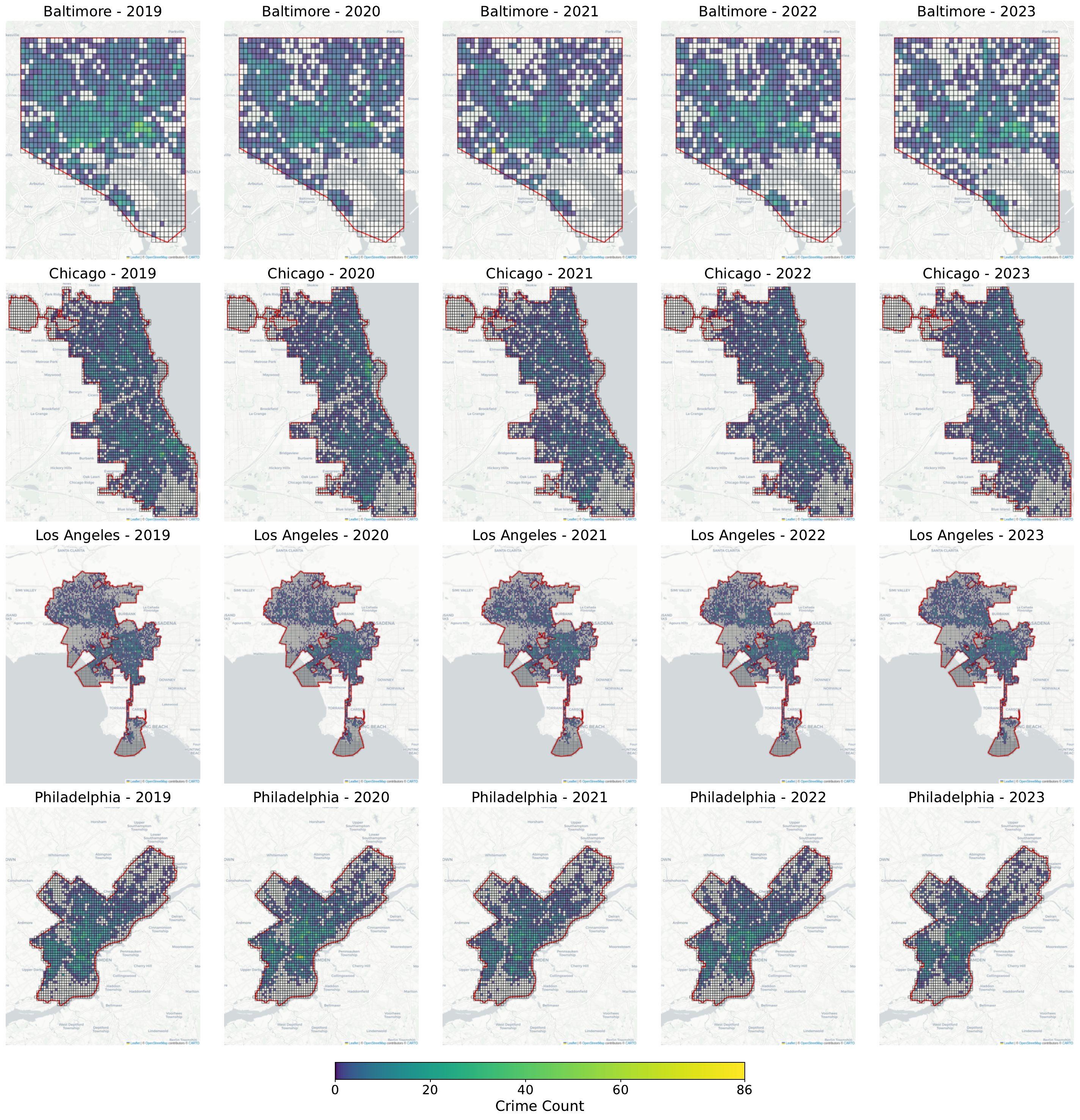}
        \caption{Map of the crime distribution for each of the four U.S.\ cities for each of the years between 2019 and 2023.}
        \label{fig:crime-distrib}
\end{figure}
\section{People-to-POI ratio}\label{poi_ratio}
\begin{figure}[H]
    \centering\includegraphics[width=0.735\textwidth]{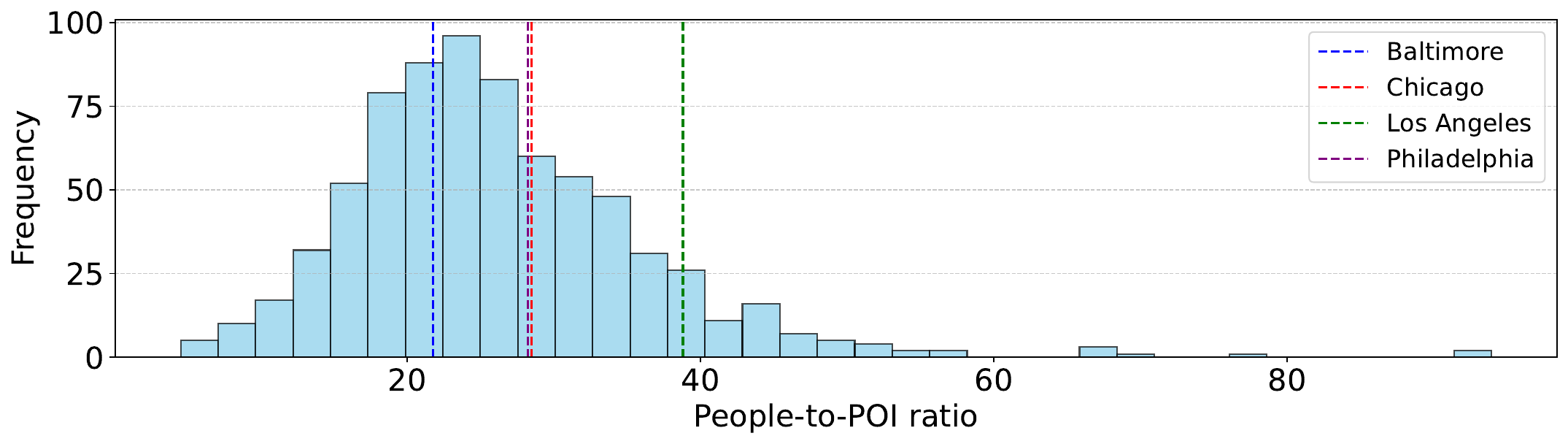}
        \caption{Histogram of the distribution of the people-to-POI ratio for the 735 cities in the U.S., with vertical lines indicating the position of our four cities within this distribution.}
        \label{fig:crime-ratio}
\end{figure}
\section{POI distribution}\label{poi_distrib}
\begin{figure}[H]
    \centering
    \begin{subfigure}{0.47\textwidth}
        \centering
        \includegraphics[width=\textwidth]{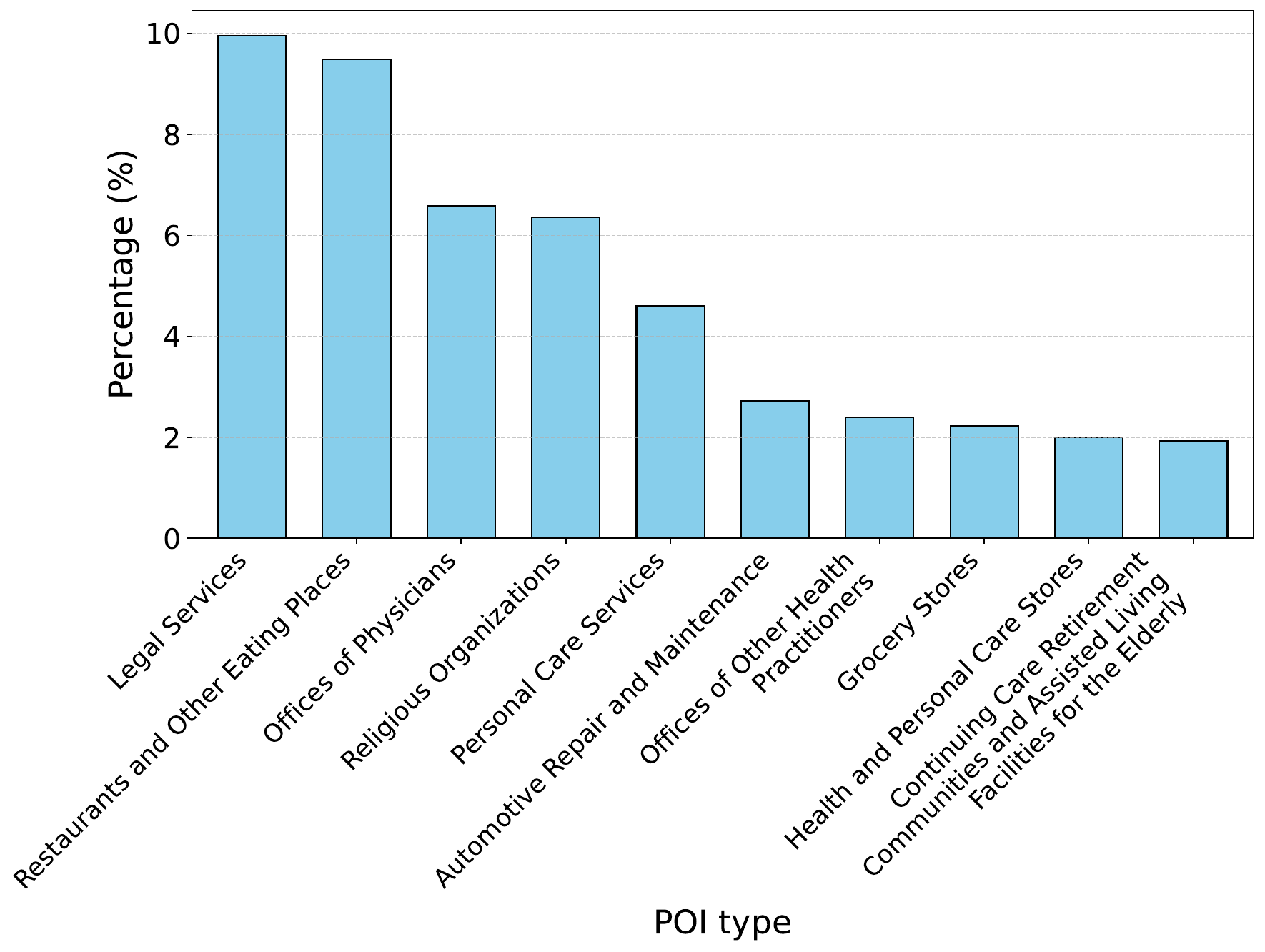}
        \caption{Baltimore.}
    \end{subfigure}%
    \begin{subfigure}{0.47\textwidth}
        \centering
        \includegraphics[width=\textwidth]{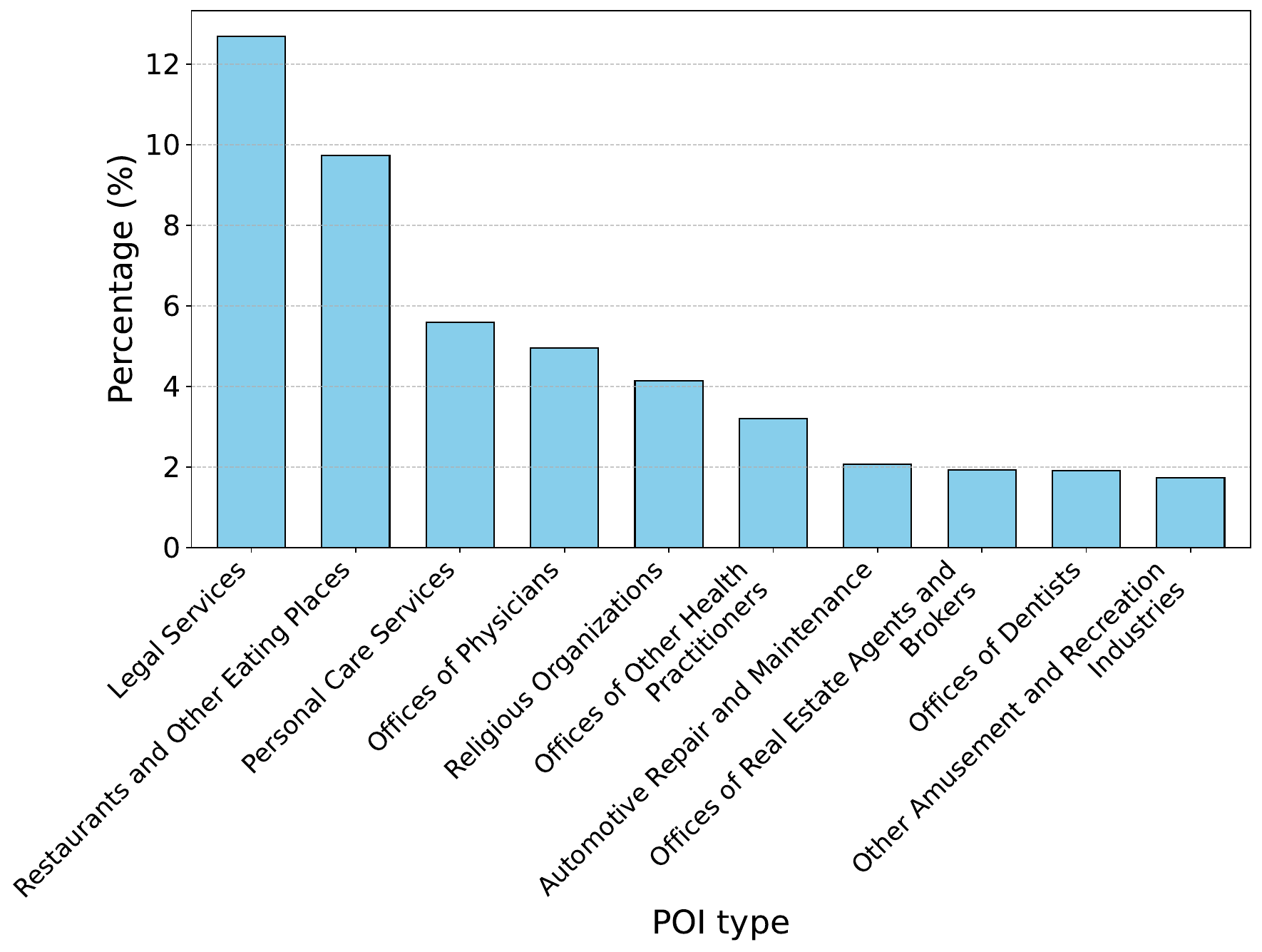}
        \caption{Chicago.}
    \end{subfigure} \\
    \begin{subfigure}{0.47\textwidth}
        \centering
        \includegraphics[width=\textwidth]{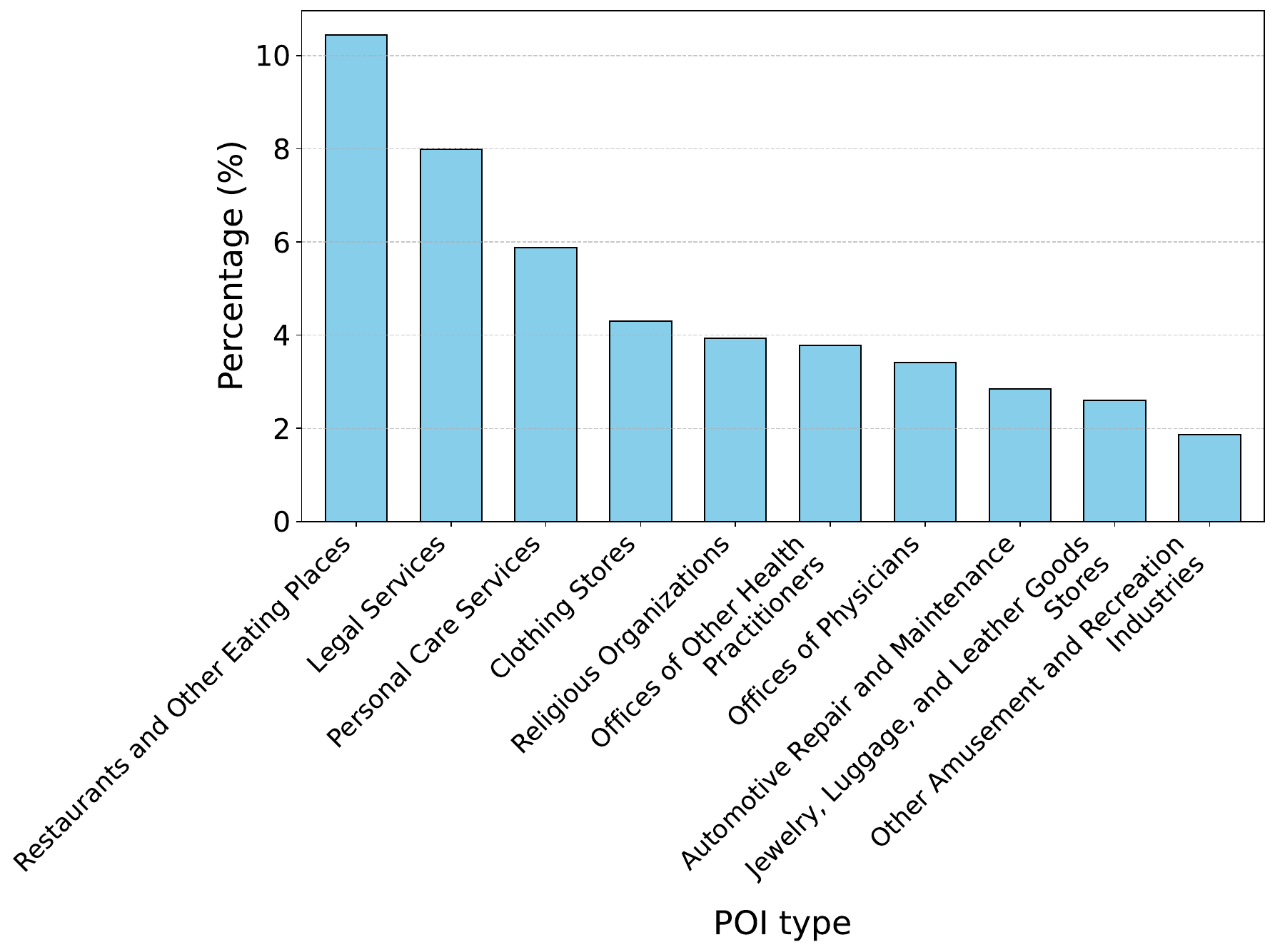}
        \caption{Los Angeles.}
    \end{subfigure}%
    \begin{subfigure}{0.47\textwidth}
        \centering\includegraphics[width=\textwidth]{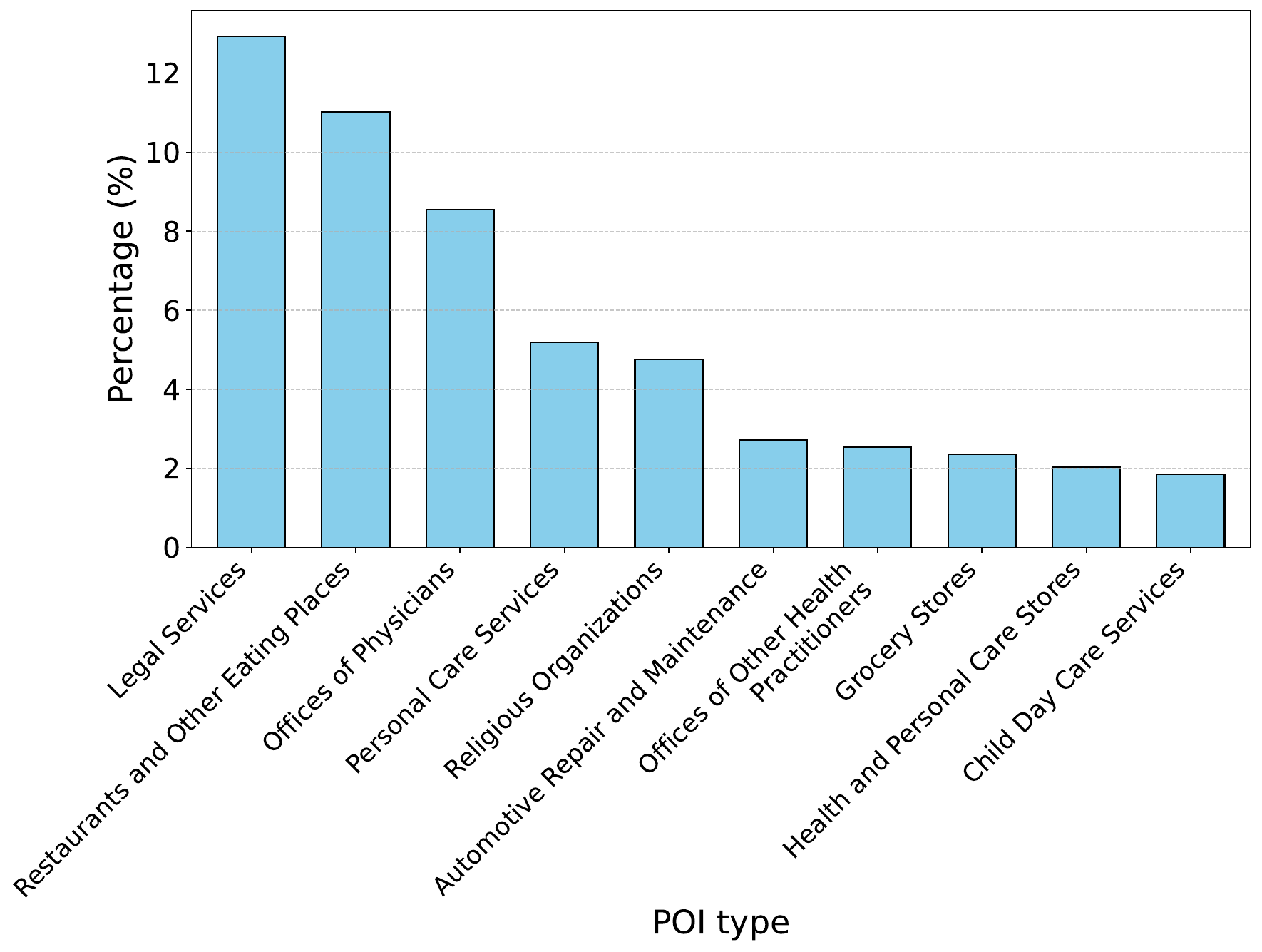}
        \caption{Philadelphia.}
    \end{subfigure}
    \caption{Top 10 POI types by percentage for each of the four U.S.\ cities between 2019 and 2023.}
    \label{fig:poi-type}
\end{figure}

\section{POI Categories}\label{secA}
\begin{enumerate}
    \item \textbf{Utilities and Construction:} ``Building Material and Supplies Dealers'', ``Foundation, Structure, and Building Exterior Contractors'', ``Building Equipment Contractors'', ``Building Finishing Contractors'', ``Other Specialty Trade Contractors'', ``Utility System Construction'', and ``Electric Power Generation, Transmission and Distribution''.
    \item \textbf{Manufacturing}:
            ``Bakeries and Tortilla Manufacturing'', ``Beverage Manufacturing'',
``Other Miscellaneous Manufacturing'',
``Glass and Glass Product Manufacturing'',
``Other Wood Product Manufacturing'',
``Metalworking Machinery Manufacturing'',
``Steel Product Manufacturing from Purchased Steel'',
``Greenhouse, Nursery, and Floriculture Production'', and
``Printing and Related Support Activities''.
    \item \textbf{Retail and Wholesale trade}:
            ``Shoe Stores'',
``Grocery Stores'',
``Clothing Stores'',
``Electronics and Appliance Stores'',
``Health and Personal Care Stores'',
``Gasoline Stations'',
``Used Merchandise Stores'',
``Automotive Parts, Accessories, and Tire Stores'',
``Jewelry, Luggage, and Leather Goods Stores'',
``Beer, Wine, and Liquor Stores'',
``Specialty Food Stores'',
``Home Furnishings Stores'',
``Other Motor Vehicle Dealers'',
``Office Supplies, Stationery, and Gift Stores'',
``General Merchandise Stores, including Warehouse Clubs and Supercenters'',
``Department Stores'',
``Book Stores and News Dealers'',
``Automobile Dealers'',
``Furniture Stores'',
``Lawn and Garden Equipment and Supplies Stores'',
``Hardware, and Plumbing and Heating Equipment and Supplies Merchant Wholesalers'',
``Machinery, Equipment, and Supplies Merchant Wholesalers'',
``Drugs and Druggists' Sundries Merchant Wholesalers'',
``Chemical and Allied Products Merchant Wholesalers'',
``Petroleum and Petroleum Products Merchant Wholesalers'',
``Motor Vehicle and Motor Vehicle Parts and Supplies Merchant Wholesalers'',
``Miscellaneous Durable Goods Merchant Wholesalers'',
``Grocery and Related Product Merchant Wholesalers'',
``Household Appliances and Electrical and Electronic Goods Merchant Wholesalers'',
``Lumber and Other Construction Materials Merchant Wholesalers'',
``Direct Selling Establishments'',
``Other Miscellaneous Store Retailers'',
``Professional and Commercial Equipment and Supplies Merchant Wholesalers'',
and ``Sporting Goods, Hobby, and Musical Instrument Stores''.
    \item \textbf{Transportation and warehousing}:
            ``Support Activities for Air Transportation'',
``Specialized Freight Trucking'',
``Rail Transportation'',
``Taxi and Limousine Service'',
``Other Transit and Ground Passenger Transportation'',
``Transit and Ground Passenger Transportation'',
``Scenic and Sightseeing Transportation'',
``Support Activities for Road Transportation'',
``Freight Transportation Arrangement'',
``Support Activities for Water Transportation'',
``Warehousing and Storage'',
``Automotive Equipment Rental and Leasing'',
``Commercial and Industrial Machinery and Equipment Rental and Leasing'',
``Interurban and Rural Bus Transportation'',
and ``Postal Service''.
    \item \textbf{Business and Professional Services}:
           ``Investigation and Security Services'',
``Activities Related to Credit Intermediation'',
``Offices of Real Estate Agents and Brokers'',
``Other Professional, Scientific, and Technical Services'',
``Accounting, Tax Preparation, Bookkeeping, and Payroll Services'',
``Management, Scientific, and Technical Consulting Services'',
``Advertising, Public Relations, and Related Services'',
``Legal Services'',
``Data Processing, Hosting, and Related Services'',
``Architectural, Engineering, and Related Services'',
``Specialized Design Services'',
``Business Support Services'',
``Employment Services'',
``Travel Arrangement and Reservation Services'',
``Freight Transportation Arrangement'',
``Management of Companies and Enterprises'',
``Activities Related to Real Estate'',
``Agencies, Brokerages, and Other Insurance Related Activities'',
``Cable and Other Subscription Programming'',
``Consumer Goods Rental'',
``Depository Credit Intermediation'',
``General Rental Centers'',
``Insurance Carriers'',
``Lessors of Real Estate'',
``Motion Picture and Video Industries'',
``Nondepository Credit Intermediation'',
``Other Financial Investment Activities'',
``Other Information Services'',
``Radio and Television Broadcasting'',
``Sound Recording Industries'',
and ``Wired and Wireless Telecommunications Carriers''.
    \item \textbf{Educational services}:
           ``Elementary and Secondary Schools'',
``Colleges, Universities, and Professional Schools'',
``Junior Colleges'',
``Technical and Trade Schools'',
``Other Schools and Instruction'',
and ``Educational Support Services''.
    \item \textbf{Health care and social assistance}:
    ``Offices of Physicians'',
``Offices of Other Health Practitioners'',
``Specialty (except Psychiatric and Substance Abuse) Hospitals'',
``General Medical and Surgical Hospitals'',
``Psychiatric and Substance Abuse Hospitals'',
``Outpatient Care Centers'',
``Nursing Care Facilities (Skilled Nursing Facilities)'',
``Continuing Care Retirement Communities and Assisted Living Facilities for the Elderly'',
``Home Health Care Services'',
``Individual and Family Services'',
``Community Food and Housing, and Emergency and Other Relief Services'',
``Residential Intellectual and Developmental Disability, Mental Health, and Substance Abuse Facilities'',
``Child Day Care Services'',
``Medical and Diagnostic Laboratories'',
``Nursing and Residential Care Facilities'',
``Offices of Dentists'',
and ``Other Ambulatory Health Care Services''.
    \item \textbf{Arts, entertainment, and recreation}:
            ``Museums, Historical Sites, and Similar Institutions'',
``Amusement Parks and Arcades'',
``Spectator Sports'',
``Other Amusement and Recreation Industries'',
``Performing Arts Companies'',
``Promoters of Performing Arts, Sports, and Similar Events'',
``Social Advocacy Organizations'',
``Civic and Social Organizations'',
and ``Gambling Industries''.
    \item \textbf{Accommodation and food services}:
            ``Traveler Accommodation'',
``Special Food Services'',
``Restaurants and Other Eating Places'',
``Drinking Places (Alcoholic Beverages)'',
and ``RV (Recreational Vehicle) Parks and Recreational Camps''.
    \item \textbf{Public administration}:
           ``Justice, Public Order, and Safety Activities'',
``Administration of Economic Programs'',
``Administration of Human Resource Programs'',
and ``National Security and International Affairs''.
    \item \textbf{Other services}:
           ``Florists'',
``Other Personal Services'',
``Religious Organizations'',
``Personal and Household Goods Repair and Maintenance'',
``Drycleaning and Laundry Services'',
``Death Care Services'',
``Personal Care Services'',
``Social Assistance'',
``Grantmaking and Giving Services'',
``Couriers and Express Delivery Services'',
``Waste Management and Remediation Services'',
``Remediation and Other Waste Management Services'',
``Services to Buildings and Dwellings'',
``Waste Treatment and Disposal'',
``Waste Collection'',
``Automotive Repair and Maintenance'',
and ``Electronic and Precision Equipment Repair and Maintenance''.
\end{enumerate}

\section{Sociodemographic Variables}\label{secB}
\textbf{Gender}:
\begin{enumerate}
\setlength{\itemindent}{1.25em}
    \item Percentage of females.
\end{enumerate}
\textbf{Age:}
\begin{enumerate}[start=2,itemsep=0.1em]
\setlength{\itemindent}{1.25em}
    \item Median age of males.
    \item Median age of females.
\end{enumerate}
\textbf{Race/ethnicity:}
\begin{enumerate}[start=4,itemsep=0.1em]
\setlength{\itemindent}{1.25em}
    \item Percentage of white alone.
    \item Percentage of black or African American.
    \item Percentage of American Indian and Alaskan native.
    \item Percentage of Asian alone.
    \item Percentage of native Hawaiian and other.
    \item Percentage of some other race alone.
\end{enumerate}
\textbf{Employment (population over 16 years old):}
\begin{enumerate}[start=10,itemsep=0.1em]
\setlength{\itemindent}{1.25em}
    \item Percentage employed.
    \item Percentage unemployed.
    \item Percentage in the armed forces.
    \item Percentage not in the labor force.
\end{enumerate}
\textbf{Income:}
\begin{enumerate}[start=14]
\setlength{\itemindent}{1.25em}
    \item Median household income in the past 12 months.
\end{enumerate}
\textbf{Education (population over 25 years old):}
\begin{enumerate}[start=15,itemsep=0.1em]
\setlength{\itemindent}{1.25em}
    \item Percentage with no schooling.
    \item Percentage with a regular high school diploma.
    \item Percentage with a Bachelor's degree.
    \item Percentage with a professional school degree.
\end{enumerate}
\textbf{Marital status:}
\begin{enumerate}[start=19,itemsep=0.1em]
\setlength{\itemindent}{1.25em}
    \item Percentage of males never married.
    \item Percentage of males currently married.
    \item Percentage of males widowed.
    \item Percentage of males divorced.
    \item Percentage of females never married.
    \item Percentage of females currently married.
    \item Percentage of females widowed.
    \item Percentage of females divorced.
\end{enumerate}

\section{POI diversity}\label{poi_diversity}
\begin{figure}[H]
        \centering
         \includegraphics[width=1\textwidth]{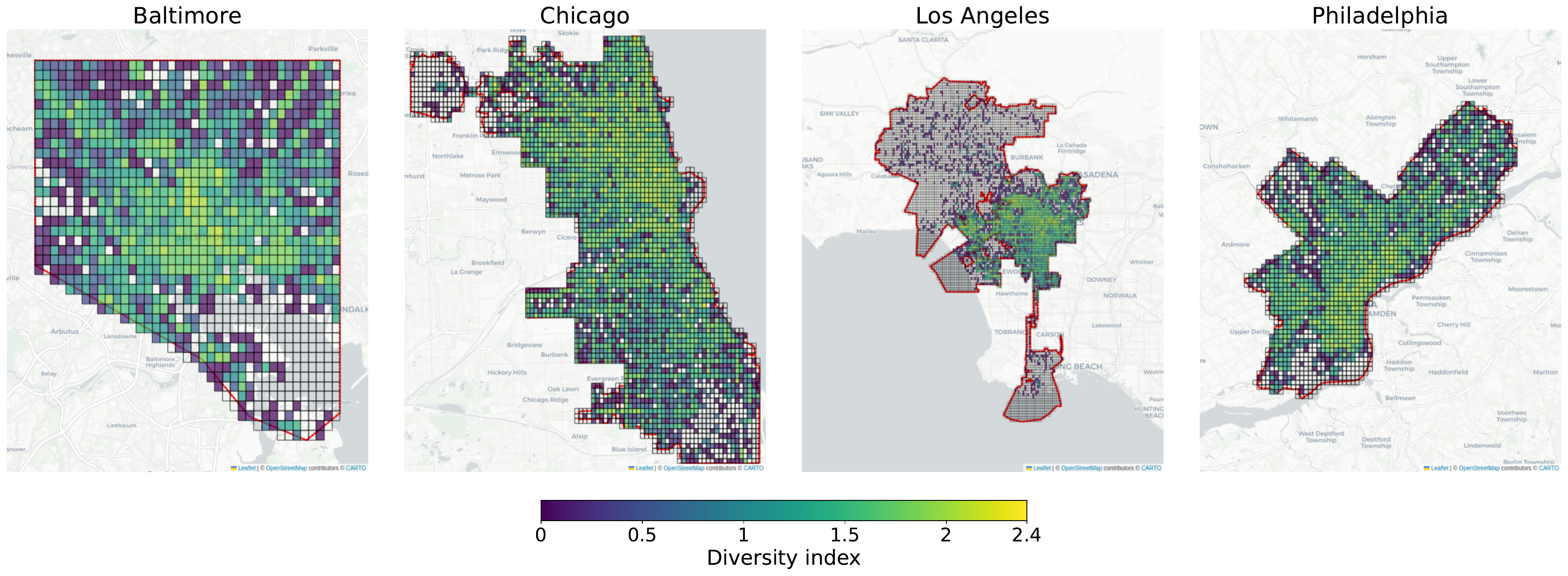}
        \caption{Map of the POI diversity distribution for each of the four U.S.\ cities between 2019 and 2023.}
        \label{fig:poi-div-distrib}
\end{figure}
\newpage
\section{Models hypterparameters}\label{secC}
\begin{table}[h]
    \fontsize{10pt}{10pt}\selectfont
    \centering
    \begin{tabularx}{\linewidth}{c l l l X}
    \toprule
    Model & Hyperparameter & Value & & Description \\
    \midrule
    \multirow{10}{*}{\rotatebox{90}{ConvLSTM}}
    & Num. of layers & 1,2,3*,4,5 & & Total ConvLSTM layers in the neural network. \\
    & Num. of neurons & 28 & & Number of neurons in each hidden layer. \\
    & Kernel size & (3,3)*,(5,5),(7,7) & & Shape of filter applied in convolutional layers. \\
    & Optimizer & Adam & & The optimization algorithm used for training. \\
    & Learning rate & 0.001,0.0001,0.00001* & & The step size for the optimizer. \\
    & Dropout rate & 0.8 & & Probability of dropping a unit during training. \\
    & Batch size & 55 & & Number of samples per gradient update. \\
    & Epochs & 200 & & Number of full passes through the train set. \\
    & Loss function & Binary cross-entropy & & Difference between predictions and targets. \\
    & Activation function & ReLU*,tanh & & Activation function used in the hidden layers. \\
    \midrule
    \multirow{3}{*}{\rotatebox{90}{LR}}
    & Multiclass & ovr & & One-vs-Rest strategy. \\
    & Class weight & Balanced & & The class weights are inversely proportional to their frequency. \\
    & Max iter & 400 & & Maximum number of iterations. \\
    \midrule
    \multirow{3}{*}{\rotatebox{90}{RF}}
    & Num. of trees & 100 & & Number of decision trees in the ensemble. \\
    & Class weight & Balanced & & The class weights are inversely proportional to their frequency. \\
    & Criterion & Entropy*,Gini & & Metric used to evaluate splits. \\
    \midrule
    \multirow{6}{*}{\rotatebox{90}{LSTM}}
    & Num. of layers & 3 & & Total LSTM layers in the neural network. \\
    & Num. of neurons & 28 & & Number of neurons in each hidden layer. \\
    & Optimizer & Adam & & The optimization algorithm used for training. \\
    & Learning rate & 0.0001 & & The step size for the optimizer. \\
    & Loss function & Binary cross-entropy & & Difference between predictions and targets. \\
    & Activation function & ReLU & & Activation function used in the hidden layers. \\
    \bottomrule
    \end{tabularx}
    \caption{Hyperparameter configurations for each model. In the case of hyperparameter tuning, the best-performing values are marked with an asterisk (*).}
    \label{tab:hyperparameters}
\end{table}

\section{Cells included in the analysis}\label{masks}
\begin{figure}[H]
        \centering
         \includegraphics[width=\textwidth]{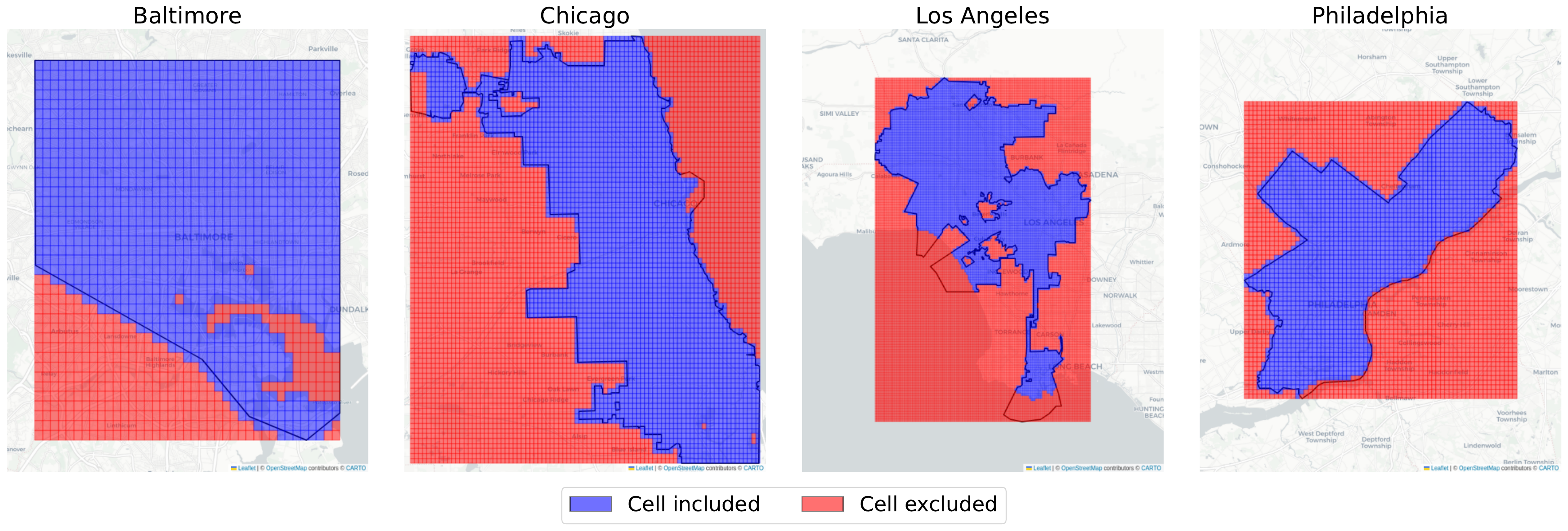}
        \caption{Map of the cells included and excluded from the analysis.}
        \label{fig:masks}
\end{figure}

\section{Average performance at different thresholds}\label{perf_thrs}
\begin{figure}[H]
        \centering
         \includegraphics[width=0.885\textwidth]{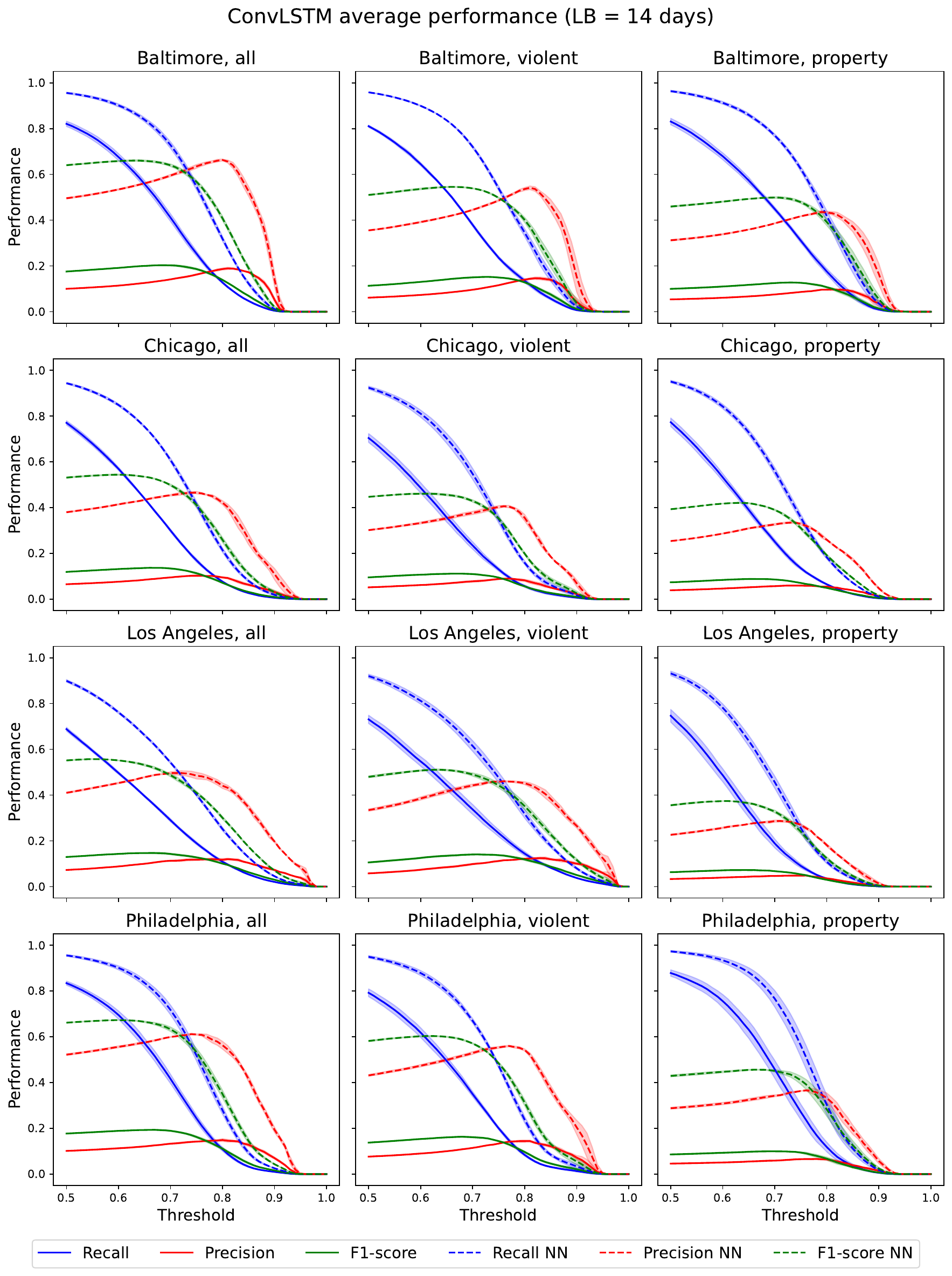}
        \caption{Average performance on the test set using ConvLSTM and an LB period of 14 days. The shaded area indicates the standard deviation over the four different random seeds.}
        \label{fig:convlstm_14_thrs}
\end{figure}

\begin{figure}[H]
        \centering
         \includegraphics[width=0.95\textwidth]{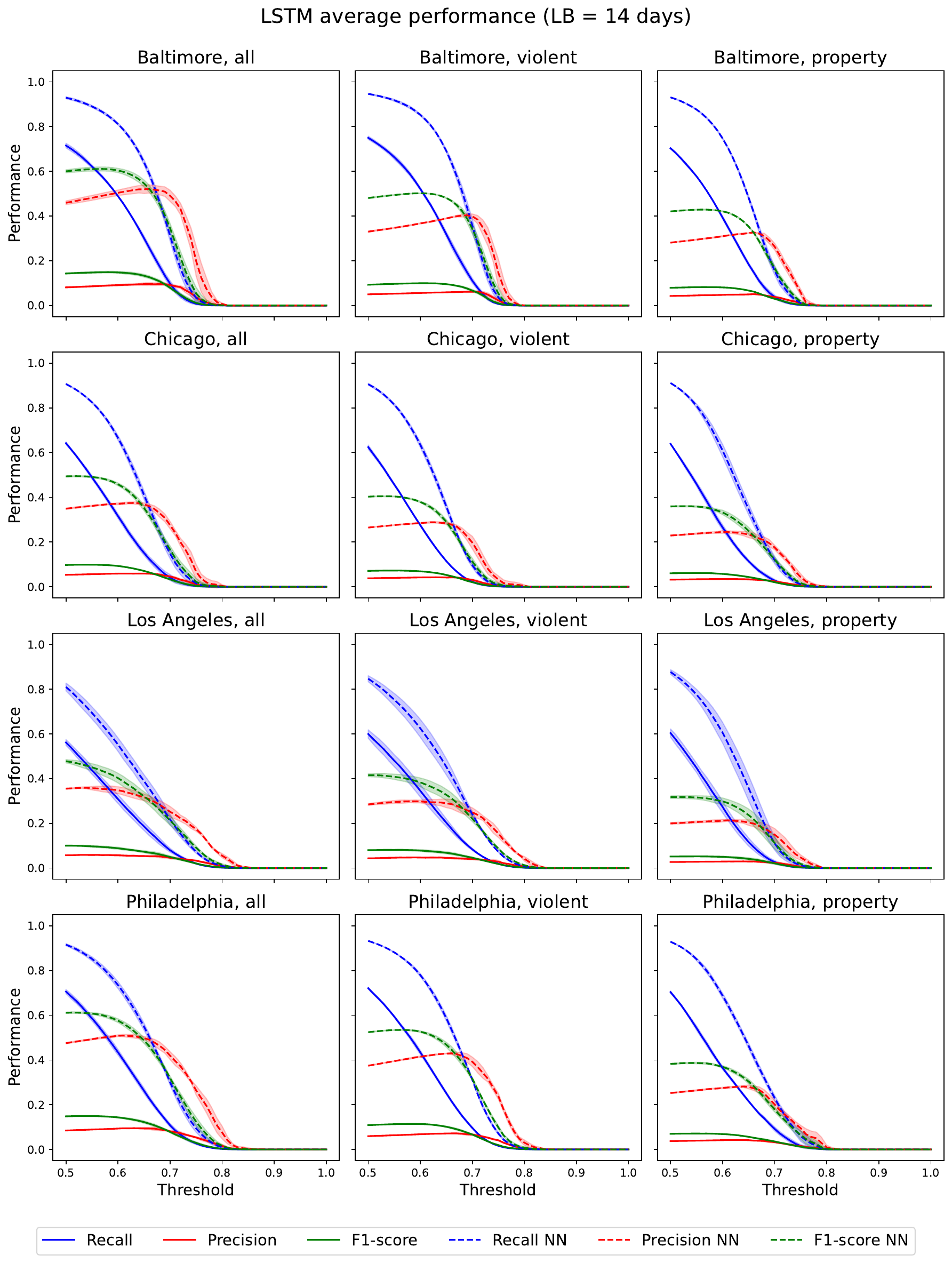}
        \caption{Average performance on the test set using LSTM and an LB period of 14 days. The shaded area indicates the standard deviation over the four different random seeds.}
        \label{fig:lstm_14_thrs}
\end{figure}

\begin{figure}[H]
        \centering
         \includegraphics[width=0.95\textwidth]{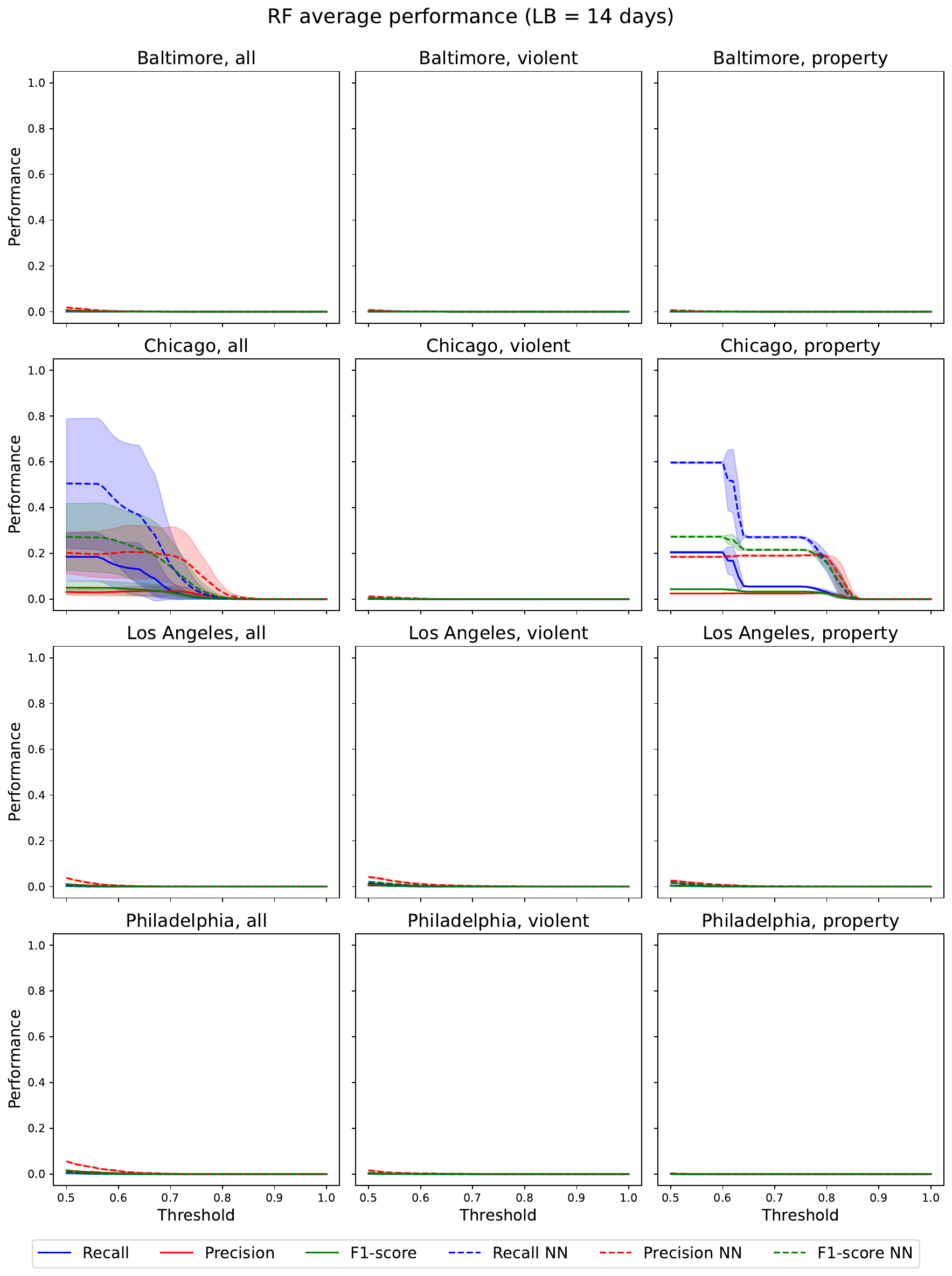}
        \caption{Average performance on the test set using RF and an LB period of 14 days. The shaded area indicates the standard deviation over the four different random seeds.}
        \label{fig:rf_14_thrs}
\end{figure}

\begin{figure}[H]
        \centering
         \includegraphics[width=0.96\textwidth]{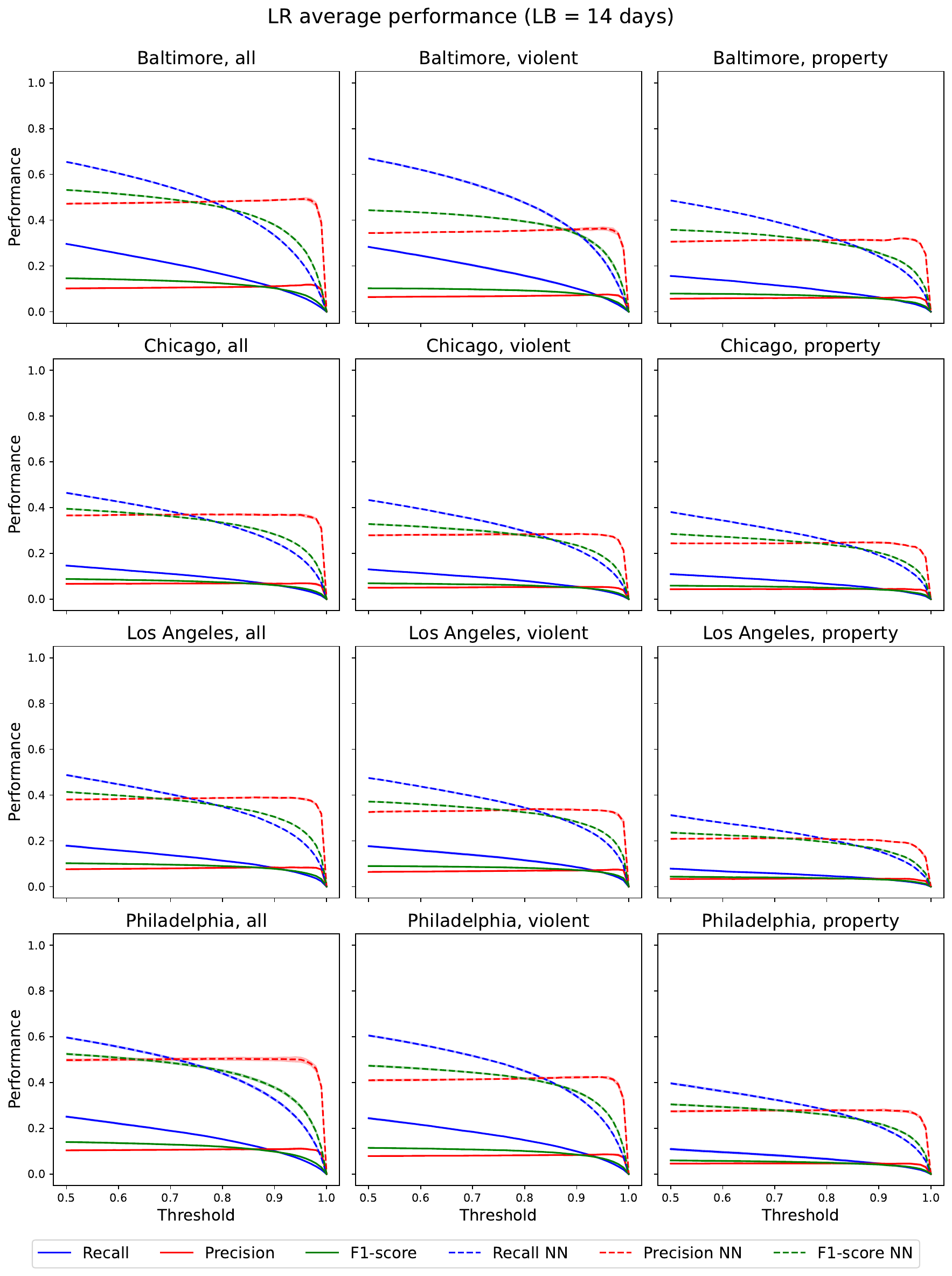}
        \caption{Average performance on the test set using LR and an LB period of 14 days. The shaded area indicates the standard deviation over the four different random seeds.}
        \label{fig:lr_14_thrs}
\end{figure}

\begin{figure}[H]
        \centering
         \includegraphics[width=0.96\textwidth]{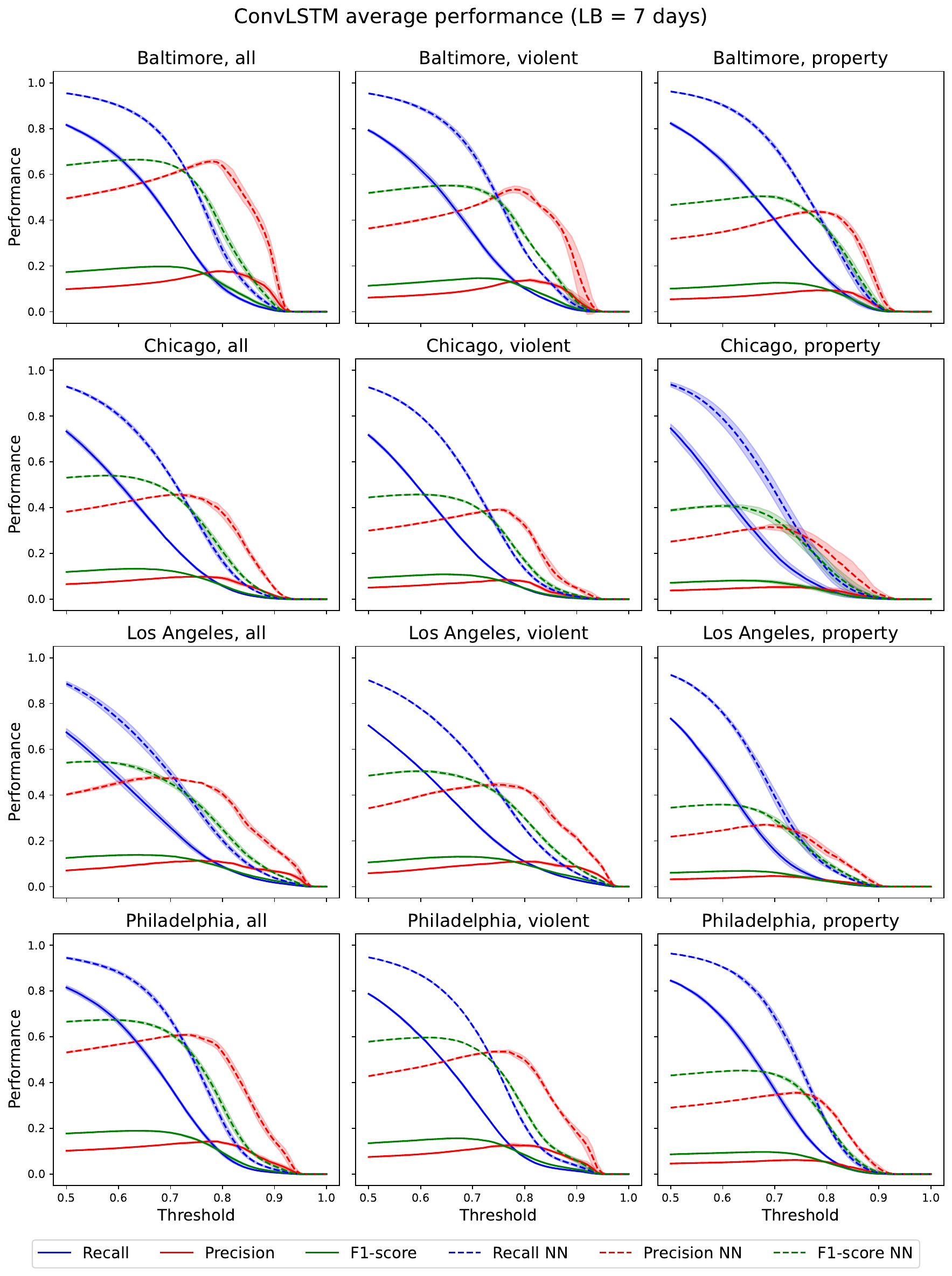}
        \caption{Average performance on the test set using ConvLSTM and an LB period of 7 days. The shaded area indicates the standard deviation over the four different random seeds.}
        \label{fig:convlstm_7_thrs}
\end{figure}

\begin{figure}[H]
        \centering
         \includegraphics[width=0.96\textwidth]{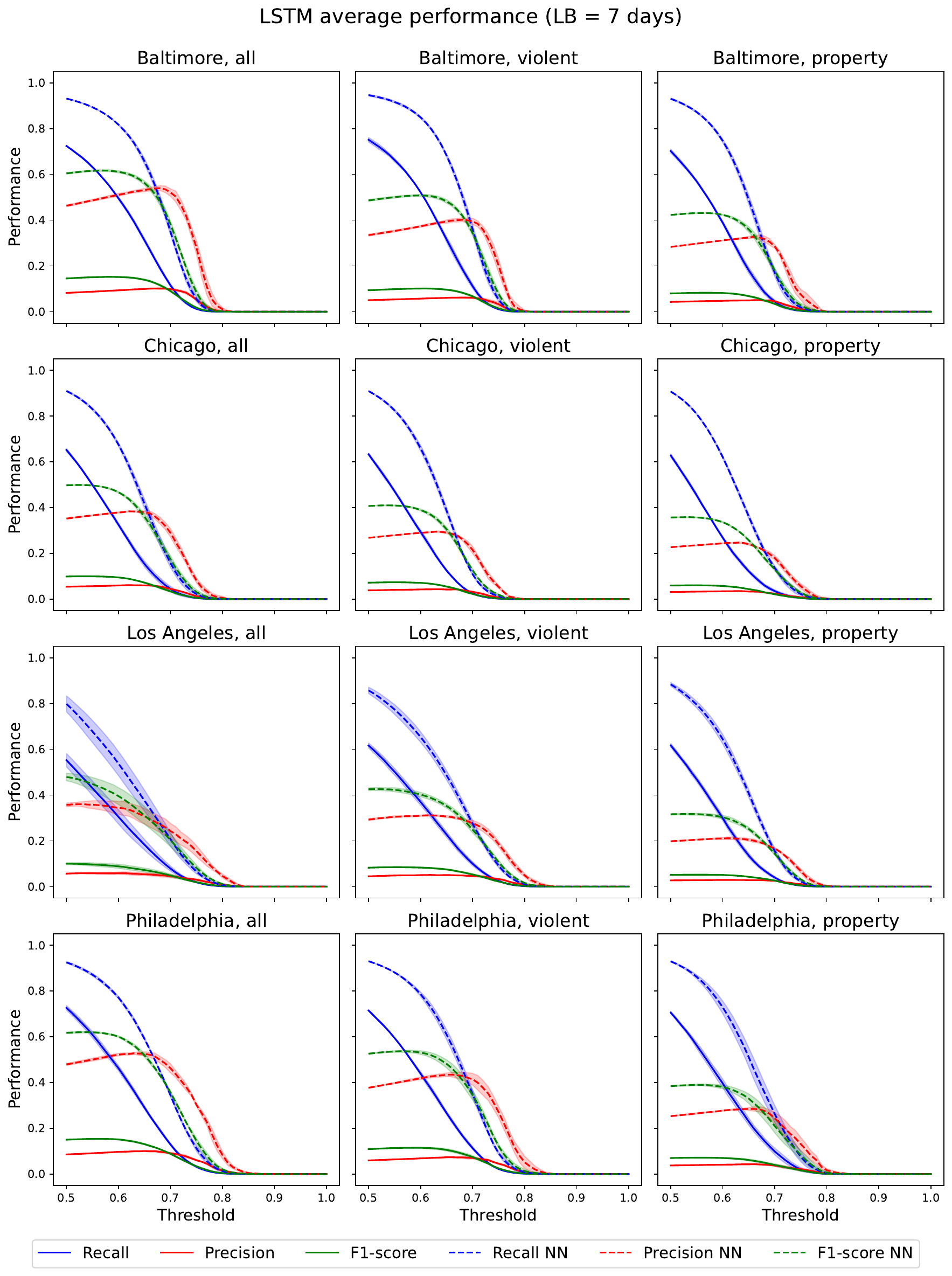}
        \caption{Average performance on the test set using LSTM and an LB period of 7 days. The shaded area indicates the standard deviation over the four different random seeds.}
        \label{fig:lstm_7_thrs}
\end{figure}

\begin{figure}[H]
        \centering
         \includegraphics[width=0.96\textwidth]{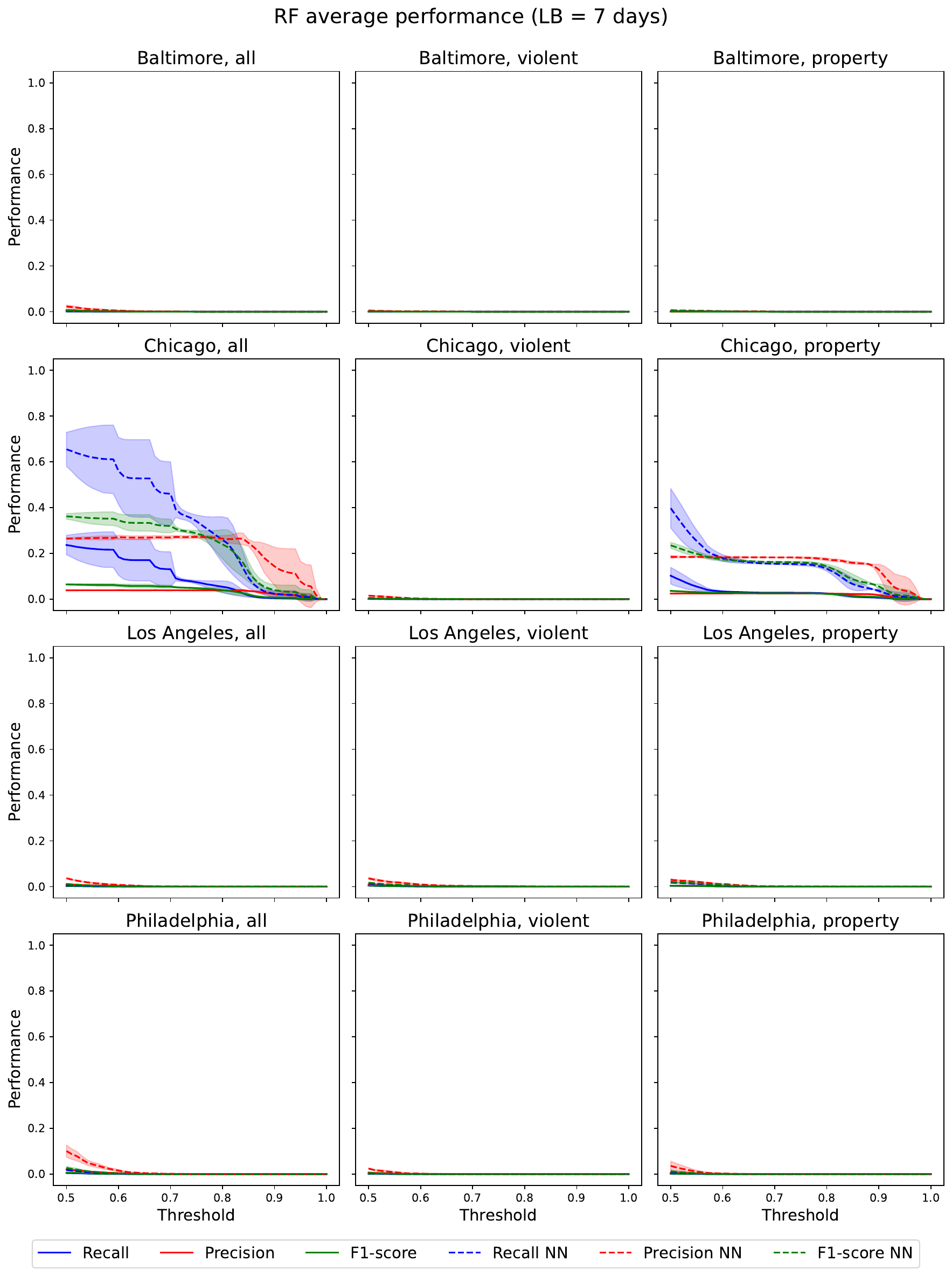}
        \caption{Average performance on the test set using RF and an LB period of 7 days. The shaded area indicates the standard deviation over the four different random seeds.}
        \label{fig:rf_7_thrs}
\end{figure}

\begin{figure}[H]
        \centering
         \includegraphics[width=0.96\textwidth]{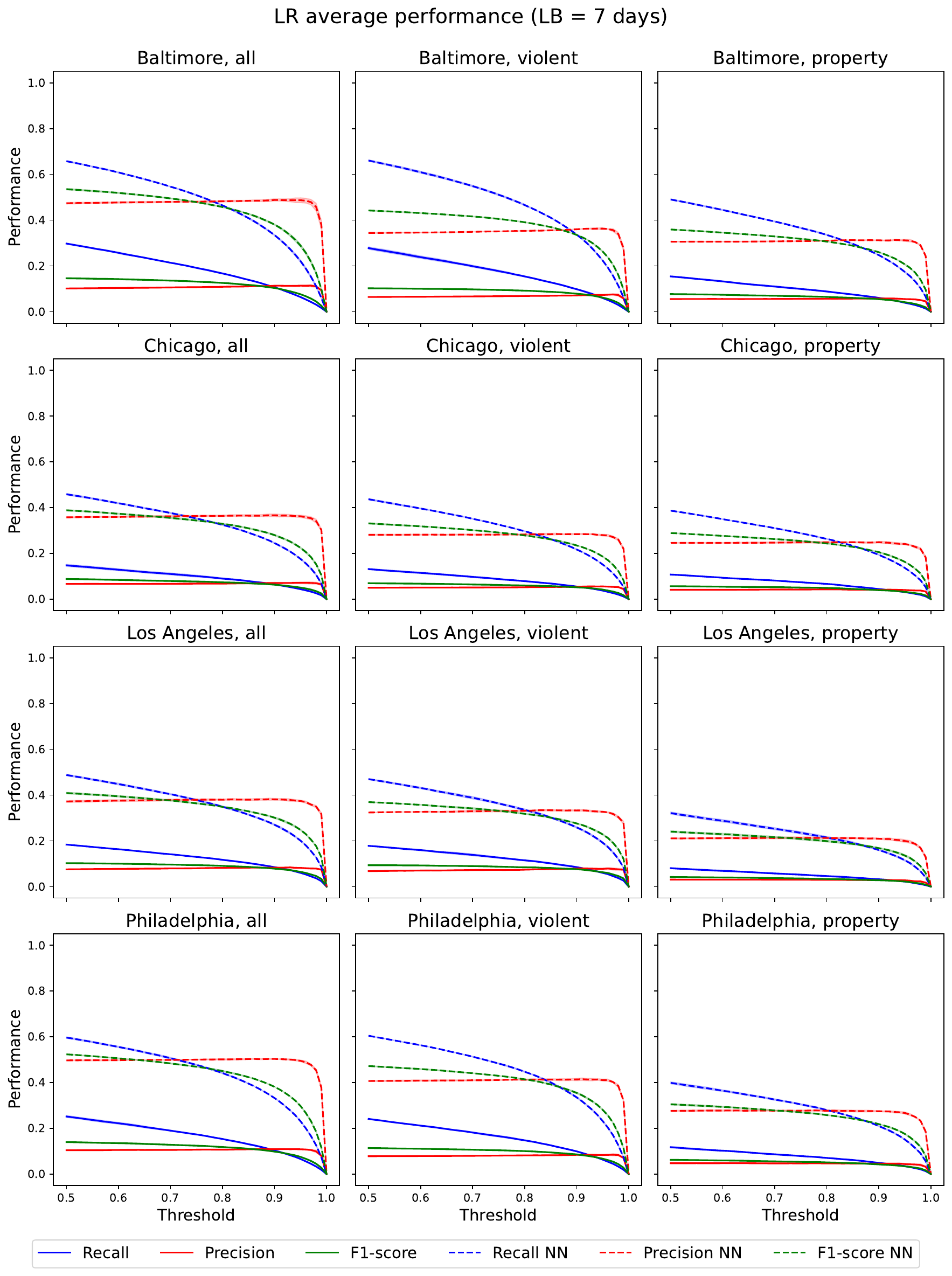}
        \caption{Average performance on the test set using LR and an LB period of 7 days. The shaded area indicates the standard deviation over the four different random seeds.}
        \label{fig:lr_7_thrs}
\end{figure}

\begin{figure}[H]
        \centering
         \includegraphics[width=0.96\textwidth]{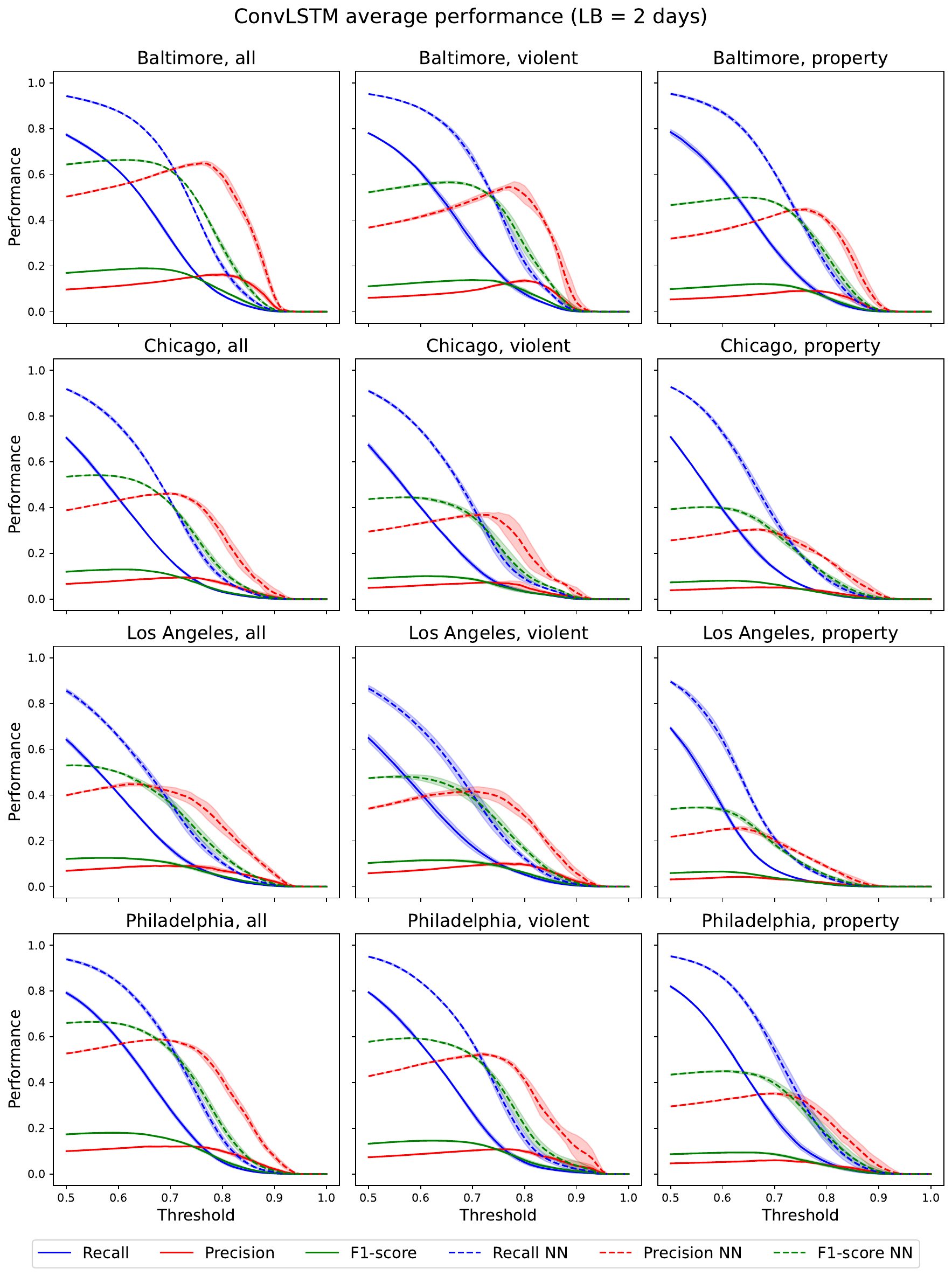}
        \caption{Average performance on the test set using ConvLSTM and an LB period of 2 days. The shaded area indicates the standard deviation over the four different random seeds.}
        \label{fig:convlstm_2_thrs}
\end{figure}

\begin{figure}[H]
        \centering
         \includegraphics[width=0.96\textwidth]{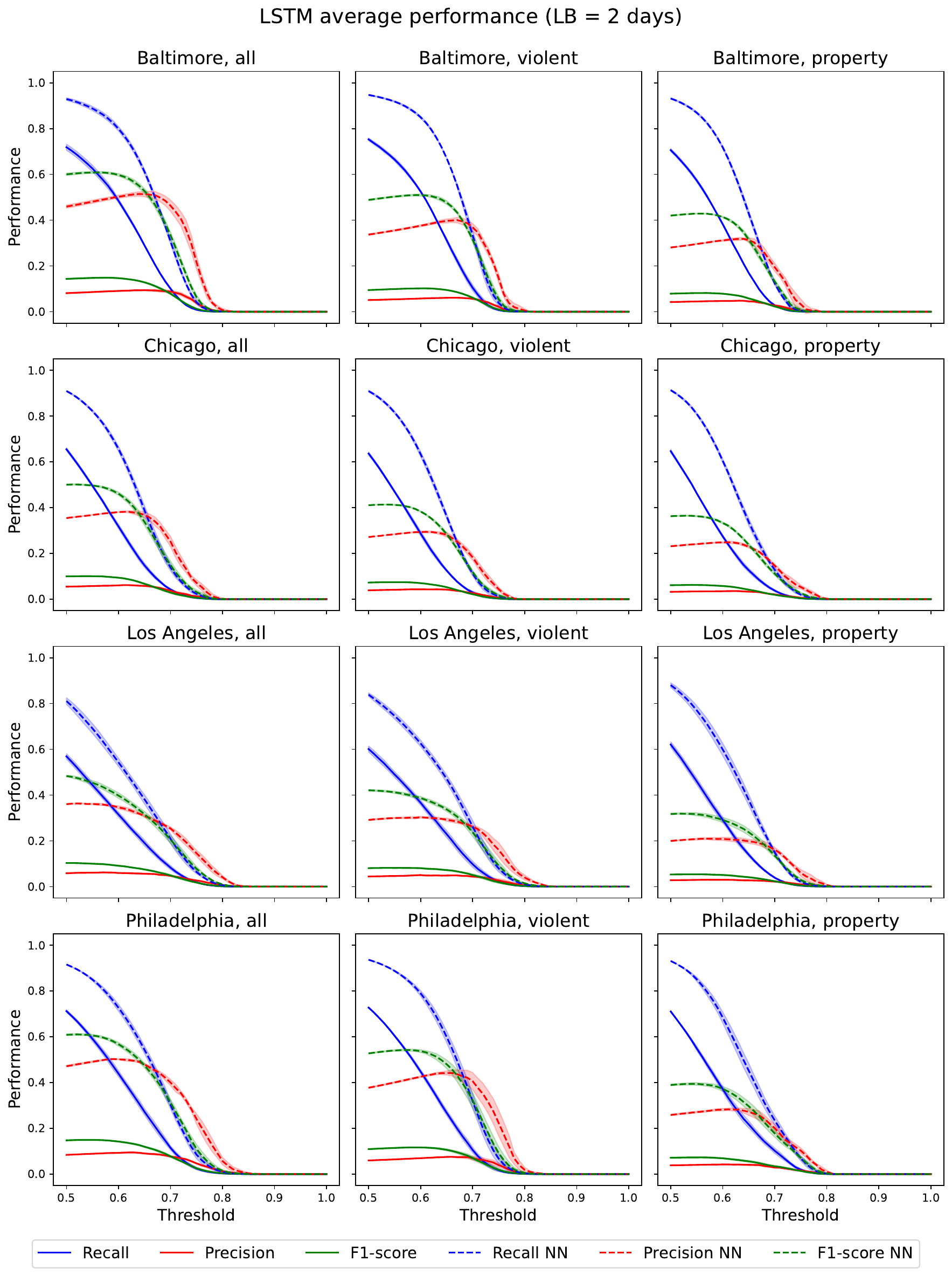}
        \caption{Average performance on the test set using LSTM and an LB period of 2 days. The shaded area indicates the standard deviation over the four different random seeds.}
        \label{fig:lstm_2_thrs}
\end{figure}

\begin{figure}[H]
        \centering
         \includegraphics[width=0.96\textwidth]{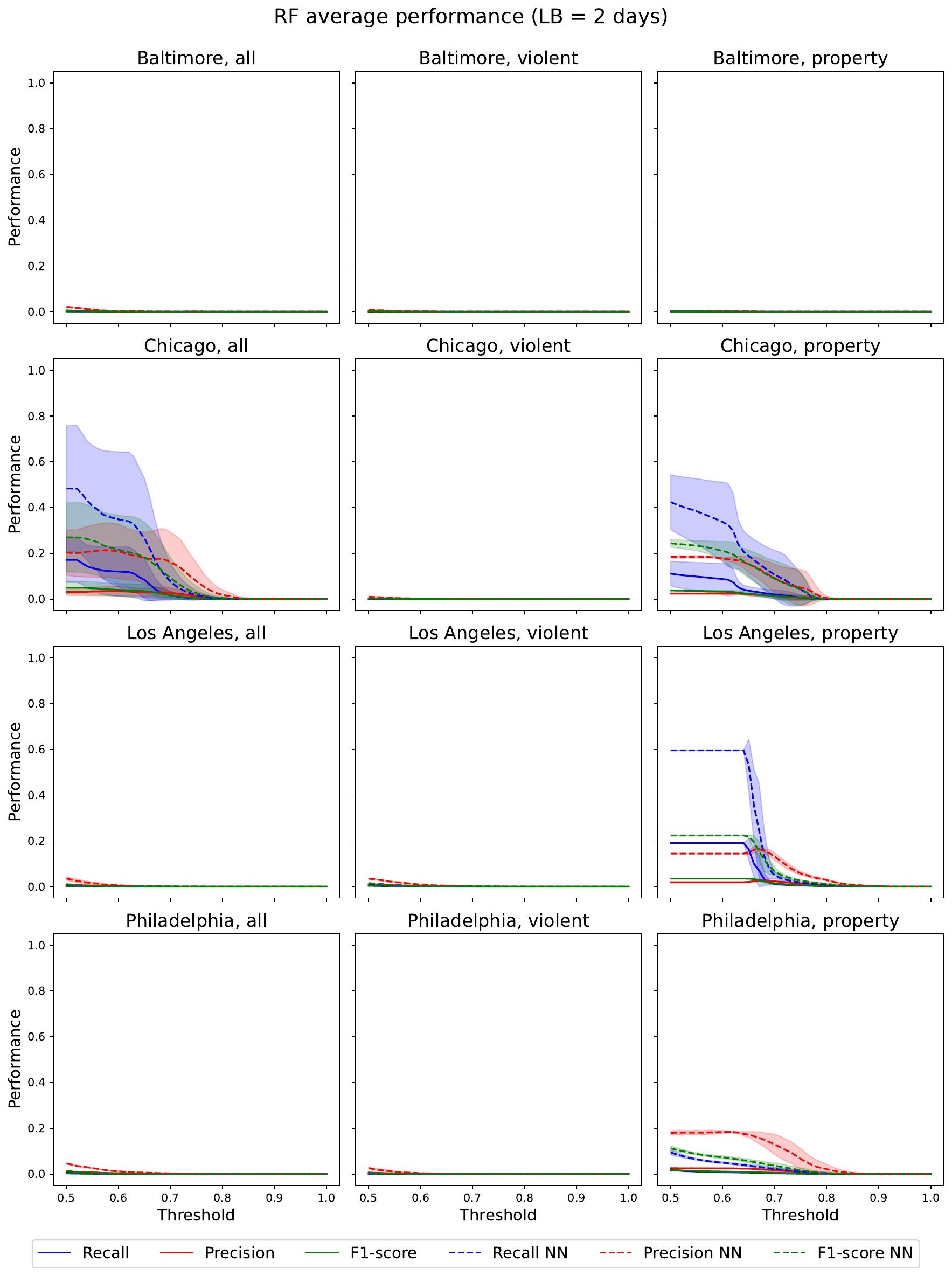}
        \caption{Average performance on the test set using RF and an LB period of 2 days. The shaded area indicates the standard deviation over the four different random seeds.}
        \label{fig:rf_2_thrs}
\end{figure}

\begin{figure}[H]
        \centering
         \includegraphics[width=0.97\textwidth]{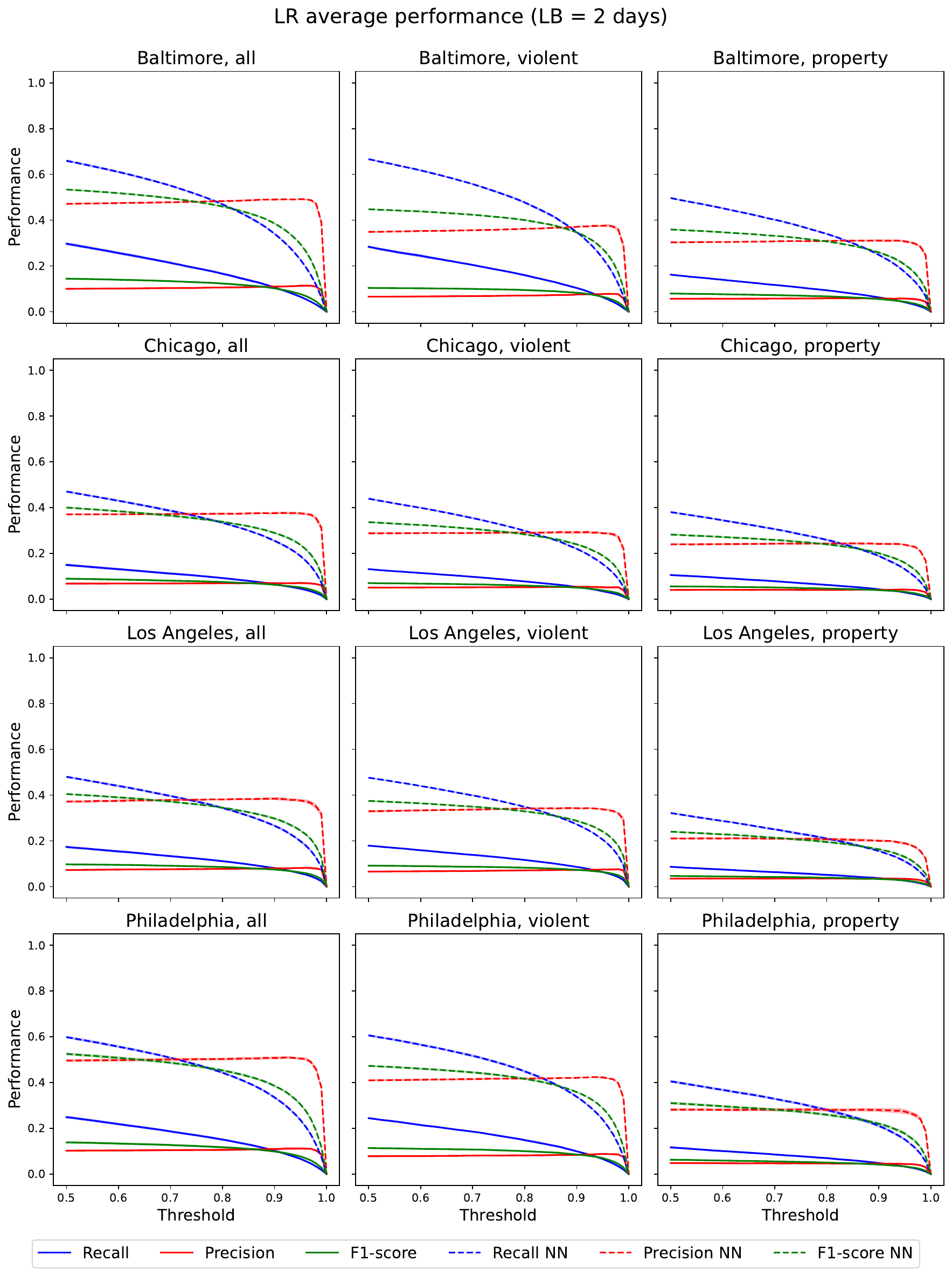}
        \caption{Average performance on the test set using LR and an LB period of 2 days. The shaded area indicates the standard deviation over the four different random seeds.}
        \label{fig:lr_2_thrs}
\end{figure}

\begin{figure}[H]
        \centering
         \includegraphics[width=0.97\textwidth]{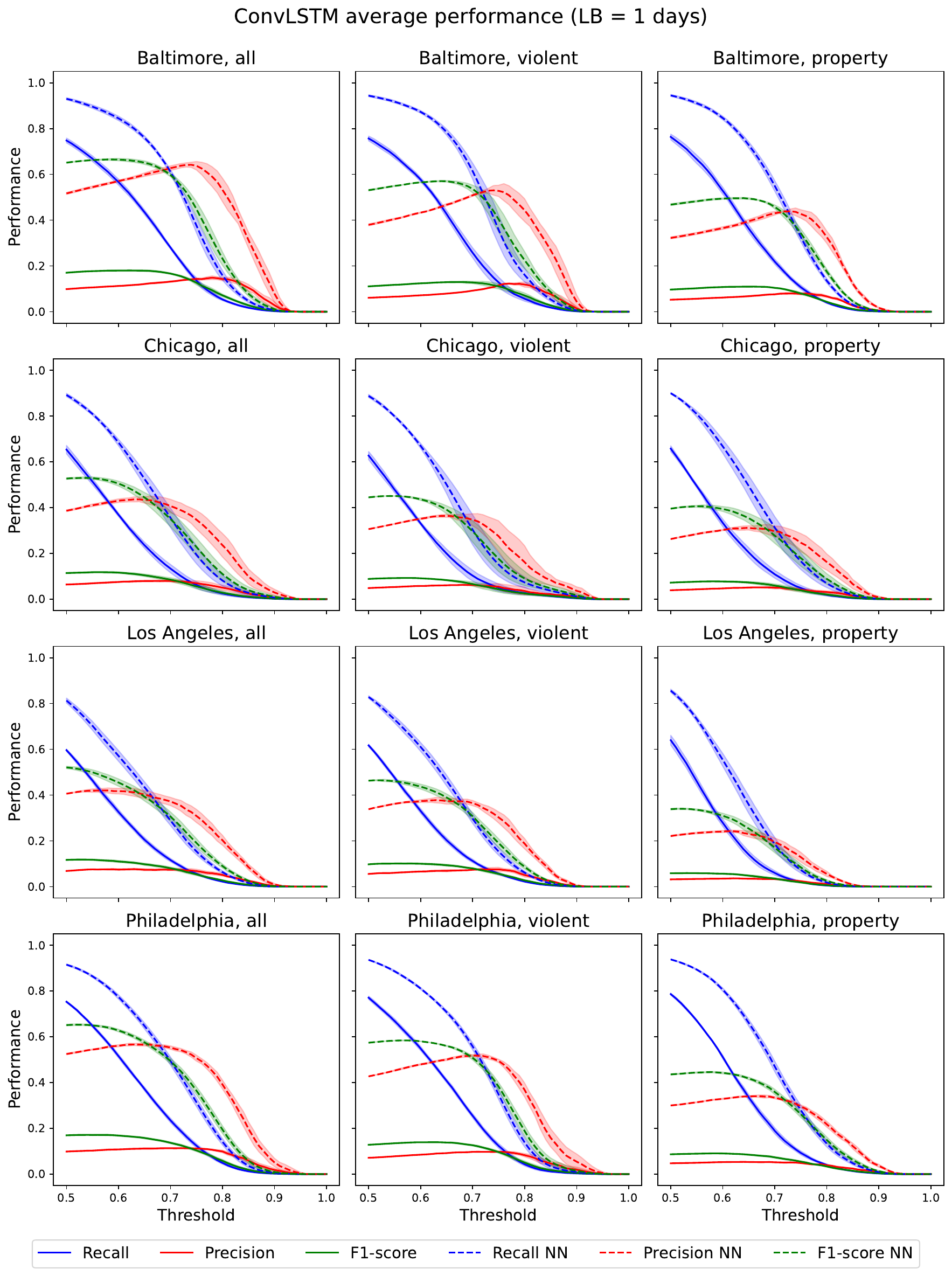}
        \caption{Average performance on the test set using ConvLSTM and an LB period of 1 day. The shaded area indicates the standard deviation over the four different random seeds.}
        \label{fig:convlstm_1_thrs}
\end{figure}

\begin{figure}[H]
        \centering
         \includegraphics[width=0.97\textwidth]{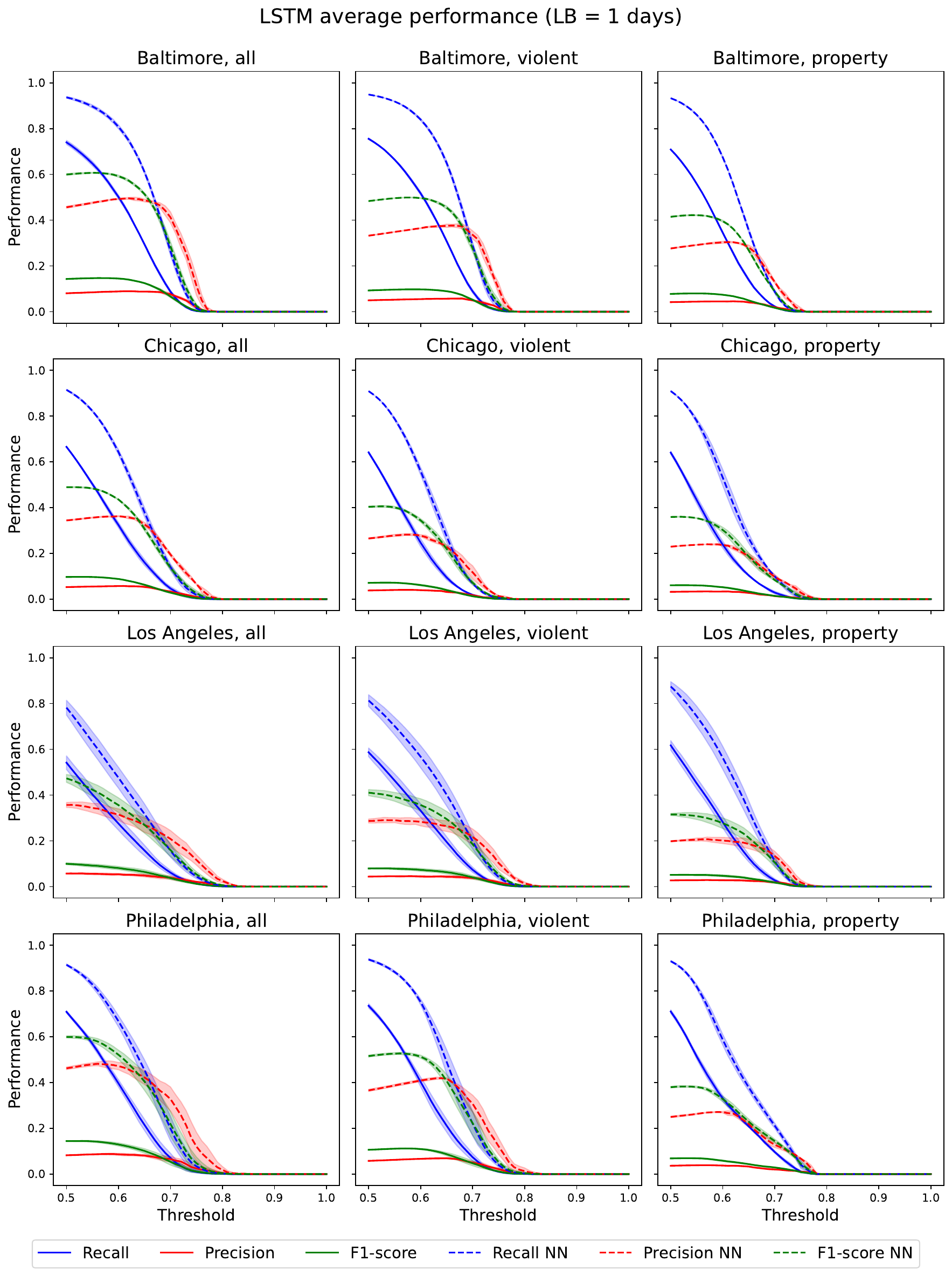}
        \caption{Average performance on the test set using LSTM and an LB period of 1 day. The shaded area indicates the standard deviation over the four different random seeds.}
        \label{fig:lstm_1_thrs}
\end{figure}

\begin{figure}[H]
        \centering
         \includegraphics[width=0.97\textwidth]{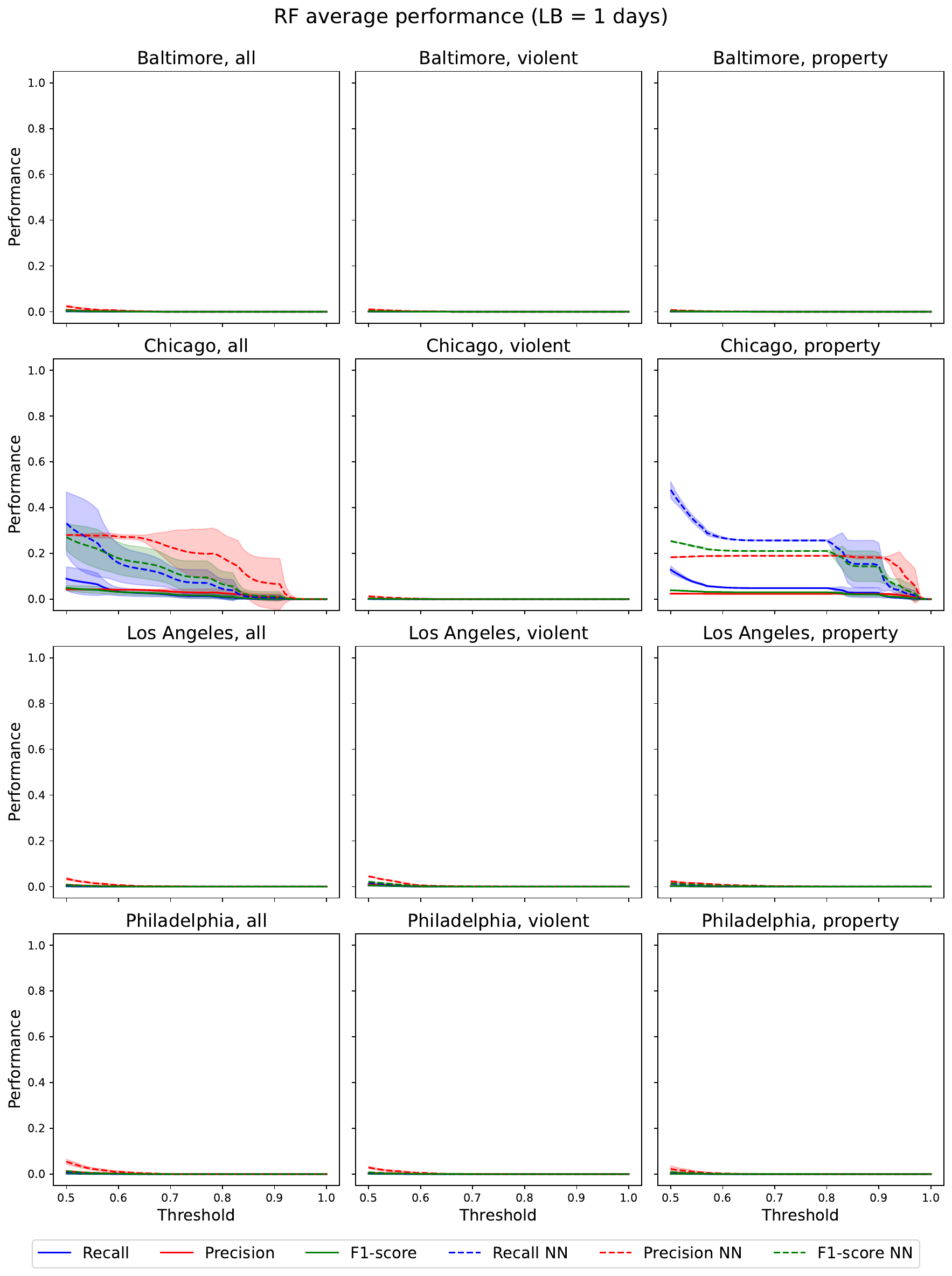}
        \caption{Average performance on the test set using RF and an LB period of 1 day. The shaded area indicates the standard deviation over the four different random seeds.}
        \label{fig:rf_1_thrs}
\end{figure}

\begin{figure}[H]
        \centering
         \includegraphics[width=0.97\textwidth]{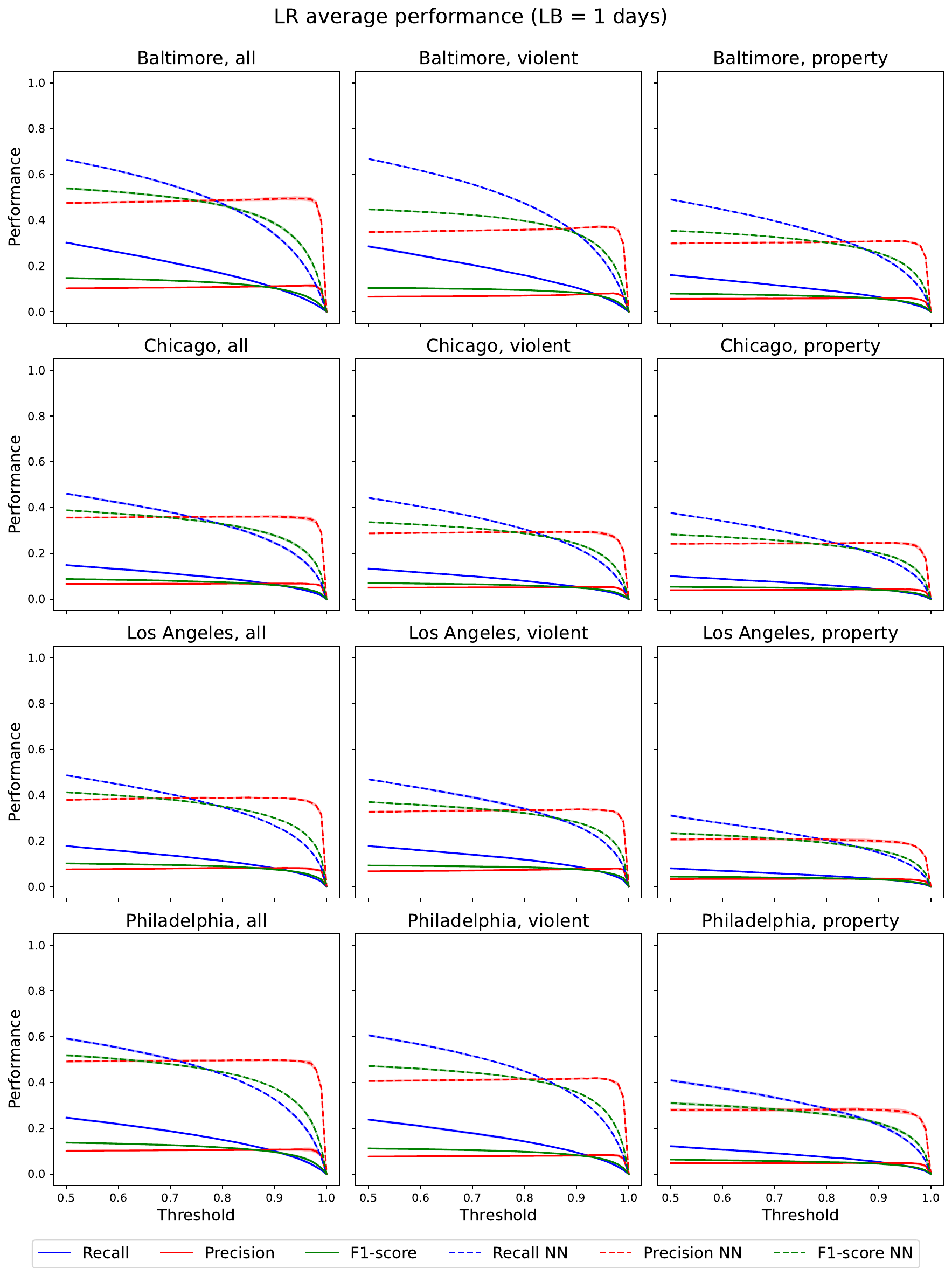}
        \caption{Average performance on the test set using LR and an LB period of 1 day. The shaded area indicates the standard deviation over the four different random seeds.}
        \label{fig:lr_1_thrs}
\end{figure}

\newpage
\section{General performance for 7- and 1-day LBs}\label{LB7andLB1}
\begin{figure}[H]
        \centering
         \includegraphics[width=0.83\textwidth]{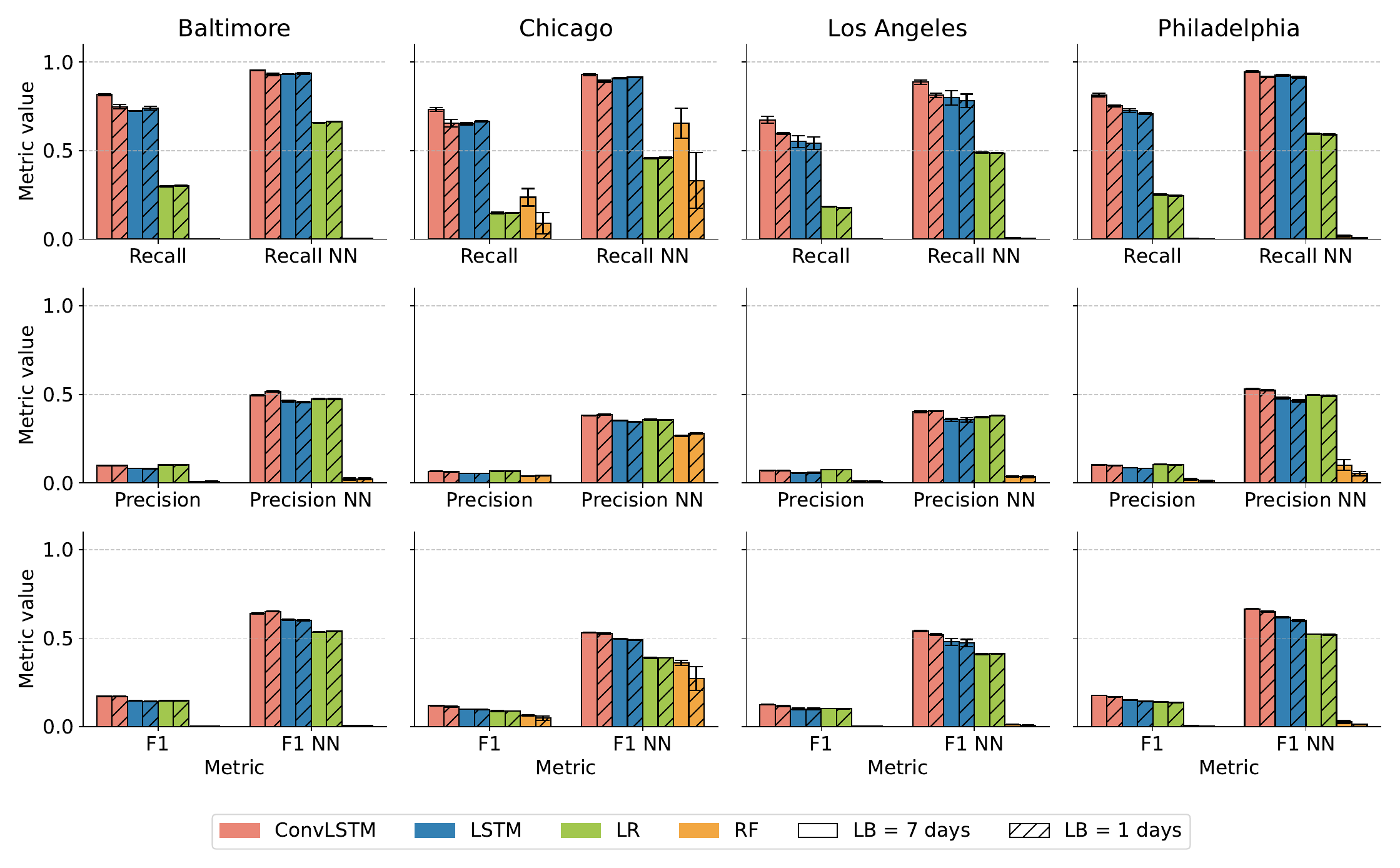}
        \caption{Average performance on the test set for our model and three baselines with and without taking into account the NN cells, considering all five crimes together and using LB periods of 7 and 1 day. Error bars indicate the standard deviation.}
        \label{fig:perf_all_lb7_lb1}
\end{figure}
\begin{figure}[H]
        \centering
         \includegraphics[width=0.83\textwidth]{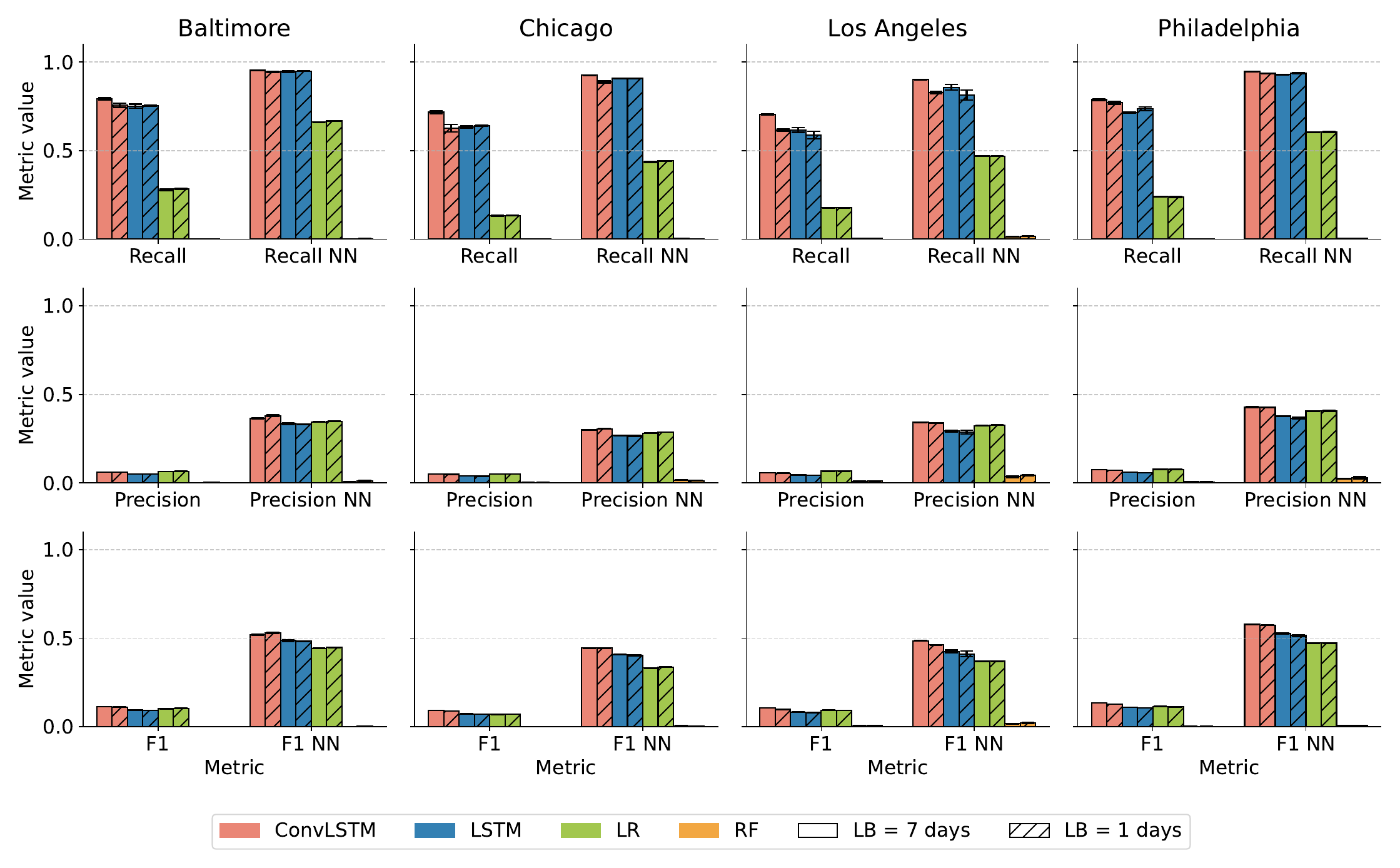}
        \caption{Average performance on the test set for our model and three baselines with and without taking into account the NN cells, considering only violent crimes and using LB periods of 7 and 1 day. Error bars indicate the standard deviation.}
        \label{fig:perf_violent_lb7_lb1}
\end{figure}

\begin{figure}[H]
        \centering
         \includegraphics[width=0.84\textwidth]{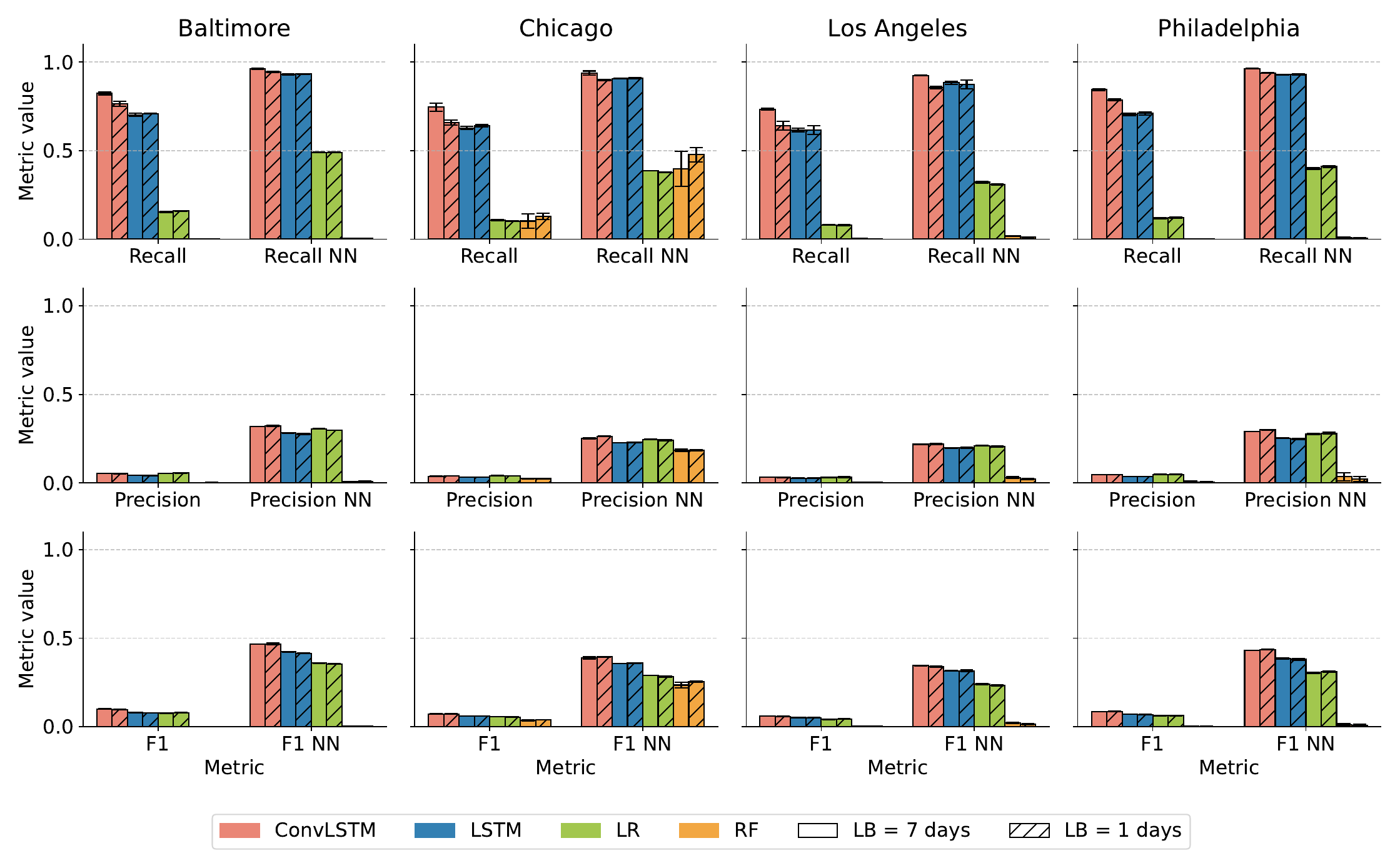}
        \caption{Average performance on the test set for our model and three baselines with and without taking into account the NN cells, considering only property crimes and using LB periods of 7 and 1 day. Error bars indicate the standard deviation.}
        \label{fig:perf_property_lb7_lb1}
\end{figure}

\section{Effect of mobility features for 7- and 1-day LBs}\label{features_extra}

\begin{table}[h]
    \fontsize{11pt}{11pt}\selectfont
    \centering
    \begin{tabular}{ccccccccccc}
    \toprule
     &  & \multicolumn{4}{c}{CMS - CM} && \multicolumn{4}{c}{CMS - CS} \\
    \cmidrule(r){3-6}
    \cmidrule(r){8-11}
    LB & Metric & Bal & Chi & Las & Phi && Bal & Chi & Las & Phi \\
    \midrule
    \multirow{6}{*}{\rotatebox{90}{7 days}}
        & Rec& -2.025 & -8.628 & -1.039 & -2.083 && 0.196 & -1.976 & -0.612 & -1.545 \\
        & Prec   & 8.264 & 7.814 & 5.247 & 4.107 && 1.193 & 4.854 & 1.509 & 1.957 \\
        & F1    & 7.171 & 6.274 & 4.632 & 3.507 && 1.077 & 4.304 & 1.528 & 1.421 \\
        & Rec NN  & -0.488 & -2.552 & -0.873 & -0.597 && -0.016 & -0.638 & 0.116 & -0.681 \\
        & Prec NN & 2.949 & 4.411 & 1.937 & 3.156 && -0.359 & 0.182 & -0.203 & 0.866 \\
        & F1 NN  & 1.826 & 2.310 & 1.169 & 1.776 && -0.222 & -0.006 & -0.053 & 0.249 \\
    \midrule
    \multirow{6}{*}{\rotatebox{90}{1 day}}
        & Rec  & 3.597 & -5.895 & 28.822 & 1.878 && 4.945 & -0.352 & -5.450 & -3.413 \\
        & Prec   & 3.859 & 10.652 & 1.875 & 4.244 && 3.662 & 6.491 & 3.913 & 4.925 \\
        & F1    & 3.719 & 9.882 & 10.795 & 4.028 && 3.624 & 6.613 & 2.240 & 3.649 \\
        & Rec NN & 1.147 & -1.084 & 21.804 & 0.599 && 1.201 & 0.335 & -2.406 & -1.532 \\
        & Prec NN  & 1.924 & 6.311 & 5.507 & 1.952 && 0.864 & 2.999 & 2.010 & 2.964 \\
        & F1 NN   & 1.533 & 4.587 & 13.724 & 1.549 && 0.973 & 2.511 & 0.375 & 1.397 \\
    \bottomrule
    \end{tabular}
    \caption{Percentage difference in model performance between the CS and CMS configurations (to assess the effect of M), and between the CM and CMS configurations (to assess the effect of S), for each city and using LB periods of 7 and 1 day. Positive values indicate improved performance, while negative values indicate a decrease.}
    \label{percent-diff-extra}
\end{table}

\begin{figure}[h]
        \includegraphics[width=1\textwidth]{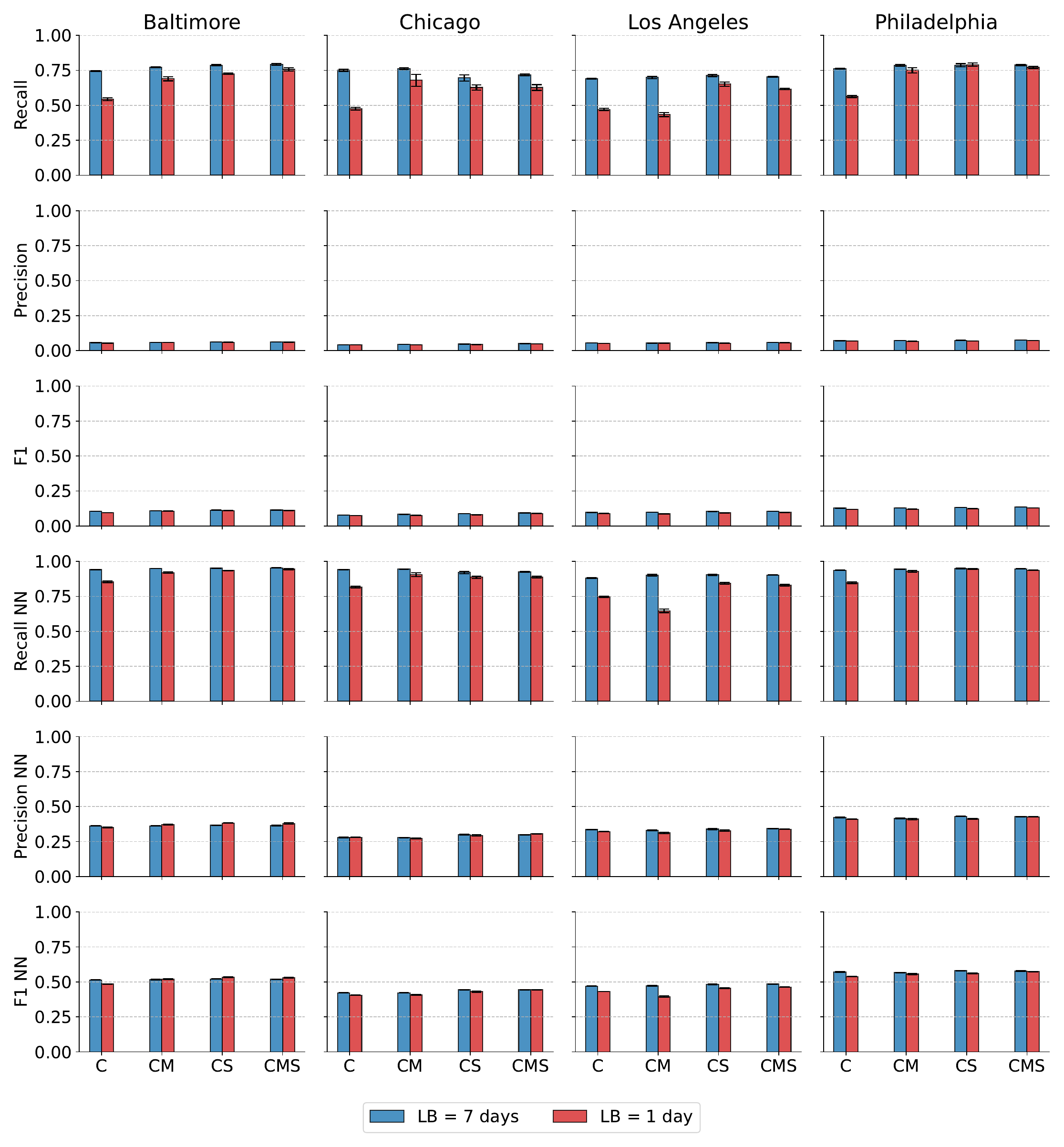}
        \caption{Average performance of the ConvLSTM model on the test set, trained using four different feature sets and LB periods of 7 and 1 day. Error bars indicate the standard deviation.}
        \label{mobility-effects-extra}
\end{figure}


\end{document}